\documentclass[10pt,twocolumn,letterpaper]{article}

\usepackage{cvpr}      %

\usepackage{amsmath,amsfonts,bm}

\def\eqref#1{equation~\ref{#1}}

\def\1{\bm{1}}

\DeclareMathAlphabet{\mathsfit}{\encodingdefault}{\sfdefault}{m}{sl}
\SetMathAlphabet{\mathsfit}{bold}{\encodingdefault}{\sfdefault}{bx}{n}

\usepackage{ifthen}
\PassOptionsToPackage{numbers, compress}{natbib}

\usepackage[utf8]{inputenc} %
\usepackage[T1]{fontenc}    %
\usepackage{url}            %
\usepackage{booktabs}       %
\usepackage{amsfonts}       %
\usepackage{nicefrac}       %
\usepackage{microtype}      %
\usepackage{xcolor}         %
\usepackage{pgfmath}
\usepackage{expl3}

\usepackage{stmaryrd}
\usepackage{tabularx}
\usepackage{xfrac}
\usepackage[inline]{enumitem}
\usepackage[export]{adjustbox}
\usepackage{soul}
\usepackage{graphicx}

\usepackage{caption}
\usepackage{subcaption}
\usepackage{fancyvrb}
\usepackage{fvextra}

\usepackage{comment}

\usepackage{amsmath}
\usepackage{amssymb}
\usepackage{array}
\usepackage{multirow}
\usepackage{color}
\usepackage{colortbl}
\usepackage{framed}
\usepackage{bm}
\usepackage{placeins} %
\usepackage{bbm}
\usepackage{xspace}
\usepackage{enumitem}
\usepackage{algorithm}
\usepackage{listings}
\usepackage{url}
\usepackage{booktabs}
\usepackage{pifont} 
\usepackage{xfrac}
\usepackage{rotating}
\usepackage{makecell}
\usepackage[accsupp]{axessibility}

\usepackage{arydshln}

\definecolor{Gray}{gray}{0.9}
\definecolor{LightCyan}{rgb}{0.88,0.95,1}
\definecolor{blond}{rgb}{0.98, 0.94, 0.75}
\definecolor{green}{rgb}{0.0, 0.62, 0.42}
\definecolor{green_into_the_wild}{rgb}{0.0, 1.0, 0.0}
\definecolor{light_green_into_the_wild}{rgb}{0.3, 0.7, 0.3}
\definecolor{red_into_the_wild}{rgb}{1.0, 0.0, 0.0}
\definecolor{yellow_into_the_wild}{rgb}{0.7, 0.7, 0.0}
\definecolor{light_blue_into_the_wild}{rgb}{0.0, 0.5, 1.0}
\definecolor{fuchsia_into_the_wild}{rgb}{0.6, 0.0, 0.6}
\definecolor{light_fuchsia_into_the_wild}{rgb}{0.4, 0.2, 0.2}
\definecolor{lightgray}{gray}{0.93}

\DeclareMathOperator*{\avg}{avg}

\newcommand{\noiseSymbol}{$\mathcal{N}$\textit{oise}}
\newcommand{\imProjSymbol}{$\mathcal{M}\textit{emory}$}%
\newcommand{\modalityGapNoStrategy}{no mitig.}

\definecolor{cvprblue}{rgb}{0.21,0.49,0.74}
\usepackage[pagebackref,breaklinks,colorlinks,allcolors=cvprblue]{hyperref}

\def\paperID{*****}
\def\confName{CVPR}
\def\confYear{2026}

\title{One Patch to Caption Them All: A Unified Zero-Shot Captioning Framework}

\author{
\begin{tabular}{cccc}
Lorenzo Bianchi\textsuperscript{1,2,*} &
Giacomo Pacini\textsuperscript{1,2,*} &
Fabio Carrara\textsuperscript{1} &
Nicola Messina\textsuperscript{1} \\
[0.3em]
\multicolumn{1}{c}{} &
Giuseppe Amato\textsuperscript{1} &
Fabrizio Falchi\textsuperscript{1} &
\multicolumn{1}{c}{}
\end{tabular}
\\[1.2em]
{\small \textsuperscript{1}ISTI-CNR, Italy} \hspace{25pt}
{\small \textsuperscript{2}University of Pisa, Italy}\hspace{25pt}
{\small \textsuperscript{*}Equal contribution}
}

\newcommand{\methodName}{Patch-ioner}

\begin{document}

\maketitle

\begin{abstract}
Zero-shot captioners are recently proposed models that utilize common-space vision-language representations to caption images without relying on paired image-text data.
To caption an image, they proceed by textually decoding a text-aligned image feature, but they limit their scope to global representations and whole-image captions.
We present a unified framework for zero-shot captioning that shifts from an image-centric to a patch-centric paradigm, enabling the captioning of arbitrary regions without the need of region-level supervision.
Instead of relying on global image representations, we treat individual patches as atomic captioning units and aggregate them to describe arbitrary regions, from single patches to non-contiguous areas and entire images.
We analyze the key ingredients that enable current latent captioners to work in our novel proposed framework.
Experiments demonstrate that backbones producing meaningful, dense visual features, such as DINO, are key to achieving state-of-the-art performance in multiple region-based captioning tasks.
Compared to other baselines and state-of-the-art competitors, our models achieve better performance on zero-shot dense captioning and region-set captioning.
We also introduce a new trace captioning task that further demonstrates the effectiveness of patch-wise semantic representations for flexible caption generation. Data and code are available at \url{https://paciosoft.com/Patch-ioner/}.

\end{abstract}
    
\section{Introduction}

Image captioning is one of the most representative tasks in vision-language understanding and has reached outstanding accuracy thanks to the availability of pre-trained vision-language backbones and large paired image-text datasets. In its basic formulation, a captioning model takes a full image as input and autonomously decides which elements must be described and up to what degree.
To enable user guidance and produce more targeted descriptions, some previous works proposed region-level captioning methods~\citep{johnson2016densecap,cornia2019show}, which take as an additional input a spatial indication (e.g., bounding boxes) specifying which image regions have to be described and, possibly, in which order.

These region-level captioning methods require expensive manually labeled data to fully supervise the model. Indeed, each sequence or set of bounding boxes for a given image should correspond to a manually written ground-truth caption describing those objects. This fully supervised solution does not scale properly.

In this paper, we propose a perspective shift that enables us to perform region-level captioning with arbitrary spatial granularity --- from a single image patch up to the entire image --- %
\textit{without requiring any form of %
region-level supervision or paired text-region data}. 
Specifically, instead of sticking with the classic setup where the subject of a captioning method is the \textit{image} --- then potentially conditioned on a set of sub-regions --- we instead build on two straightforward yet powerful ideas: i) the simplest element that we could caption is a \textit{patch}, the atomic element of an image representation in modern architectures based on vision transformers~\citep{dosovitskiy2021an}, and ii) we can easily aggregate multiple patch representations from frozen vision-language backbones to produce descriptions for arbitrarily large --- and also potentially not contiguous --- image regions.

We present a regional captioning framework which implements these ideas through a zero-shot regional captioning setup: we build on the assumptions that paired text-region data is not available at training time, and the models can be solely trained on text data, assumed to be available in large quantities. %
Our formulation offers maximum flexibility in zero-shot regional captioning tasks, producing models that effortlessly generate captions for various aggregations of image patches, ranging from individual patches to larger image regions, up to providing a caption for the entire image.

Despite the powerful perspective change that defines the patch as the new captioning unit, the problem is now entangled in a simple yet critical question: \textit{how can we craft a model able to provide patch-level captions without relying on any direct patch-level ground truth supervision?}

In the last years, vision-language foundation models like CLIP~\citep{radford2021learning,jia2021scaling,li2023blip} solved many downstream tasks in zero-shot or even training-free configurations. In particular, contrastively learned vision-language representations enabled impressive results in zero-shot settings in image classification~\citep{radford2021learning,zhai2022lit}, open-vocabulary detection~\citep{zhou2022detecting,minderer2023scaling}, and segmentation~\citep{ghiasi2022scaling,liang2023open}, or text-image retrieval~\citep{Kordopatis2025ilias}. %
Image captioning, however, cannot directly employ CLIP machinery at inference time to generate text, given that CLIP is inherently a discriminative --- and not a generative --- approach. Only recently, image captioning models became zero-shot by decoupling image encoding --- where pre-trained vision-language models like CLIP are used to create proper image and text representations --- from the actual generative module. This is the case for models like ~\cite{nukrai2022text,li2023decap,gu2023can,zeng2024meacap,fei2023transferable,yan2025entrocap,zeng2025mercap,tewel2022zerocap,su2022language}, which i) employ CLIP to leverage a shared vision-language semantic space, and ii) train a text decoder on solely text samples to recover the text back from the CLIP textual feature. This requires nothing more than a pre-trained contrastive model and a large set of sole text samples to craft a powerful captioner.

In this paper, we show that
many zero-shot captioners, paired with the right components, can be easily restructured to perform zero-shot \textit{region-based} captioning. 
Therefore, we identify and study in detail the most critical components of this novel zero-shot regional captioning framework. Particularly, we focus our attention on the pre-trained vision-language contrastive backbone, which should be able, unlike CLIP, to create meaningful patch representations. To this aim, we largely explore DINO-based ~\citep{caron2021emerging,DBLP:journals/tmlr/OquabDMVSKFHMEA24} variants, having better localized capabilities than CLIP. On top of this, we address multiple modality-gap mitigation strategies helping the text decoder to correctly interpret visual features without the need for paired image-text data, as well as a study on different patch aggregation methods. %

By analyzing existing components and employing vision backbones able to output patch-level meaningful representations like DINO, we show that we can enable many zero-shot captioners to reach state-of-the-art or comparable results in many zero-shot captioning task variants requiring captioning sub-parts of the entire image --- \textit{dense captioning}~\citep{johnson2016densecap}, \textit{region-set captioning}~\citep{cornia2019show}, up to the standard image captioning~\citep{tewel2022zerocap} where the region to caption extends over the entire image.
To better showcase the effectiveness of our framework in extreme patch-based captioning scenarios, we also introduce the %
\textit{trace captioning} task, requiring the captioning of an image region specified by a mouse trace.

To summarize, our contributions are the following:
\begin{enumerate*}[label=\alph*)]
\item we reformulate captioning by shifting perspective from the \textit{image-to-caption} approach to a \textit{patch-to-caption} one, unifying local and global tasks in one framework which does not require region-level supervision, through the exploitation of frozen vision-language backbones,
\item we repurpose existing models to work within this novel framework, by analyzing the role of the key components, with a special attention to the vision backbone, %
\item we show the performance of these models on four zero-shot captioning tasks spanning different region granularity, from captioning few patches to the whole image, showing the effectiveness of the proposed perspective shift proposed by our framework despite its simplicity.
\end{enumerate*}

\begin{figure*}[t]
    \centering
    \includegraphics[width=\linewidth]{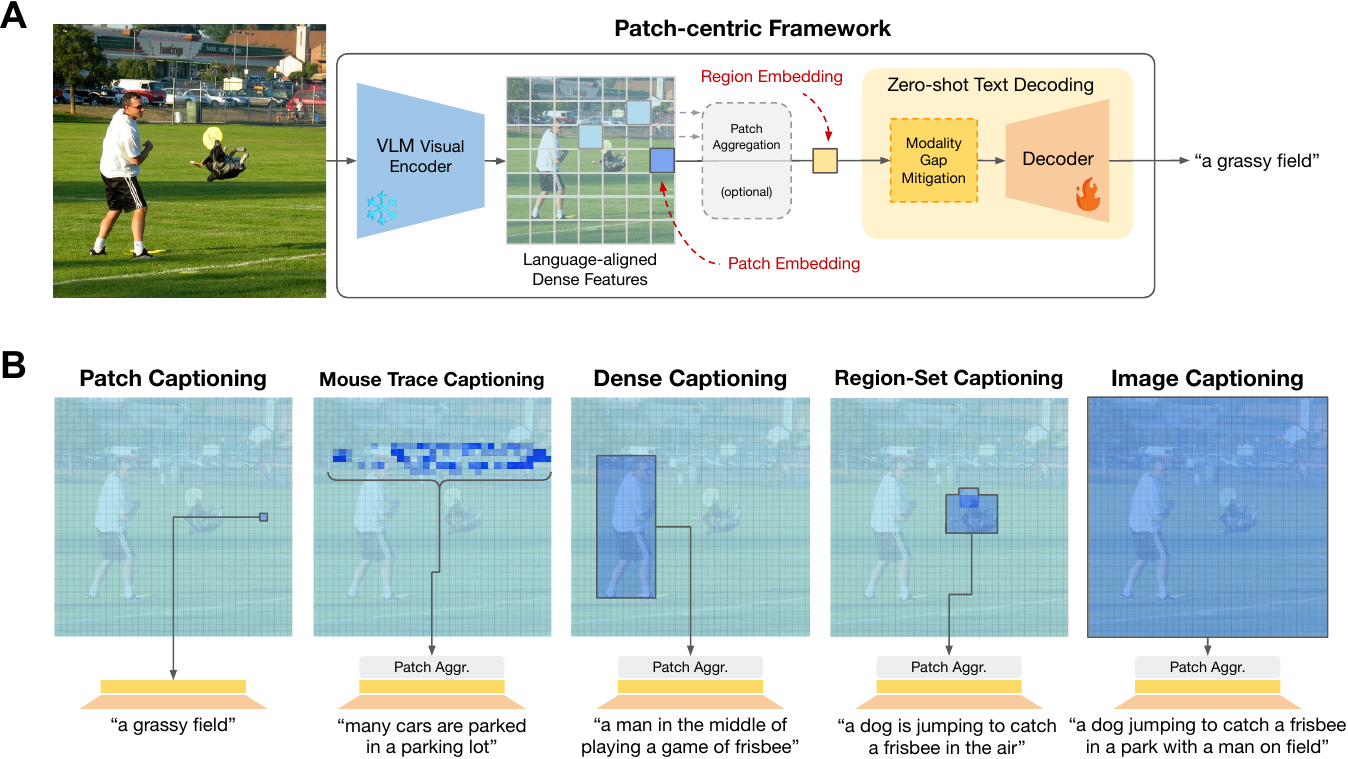}
    \caption{\textbf{Patch-centric framework for unified zero-shot captioning}. %
    \textsf{\textbf{A}}. Overview of our framework. First, we extract language-aligned dense patch embeddings from the image using a VLM. Given a region, we select the underlying patches and aggregate their features to obtain a region representation. Finally, we obtain the region caption by applying a zero-shot text decoder, that is a) conditioned on the latent region representation, b) trained on text-only data, and c) equipped with a mechanism to handle the modality gap present in vision-language common spaces. This enables regional captioning without requiring region-level supervision. %
    \textsf{\textbf{B}}. By aggregating patch-level features from arbitrary image regions, we can flexibly handle multiple captioning tasks across spatial granularities in a unique model.
    }
    \label{fig:teaser}
\end{figure*}

\section{Related Work}

\textbf{Language-aligned Dense Image Representations}
are crucial for our goal of captioning at patch level.
Vision-language models (VLM) like CLIP~\citep{radford2021learning} introduced a powerful approach to learning global modality representations in a shared space via contrastive learning, paving the way to solve several downstream tasks, including captioning~\citep{mokady2021clipcap,cornia2024generating}.  %
However, in zero-shot settings, CLIP-like representations are known to struggle with dense tasks due to misalignment between local visual patches and fine-grained semantics~\citep{zhong2022regionclip,ranasinghe2023perceptual,bica2024improving}.
On the other hand, visual-only self-supervised models (SSM) like DINO~\citep{caron2021emerging,DBLP:journals/tmlr/OquabDMVSKFHMEA24} %
excel in local semantic modeling but lack a bridge with language.
Recent works like SILC~\citep{naeem2024silc}, DINO.txt~\citep{jose2024dinov2} and SigLIP~2~\citep{tschannen2025siglip2} %
attempt to get the best of both worlds by combining DINO- and CLIP-like training objectives, aiming to obtain language-aligned dense representations.
INViTE~\citep{ICLR2024_99120406} modifies CLIP’s visual encoder by zeroing the attention weights from each patch to all the others in the last layers, leading to more semantic patch features.
DenseCLIP~\citep{rao2022denseclip} and RegionCLIP~\citep{zhong2022regionclip} extend CLIP with additional region-level supervision. RegionCLIP leverages a region-proposal network to construct region-text pairs from image-text datasets, while DenseCLIP introduces a pixel-to-text matching loss to strengthen the alignment between local regions and textual concepts.
Other methods instead exploit already existing VLMs and SSMs to get the same properties with minimal or no training: Talk2DINO~\citep{barsellotti2024talking} connects language to the DINOv2 space by mapping CLIP textual representations to DINOv2 patches. ProxyCLIP~\citep{lan2024proxyclip} instead leverages DINO's attention maps to improve the local properties of the CLIP visual embeddings of patches.

\textbf{Zero-shot Image Captioning}
methods rely mostly on global CLIP representations to guide text generation. \textit{Early-guided} decoding methods take CLIP visual features as input and introduce adaptation techniques to reduce the visual-textual modality gap~\citep{liang2022mind}.
DeCap~\citep{li2023decap} projects CLIP visual features into a more text-aligned space using a memory of texts as basis, while CapDec~\citep{nukrai2022text} and CLOSE~\citep{gu2023can} inject noise during text-only training to enable decoding also from the CLIP visual space.
Diffusion Bridge~\citep{lee2025diffusionbridge} further mitigates this gap by applying a diffusion model to refine visual features toward the textual manifold.
To improve the generation ViECap~\citep{fei2023transferable}, MeaCap~\citep{zeng2024meacap}, MERCap~\citep{zeng2025mercap} and EntroCap~\citep{yan2025entrocap} leverage \textit{external knowledge} to condition the decoding together with the CLIP image representation.
\textit{Late-guided} decoding methods instead use CLIP as a scoring or optimization signal rather than direct input. ZeroCap~\citep{tewel2022zerocap} leverages CLIP gradients to steer the cached context during text generation, while MAGIC~\citep{su2022language} optimizes token selection based on CLIP similarity scores.
However, all the above approaches rely on global representations, which are not well-suited for capturing localized semantic details, making them suboptimal for patch-level or region-level captioning in zero-shot settings.

\textbf{Region-level Captioning} comprises several tasks in which models are asked to produce natural language descriptions based on sub-parts of an image. They pose additional challenges as naively captioning the cropped regions or feature maps often induces a loss of the global context of the image and, thus, misinterpretation of the region.
For this reason, zero-shot solutions to this family of problems are still underexplored.
For \textit{controllable captioning}~\citep{cornia2019show} ---  the generation of an image caption controlled by a set or sequence of regions --- and \textit{dense captioning}~\citep{johnson2016densecap} --- the localization and captioning of salient regions of an image --- state-of-the-art solutions like CAG-Net~\citep{yin2019context}, GRiT~\citep{wu2024grit}, ControlCap~\citep{zhao2024controlcap}, and FlexCap~\citep{dwibedi2025flexcap} provide good performance but need supervision with ground-truth boxes.
Recent works \citep{guo2024regiongpt}, \citep{hua2025finecaption} moved towards the direction of arbitrary regions captioning exploiting region-level supervision, and 
the Localized Narratives dataset~\cite{pont2020connecting} --- comprising images, timed captions, and timed mouse tracks --- provide the ingredients for evaluating captioning also at track- or patch-level.
We propose a unique framework to tackle captioning at various granularities, from image- to patch-level, in a \textit{zero-shot setting}.

\section{A Patch-centric Framework}
\label{sec:method}

\paragraph{Overview.}
Figure \ref{fig:teaser} %
provides an overview of our framework, which reformulates captioning around image patches as the fundamental visual units. An input image is first encoded by a frozen vision–language backbone into dense, language-aligned patch embeddings. A region is then represented by aggregating the embeddings of its constituent patches, and a text decoder generates a natural-language caption from this aggregated representation.
\paragraph{Assumptions.}
Our approach operates under three main assumptions: (i) no region–text supervision is available during training, meaning the model never observes captions paired with localized image regions, (ii) the vision–language backbone remains frozen, providing patch-level representations that reside in, or can be projected into, a joint embedding space shared with the text encoder, and (iii) no image-text supervision is available to train the text decoder, which instead can be trained exclusively on text data, learning to reconstruct captions from their textual embeddings.

At inference, the decoder interprets visual features through the shared multimodal space, enabling region-level captioning without any region–text or image-text annotations.  
The following subsections detail how each component is instantiated, how regions are represented through a parameter-free patch aggregation, and how we address the modality gap between visual and textual representations.

\subsection{Motivation and Formulation}
\label{sec:regional-captioning}

\paragraph{Learned region fusion requires regional data.}
Traditional regional captioning~\citep{johnson2016densecap, dwibedi2025flexcap, zhao2024controlcap} follows an early injection of region specification in the model. Formally, given an image $I$ and a region $R$, the region caption $t$ is modeled as $t = \mathcal{D}(I, R)$. Such models require region-caption annotations and often use dedicated models or losses per captioning task or granularity. A step forward can be moved by introducing a formulation which disentangles image encoding and postpones region specification, defining $t = \mathcal{D}(\psi(I), R)$, where $\psi(I)$ provides a visual representation of the image independent of the region $R$, and $\mathcal{D}$ performs late region selection and text decoding.
In order to avoid training the whole pipeline using region-level labels, we further decompose the decoder module $\mathcal{D}(\cdot)$ into two distinct modules: a parameter-free fixed aggregation $\operatorname{agg}_R$ of patch-level representations, and an actual text decoder $\phi(\cdot)$ not directly conditioned on regions $R$, so that $t = \phi(\operatorname{agg}_R(\psi(I), R))$. We will detail these two factorized components in the following paragraphs.

\paragraph{Parameter-free patch aggregation.}
Let $I \in \mathbb{R}^{H \times W \times 3}$ be an image split into non-overlapping patches of size $P \times P$. Each patch is encoded with a vision backbone $\psi_v$, yielding a dense grid of patch-level embeddings $V = \psi_v(I) = \{\mathbf{v}_i\} \in \mathbb{R}^{\frac{H}{P} \times \frac{W}{P} \times D}$, where $\mathbf{v}_i \in \mathbb{R}^D$.  Assuming that $\psi_v(I)$ extracts this spatial grid of patch embeddings, a region definition $R$ selects a subset of patch features that we aggregate to obtain the region embedding $\mathbf{v}_R = \operatorname{agg}_R(\psi_v(I))$. We describe this aggregation as $\mathbf{v}_S = \sum_{i \in S} w_i \mathbf{v}_i$, where $S$ is the set of indices of patches that underlie the region, and $w_i$ are aggregation weights. While several aggregation functions can be considered (e.g., uniform, gaussian, attention-based), we found that the specific choice has limited impact in regional captioning (see SM\S\ref{sec:aggr-size}) and report results with mean aggregation ($w_i = \sfrac{1}{|S|}$). Using a set operator as aggregation gives us the flexibility to aggregate arbitrary sets of patches, and thus, define regions as boxes, masks, traces, single patches, or full-image grids.
Empirically, we also find that not all $\psi_v$ are suitable to extracting patch-level meaningful visual representations; transformer architectures pretrained with dense local contrastive objectives (such as DINO) are significantly more robust at the patch level, as validated in \S\ref{sec:backbone-selection}.

\paragraph{Zero-shot decoding.}
\label{par:zero-shot-decoding}
We train a text decoder $\phi: \mathbb{R}^{D} \to \mathcal{T}$ using a prefix language modeling approach, where the decoder reconstructs a caption $t \in \mathcal{T}$ from its text embedding $\psi_\text{t}(t)$. If $\psi_\text{t}(t)$ is aligned to the visual encoder $\psi_v(I)$, we could directly decode the visual embeddings using a text-only decoder trained on text-only data. In such a case, we can finally decode the region representation into a natural language caption $t = \phi(\mathbf{v}_R)$. However, the assumption that $\phi(\cdot)$ can digest features from $\psi_v(I)$ while being trained to reconstruct features from $\psi_\text{t}(t)$ is often too optimistic due to the prominent modality gap~\citep{liang2022mind}. In fact, text and image representations, despite being in the same multimodal space, occupy different, separated subspaces. To obtain a decoder applicable to visual embeddings, we experimented with three mitigation strategies for this gap. The first strategy, following \cite{li2023decap}, introduces a projection step at inference that maps visual features into the text subspace using a memory of text embeddings. The second, inspired by \cite{gu2023can} and \cite{nukrai2022text}, trains the decoder under input perturbations, enforcing robustness to visual embeddings laying in another subspace. 
Finally, following \cite{lee2025diffusionbridge}, we explore a diffusion-based strategy that applies a diffusion model trained on the textual manifold to the patch-level representation, effectively "bridging" the gap by iteratively refining the visual features toward the text subspace.
We analyze the effects of the modality gap mitigation strategies in SM\S\ref{sec:suppl:modality-gap}.

Overall, our formulation offers three main advantages: a) it is zero-shot by design, in the sense that it only requires textual descriptions to be trained and does not require paired image-text samples to train the text decoder, b) any region (e.g., whole image, box, mask, free-form trace, single point) is addressed identically, supporting a modular, general-purpose regional captioning pipeline, and c) our method requires only a single forward pass of the vision backbone to extract patch features for an entire image, which can then be reused to caption multiple regions without rerunning the full pipeline for each one.

\subsection{From Patches to Regions}
\label{sec:region-captioning}

Building on patch-level zero-shot captioning, our framework can generate captions for arbitrary regions of an image.
A region is defined as a set of patches, and its representation is obtained by averaging the embeddings of its constituent patches.
This simple formulation unifies several existing region-level captioning tasks, which differ only in how the relevant patches are selected, allowing us to address them without task-specific modifications.
In the following, we describe the tasks considered in our evaluation and their induced patch selections.

\textbf{Image Captioning} 
involves generating a single caption that describes the entire image. To achieve this, we derive a global representation $\mathbf{v}_I = \avg_i(\mathbf{v}_i)$ by aggregating the feature embeddings of all patches $\{\mathbf{v}_i\}$ within the image $I$.

\textbf{Dense Captioning}
requires %
locating salient regions in an image and generating their %
descriptions. %
Following defined evaluation protocols~\citep{johnson2016densecap}, we focus on captioning already defined boxes, effectively removing the localization subtask, which can be tackled using additional region-proposal models. Given a bounding box $B$ %
and the set $S_B$ of indexes of patches that intersect with $B$,
we obtain the representation of the region $\mathbf{v}_B = \avg_{i \in S_B}(\mathbf{v}_i)$. %

\textbf{Region-set Captioning} consists of generating a single caption for multiple regions within an image, where each region is specified by a distinct bounding box.
Given an image $I$ and a set of bounding boxes $\mathfrak{B} = \{B_1, B_2, \dots, B_K\}$, %
we define $S_{B_k}$ as the set of patches that intersect with the $k$-th bounding box in $\mathfrak{B}$.
To represent the entire set of regions, we aggregate the feature embeddings from all selected patches across all bounding boxes, which results in a combined region-level representation%
\begin{equation}\label{eq:region}
\mathbf{v}_\mathfrak{B} 
= \frac{1}{\sum_{B \in \mathfrak{B}} |S_B|}
   \sum_{B \in \mathfrak{B}} \; \sum_{i \in S_B} \mathbf{v}_i\,.
\end{equation}
Note that if a patch appears in more than one box, it is weighted more in the average.

\paragraph{Trace Captioning.}  
To demonstrate the flexibility of our approach, we introduce \textit{Trace Captioning}, a novel task in which the region of interest is specified by a mouse trace $T = \{p_1, \dots, p_L\}$ with $L$ points.
Each point $p_j$ is mapped to the corresponding patch index $i_j$, yielding the sequence $S_T = [i_1, \dots, i_L]$.
The trace-level representation is then obtained by averaging the embeddings of the selected patches
\begin{equation}\label{eq:trace-agg}
\mathbf{v}_T = \frac{1}{L} \sum_{j=1}^L \mathbf{v}_{i_j}\,.
\end{equation}
Unlike box-based settings, this formulation allows for free-form, user-specified regions and thus enables interactive and fine-grained localized descriptions, expanding the scope of region-level captioning.

\section{Experiments}
\label{sec:experiments}

In this section, we first quantitatively assess the role of the backbone in our novel framework (\S\ref{sec:backbone-selection}).
Then, we measure the performance of our new formulation over state-of-the-art zero-shot captioning methods, evaluating them %
region-specific zero-shot tasks (\S\ref{sec:sota}).
Other interesting yet less influential studies --- like the role of the modality-gap mitigation strategy and the patch aggregation operator $\operatorname{agg}_R(\cdot)$ --- are available in SM \S\ref{sec:aggr-size} and \S\ref{sec:suppl:modality-gap}.

To create a common ground, we first introduce the employed datasets and metrics in the following section.

\subsection{Datasets and Metrics}

\paragraph{Trace Captioning.}
We build a benchmark for Trace Captioning exploiting Localized Narratives~\citep{pont2020connectingLocalizedNarratives} --- a dataset in which
annotators vocally described objects in images while moving the mouse pointer over the described object.
The dataset provides temporal annotated voice transcriptions and mouse traces for the images of many standard captioning datasets. %
We took the labeled COCO \citep{lin2014microsoft, chen2015microsoft} with splits defined in \cite{karpathy2015deep} and Flickr30K%
\citep{young2014flickr30k} test subsets to build the trace captioning evaluation datasets. 
We split long traces and transcriptions for each image into sentences, and we discard the parts of the traces that are not temporally located between the start and the end of each sentence.
We discarded noisy sentences --- such as the ones describing image properties (\textit{The image is blurred}, \textit{the image is edited}, ...) and rewrote each sentence removing uncertainties typical of voice descriptions in a more concise and caption-like style through a few-shots prompted LLM (LLama 3 \cite{dubey2024llama}).
After annotation cleaning, 51 COCO images resulted without clean sentences and were discarded.
Each sub-trace and relative sentence comprise an independent sample, that is, we ignore the temporal information of consecutive sentences and focus on evaluating the description of each sub-trace only.
The final benchmark includes 4,949 COCO images with 14,283 captions (average caption length: 7.8 words) and 1,000 Flickr30K images with 26,614 captions (average caption length: 6.7 words).
Samples and additional details are available in \S\ref{sec:trace-details}.

\paragraph{Dense Captioning.}
We assess the performance on dense captioning tasks following the evaluation procedure of \cite{johnson2016densecap}, omitting the bounding box proposal and evaluating only the bounding box captioning task, using ground-truth boxes as input for the models.
In addition to standard caption metrics, for this task, we also report the mAP as originally defined by \cite{johnson2016densecap}.
We use the Visual Genome (VG) v1.2 ~\citep{johnson2016densecap,krishna2017visual} and VG-COCO test splits~\citep{li2019learning}. %
The former comprises 5000 images from VG, while the latter contains 2476
images present in both VG and COCO. %
Both contain multiple bounding box annotations per image with descriptions.

\paragraph{Region-Set Captioning}
We follow the evaluation protocol of \cite{cornia2019show} that originally introduced region-set captioning.
We use the Flickr30K Entities~\citep{plummer2015flickr30kentities} and the COCO Entities~\citep{cornia2019show} datasets.
Each record comprises an image, a set of bounding boxes of variable length, and a ground-truth controlled caption.
We evaluate on the test splits, comprising images from the Karpathy \citep{karpathy2015deep} splits, which consist of 3569 and 1000 images for COCO and Flickr30k versions, respectively.

\paragraph{Image Captioning}
We follow the standard evaluation for zero-shot captioners, generating captions for the 5000 images in Karpathy's COCO test split. %
We compare with DeCap \citep{li2023decap}, CLOSE \citep{gu2023can}, ZeroCap \citep{tewel2022zerocap}, MAGIC \citep{su2022language}, ViECap \citep{fei2023transferable}, CapDec \citep{nukrai2022text}, EntroCap \citep{yan2025entrocap}, MeaCap\footnote{We picked the MeaCap$_{InvLM}$, which uses a GPT-2 decoder and achieves the highest scores in the zero-shot captioning task.}~\citep{zeng2024meacap}, MERCap \citep{zeng2025mercap}, IFCap~\citep{lee2024ifcap}, and Diffusion Bridge~\citep{lee2025diffusionbridge}.

\paragraph{Metrics.}

All datasets used in our evaluation provide a ground-truth caption for each annotation.
We therefore adopt standard captioning metrics to assess the similarity between generated and reference captions.
In particular, we focus on CIDEr (C)~\citep{vedantam2015cider}, which captures syntactic overlap, and RefPAC Score (P)~\citep{sarto2023pac}, a more recent metric that quantifies semantic similarity independently of caption phrasing.
For completeness, results with other traditionally used metrics --- BLEU@4~\citep{papineni2002bleu}, METEOR~\citep{banerjee2005meteor}, ROUGE-L~\citep{lin-2004-rouge}, and SPICE~\citep{anderson2016spice} --- are reported in supplementary materials, as they follow the same trends. For the image captioning task, we additionally report CLIP-Score~\citep{hessel2021clipscore}, which measures the alignment between an image and its generated caption in the joint CLIP space. This metric is not applicable to region-set, trace, or dense captioning tasks, where captions describe local regions rather than the entire image.

\subsection{Backbone Selection}
\label{sec:backbone-selection}
The choice of the visual backbone is crucial for our patch-centric framework, as the quality and semantic richness of the patch features directly impact the captioning performance.
We tested several state-of-the-art vision-language models pre-trained \emph{without region-level supervision} and evaluated their effectiveness within the framework.
We tested vanilla \textbf{CLIP}~\citep{radford2021learning}, three CLIP adaptations for dense tasks --- \textbf{DenseCLIP}~\citep{rao2022denseclip}, \textbf{INViTE}~\citep{ICLR2024_99120406}, and \textbf{ProxyCLIP}~\citep{lan2024proxyclip} ---, \textbf{SigLIP2}~\cite{tschannen2025siglip2}, and two methods with visual encoders based on DINOv2~\citep{oquab2023dinov2} --- 
\textbf{DINO.txt}~\citep{jose2025dinov2} and
\textbf{Talk2DINO}~\citep{barsellotti2024talking}. In \S\ref{sec:backbone-details} we provide further details on each backbone adopted.

Patches are aggregated as described in \S\ref{sec:region-captioning}, while for the zero-shot decoder, we align with the setting of \cite{li2023decap}, and use a prefix GPT-2 style decoder (SM\S\ref{sec:implementation-details} reports implementation details), with a memory-based latent projection approach as mitigation strategy for handling the modality gap.
Specifically, before decoding, the region representation $\mathbf{v}$ is projected into the text embedding space as a similarity-weighted linear combination of memory elements, $\mathbf{v}_\text{proj} = M\,\alpha$ with $\alpha = \mathrm{softmax}\!\left(\tfrac{1}{\tau} M^\top \mathbf{v}\right)$, where $M = [\mathbf{m}_1, \dots, \mathbf{m}_N]$ stores the text embeddings $\mathbf{m}_j = \psi_t(t_j)$ and $\tau > 0$ controls the sharpness of the weighting distribution.
This choice enables us to be directly comparable with other works using the same architecture in the all the following experiments, besides also performing the best among the tested zero-shot decoder methods (see SM\S\ref{sec:suppl:modality-gap}).
In SM\S\ref{sec:aggr-size}, we also study how the choice of the memory bank $M$ affects the captioning performance, providing an upper bound to metrics.

\begin{table*}[t]
\centering
\resizebox{\textwidth}{!}{
\begin{tabular}{lllccccccccc}
    \toprule
    \multicolumn{3}{l}{\makecell[l]{Captioning Task: \\ (Dataset)}} & \multicolumn{2}{c}{\makecell{Trace \\ (COCO)}} & \multicolumn{2}{c}{\makecell{Dense \\ (VG v1.2)}} & \multicolumn{2}{c}{\makecell{Region-Set \\ (COCO Entities)}} & \multicolumn{3}{c}{\makecell{Image \\(COCO)}} \\
    \cmidrule(lr){1-3}\cmidrule(lr){4-5}\cmidrule(lr){6-7}\cmidrule(lr){8-9}\cmidrule(lr){10-12}
     & Vision Backbone & Text Backbone & C & P & C & P & C & P & C & P & CLIP-S \\
    \midrule
    CLIP & CLIP B/16 & CLIP B/16 & 10.9 & 75.0 & 10.9 & 74.2 & 41.6 & 78.8 & 42.1 & 84.0 & 66.2 \\
DenseCLIP & CLIP B/16 & CLIP B/16 & 18.6 & 75.3 & 19.9 & 75.2 & 51.0 & 77.6 & 28.0 & 77.0 & 57.3 \\
INViTE & CLIP B/16 & CLIP B/16 & 13.8 & 76.4 & 16.8 & 77.3 & 43.3 & 78.9 & 21.3 & 79.1 & 60.6 \\
ProxyCLIP & DINO B/8 + CLIP B/16 & CLIP B/16 & 16.7 & 75.7 & 15.7 & 76.0 & 41.2 & 78.4 & 28.7 & 79.0 & 61.7 \\
ProxyCLIP & DINOv2 B/14 + CLIP B/16 & CLIP B/16 & 16.5 & 75.7 & 15.5 & 76.0 & 40.6 & 78.5 & 27.4 & 78.6 & 61.0 \\
SigLIP2 & SigLIP2 B/16 & SigLIP2 B/16 & 18.3 & 73.6 & 19.8 & 73.4 & 47.2 & 76.7 & 27.7 & 77.0 & 56.4 \\
DINO.txt & DINOv2 B/14 & DINO.txt & \underline{23.2} & \textbf{78.8} & \underline{23.4} & \underline{78.5} & \underline{91.8} & \underline{86.3} & \underline{67.8} & \underline{87.2} & \underline{70.8} \\
Talk2DINO & DINOv2 B/14 & Talk2DINO & \textbf{27.9} & \underline{78.7} & \textbf{31.9} & \textbf{78.8} & \textbf{109.1} & \textbf{87.5} & \textbf{69.2} & \textbf{87.4} & \textbf{72.8} \\
\bottomrule
\end{tabular}
}
\caption{\textbf{Vision-Language Backbones.} CIDEr (C) and RefPAC-S (P) across four captioning tasks.
}
\label{tab:backbones-over-tasks}
\end{table*}

\autoref{tab:backbones-over-tasks} shows that backbone effectiveness in our framework is closely tied to capturing fine-grained local semantics. Standard CLIP performs poorly, indicating that its patch tokens lack the spatial detail needed for our tasks~\citep{mukhoti2023open,ranasinghe2023perceptual,bica2024improving}.
Backbones that strengthen CLIP local representations, such as INViTE and DenseCLIP, achieve stronger results, supporting our hypothesis.
The best performance comes from DINOv2-based models, including DINO.txt and Talk2DINO, with the latter emerging as the most effective encoder. This underscores the importance of semantically rich patch-level features for high-quality region-level captions.
For this reason, we show results using Talk2DINO as the default backbone in subsequent experiments.

\subsection{Comparison with SOTA}
\label{sec:sota}

Despite significant advances in zero-shot image captioning, we are unaware of any prior methods specifically tailored for zero-shot regional captioning tasks. Existing zero-shot captioners are usually evaluated only at the level of whole images, without any mechanisms to natively attend to arbitrary regions.
To rigorously quantify the benefit of our approach, in \autoref{tab:results_consolidated}, we compare against both state-of-the-art zero-shot image captioners and adapted baselines:
(i) \textit{whole-image zero-shot captioners} in their standard setting or applied to region crops, simulating regional captioning by isolating local content, and
(ii) \textit{region-supervised encoders}, thus outside our no-region-label setting, that leverage mask-based (AlphaCLIP, \cite{sun2024alphaclip}) or crop-based (RegionCLIP, \cite{zhong2022regionclip}) attention coupled with the same zero-shot decoder, allowing them to attend to specific regions.
This design ensures that our evaluation covers the strongest available baselines for both image-level and region-level zero-shot captioning.
While we exclude models trained on region-level data \cite{guo2024regiongpt,hua2025finecaption}, in \S\ref{sec:suppl:comparison-lmms} we compare our framework against representative LMMs. Despite their massive end-to-end pre-training on image-text pairs, our text-only method achieves superior performance on fine-grained regional tasks.

In addition to our strongest model (i.e., Talk2DINO with the memory-based mitigation strategy), we also report other combinations that express existing zero-shot captioning approaches (CLOSE~\citep{gu2023can}, CapDec~\citep{nukrai2022text}, VieCap~\citep{fei2023transferable}, MeaCap~\citep{zeng2024meacap} and Diffusion Bridge~\citep{lee2025diffusionbridge}) but replacing the original CLIP backbone.
Specifically, CLOSE and CapDec can be expressed by choosing the noise-injection mitigation strategy and using the standard decoder pipeline described in \S\ref{sec:backbone-selection}.
The same applies to ViECap and MeaCap, although with the addition of extra knowledge in the text decoding step:
ViECap uses external entity-aware prompts, while MeaCap leverages also structured concept retrieval from a knowledge base.
Finally, Diffusion Bridge is implemented by applying a diffusion model trained on the textual manifold to the patch-level representation.

We use two datasets for each task --- a COCO-derived dataset and an additional dataset such as Visual Genome~\citep{krishna2017visual} or Flickr30k~\citep{young2014flickr30k}. %
Figure~\ref{fig:examples} shows qualitative results.

\paragraph{Patch-centric captioning excels in local, fine-grained tasks.}
In trace and dense captioning tasks, which emphasize local visual content, our patch-centric framework significantly outperforms all baselines across metrics.
In trace captioning (\autoref{tab:results_consolidated}, 1st group), our formulation outperforms image-based captioners and crop-based adaptations.
Models relying on global CLS representations fail to capture the precise objects and attributes under the trace.
Even AlphaCLIP, a region-supervised backbone applicable to traces, lags behind, underlying intrinsic limitations of the pretrained CLIP backbone. 
Dense captioning shows a similar trend (\autoref{tab:results_consolidated}, 2nd group), with our models outperforming baselines.
For this task, we report crop-based adaptations for DeCap, ViECap, MeaCap, IFCap, and Diffusion Bridge, as they provide a stronger baseline with respect to the same models applied to the global CLS (see \autoref{tab:fullTableDenseCaptvg12} in SM).
Although isolating regional content, these models discard broader contextual cues that are crucial for coherent dense descriptions.

\paragraph{Patch aggregation extends seamlessy to context-aware captioning.}
Also in region-set captioning (\autoref{tab:results_consolidated}, 3rd group), our models achieves state-of-the-art results, outperforming both zero-shot baselines and region-supervised models.
Region-set captioning tends to align more closely with image-level captioning rather than strictly focusing on localized regions (see \autoref{fig:examples} and \autoref{fig:more-examples} in SM), as regions are intended to control an image-level caption\footnote{This is expected since the ground-truth captions in the COCO Entities dataset originate from the image-level annotations of COCO %
\cite{cornia2019show}.}.
Thus, global models also tend to perform well on those tasks, showing a narrower gap with respect to ours.
By aggregating patch embeddings from arbitrary and possibly disjoint sets of regions, the model produces coherent and contextually rich captions that align with the collective semantics of the chosen areas.
In contrast, whole-image methods cannot naturally incorporate regional cues, limiting their effectiveness in this setting.
Importantly, our patch-based aggregation even surpasses AlphaCLIP, which relies on explicit mask supervision. %

\paragraph{Patch-centric models deliver comparable performance on whole-image captioning.}
In whole-image captioning (\autoref{tab:results_consolidated}, 4th group), our results remain competitive with the strongest zero-shot captioners but are slightly behind \textit{dedicated} image-centric architectures such as MERCap~\citep{zeng2025mercap} and EntroCap~\citep{yan2025entrocap}.
For this task, we report the performance of our models using an attention-based weighting, when helpful, as it usually performs marginally better than the standard average patch aggregation and thus represents the best available model for this task and reaches the smaller gap with state-of-the-art models (see SM\S\ref{sec:aggr-size}).
Notably, adding structured external knowledge or filtered retrieval (as in ViECap or MeaCap) on noise-based decoders improves fluency and informativeness, suggesting such modules are complementary for strong regional semantics. However, they compare similarly to the model using the memory-based decoder.

\newcommand{\m}[1]{\multicolumn{1}{c}{\large #1}}

\begin{table*}[t]
\definecolor{lightergray}{gray}{0.95}
\newcolumntype{g}{>{\columncolor{lightergray}}c}
\setlength{\tabcolsep}{3pt}
\centering
\resizebox{\textwidth}{!}{
\begin{tabular}{lggccgggcccggccgggccc}
\toprule
& \multicolumn{4}{c}{Trace Captioning} & \multicolumn{6}{c}{Dense Captioning} & \multicolumn{4}{c}{Region-Set Captioning} & \multicolumn{6}{c}{Image Captioning} \\
\cmidrule(lr){2-5} \cmidrule(lr){6-11} \cmidrule(lr){12-15} \cmidrule(lr){16-21}
& \multicolumn{2}{c}{COCO} & \multicolumn{2}{c}{Flickr30k} & \multicolumn{3}{c}{VG v1.2} & \multicolumn{3}{c}{VG-COCO} & \multicolumn{2}{c}{COCO Entities} & \multicolumn{2}{c}{Flickr30k Entities} & \multicolumn{3}{c}{COCO} & \multicolumn{3}{c}{Flickr30k} \\
\cmidrule(lr){2-3} \cmidrule(lr){4-5} \cmidrule(lr){6-8} \cmidrule(lr){9-11} \cmidrule(lr){12-13} \cmidrule(lr){14-15} \cmidrule(lr){16-18} \cmidrule(lr){19-21}
Model & \m{C} & \m{P} & \m{C} & \m{P} & \m{mAP} & \m{C} & \m{P} & \m{mAP} & \m{C} & \m{P} & \m{C} & \m{P} & \m{C} & \m{P} & \m{C} & \m{P} & \m{CLIP-S} & \m{C} & \m{P} & \m{CLIP-S} \\
\bottomrule
\rowcolor{lightgray}
\multicolumn{21}{l}{\textbf{\textit{Whole-image Zero-shot Captioners}}} \\ \toprule
ZeroCap \citep{tewel2022zerocap} \textsubscript{CVPR'22} & - & - & - & - & - & - & - & - & - & - & - & - & - & - & 14.6 & - & - & - & - & - \\
MAGIC \citep{su2022language} \textsubscript{ArXiv'22} & - & - & - & - & - & - & - & - & - & - & - & - & - & - & 49.3 & - & - & 17.5 & - & - \\
CLOSE \citep{gu2023can} \textsubscript{ICCV'23} & - & - & - & - & - & - & - & - & - & - & - & - & - & - & 81.2 & - & - & - & - & - \\
CapDec \citep{nukrai2022text} \textsubscript{EMNLP'22} & - & - & - & - & - & - & - & - & - & - & - & - & - & - & 91.8 & - & - & 35.7 & - & - \\
EntroCap \citep{yan2025entrocap} \textsubscript{NEUCOM'25} & - & - & - & - & - & - & - & - & - & - & - & - & - & - & \underline{94.3} & - & - & \underline{41.5} & - & - \\
MERCap \citep{zeng2025mercap} \textsubscript{AAAI'25} & - & - & - & - & - & - & - & - & - & - & - & - & - & - & \textbf{96.0} & - & - & \textbf{45.6} & - & - \\

ViECap$^\dagger$ \citep{fei2023transferable} \textsubscript{ICCV'23} & {24.3} & 74.4 & 12.0 & 68.8 & 14.90$^{\diamond}$ & 26.4$^{\diamond}$ & 74.3$^{\diamond}$ & 15.2$^{\diamond}$ & 26.6$^{\diamond}$ & 74.3$^{\diamond}$ & {102.7} & 85.0 & 31.8 & 74.9 & 89.7 & 88.5 & 75.6 & 29.8 & 80.6 & 70.5 \\
MeaCap$^\dagger$ \citep{zeng2024meacap} \textsubscript{CVPR'24} & 22.5 & 74.4 & {12.6} & 69.8 & 15.01$^{\diamond}$ & 28.6$^{\diamond}$ & 75.1$^{\diamond}$ & 16.0$^{\diamond}$ & 28.9$^{\diamond}$ & 75.1$^{\diamond}$ & 97.9 & 85.2 & 38.6 & 76.4 & 86.0 & 88.6 & 77.8 & 40.1 & 82.8 & 73.9 \\
DeCap$^\dagger$ \citep{li2023decap} \textsubscript{ICLR'23} & 20.5 & 75.3 & 11.2 & {71.0} & {17.75}$^{\diamond}$ & 24.6$^{\diamond}$ & {77.8}$^{\diamond}$ & {17.8}$^{\diamond}$ & 24.9$^{\diamond}$ & {77.7}$^{\diamond}$ & 95.1 & \underline{87.4} & 39.4 & \underline{78.8} & 87.4 & {90.6} & \textbf{79.3} & 40.0 & \underline{84.8} & \textbf{76.3} \\
IFCap$^\dagger$ \citep{lee2024ifcap} \textsubscript{EMNLP'24}  & 20.8 & 73.2 & 10.0 & 67.4 & 13.7$^{\diamond}$ & 21.3$^{\diamond}$ & 72.5$^{\diamond}$ & 13.4$^{\diamond}$ & 22.8$^{\diamond}$ & 73.0$^{\diamond}$ & 95.0 & 83.9 & 30.5 & 73.3 & 86.9 & 87.3 & 74.6 & 29.4 & 79.2 & 69.2 \\ %
Diffusion Bridge$^\dagger$ \citep{lee2025diffusionbridge} \textsubscript{CVPR'25} & 21.8 & 74.8 & 10.7 & 69.8 & 14.9$^{\diamond}$ & 20.0$^{\diamond}$ & 72.8$^{\diamond}$ & 15.2$^{\diamond}$ & 20.2$^{\diamond}$ & 72.8$^{\diamond}$ & 94.7 & 85.3 & 32.9 & 75.8 & 85.1 & 88.8 & 76.7 & 33.7 & 82.1 & 72.4 \\ %
\bottomrule
\rowcolor{lightgray}
\multicolumn{21}{l}{\textbf{\textit{With Region-level Supervision}}} \\
\toprule
RegionCLIP~\citep{zhong2022regionclip}  \textsubscript{CVPR'22} + Mem. ($\simeq$ DeCap)  & - & - & - & - & 15.85 & 21.7 & 76.7 & 16.01 & 21.0 & 75.4 & - & - & - & - & 93.4 & \textbf{91.2} & 77.5 & 38.8 & 84.5 & 73.6 \\
AlphaCLIP~\citep{sun2024alphaclip} \textsubscript{CVPR'24} + Mem. ($\simeq$ DeCap)  & 21.3 & {75.4} & 11.8 & 71.0 & 14.63 & 19.1 & 73.9 & 14.82 & 19.4 & 73.8 & 95.1 & \underline{87.4} & {39.5} & \underline{78.8} & 89.7 & \underline{91.1} & \underline{78.2} & 40.9 & \textbf{85.4} & \underline{75.0} \\
\bottomrule
\rowcolor{lightgray}
\multicolumn{21}{l}{\textbf{Patch-ioner (Our Patch-based Framework)}} \\
\toprule
\rowcolor{LightCyan} 
T2D + Mem. ($\simeq$ DeCap) & 27.9 & \underline{78.7} & 18.8 & \underline{77.0} & \textbf{21.31} & \textbf{31.9} & \underline{78.8} & \textbf{21.53} & \textbf{32.3} & \textbf{78.7} & 109.1 & \textbf{87.5} & \textbf{44.1} & \textbf{79.1} & {88.5}$^\lozenge$ & {90.2}$^\lozenge$ & {76.0}$^\lozenge$ & {39.3}$^\lozenge$ & {84.2}$^\lozenge$ & {71.8}$^\lozenge$ \\
\rowcolor{LightCyan} 
T2D + Noise ($\simeq$ CLOSE, CapDec) & \underline{29.3} & 78.1 & \underline{19.3} & 75.6 & \underline{20.26} & 26.3 & 77.0 & \underline{20.33} & 26.4 & 76.9 & 97.5 & 85.6 & 37.1 & 76.5 & 65.5\color{white}{$^\dagger$} & 86.2\color{white}{$^\dagger$} & 70.9\color{white}{$^\dagger$} & 27.8\color{white}{$^\dagger$} & 80.8\color{white}{$^\dagger$} & 67.0\color{white}{$^\dagger$} \\
\rowcolor{LightCyan} 
T2D + Noise + External knowledge ($\simeq$ ViECap) & 28.2 & 78.2 & 18.5 & 76.2 & 18.43 & \underline{30.3} & \underline{77.8} & 18.43 & 30.7 & \underline{77.7} & \underline{109.3} & 86.7 & 37.8 & 77.8 & {88.5}$^\lozenge$ & 89.2$^\lozenge$ & 73.7$^\lozenge$ & 34.1$^\lozenge$ & 82.8$^\lozenge$ & 69.9$^\lozenge$ \\
\rowcolor{LightCyan} 
T2D + Noise + Filtered knowledge ($\simeq$ MeaCap) & 27.4 & \textbf{78.8} & \textbf{20.3} & \textbf{77.3} & 18.66 & \textbf{31.9} & \textbf{78.9} & 19.43 & \textbf{32.3} & \textbf{78.7} & 104.4 & {86.9} & \underline{42.3} & {78.6} & {83.0}$^\lozenge$ & {89.6}$^\lozenge$ & {74.8}$^\lozenge$ & {39.4}$^\lozenge$ & {84.4}$^\lozenge$ & {71.4}$^\lozenge$ \\
\rowcolor{LightCyan} 
T2D + Diffusion ($\simeq$ Diffusion Bridge) & \textbf{29.4} & 77.6 & 18.7 & 75.0 & 19.01 & 30.7 & 77.1 & 19.40 & \underline{31.1} & 77.0 & \textbf{114.4} & 86.6 & 40.8 & 77.1 & 92.8$^\lozenge$ & 89.0$^\lozenge$ & 74.2$^\lozenge$ & 36.4$^\lozenge$ & 82.3$^\lozenge$ & 69.0$^\lozenge$ \\ %
\bottomrule
\multicolumn{21}{l}{$^\dagger$: reproduced by us. $^{\diamond}$: Model applied to image crops. {$^\lozenge$}: attention-based weighting. }
\end{tabular}
}
\caption{
Comparison of our Patch-ioner framework, using Talk2DINO (\textbf{T2D}), with ZS methods on trace, dense, region-set, and image captioning tasks. Our approach consistently outperforms whole-image and region-supervised baselines in local, fine-grained captioning tasks, while achieving competitive results on whole-image captioning. The table reports CIDEr (C), RefPAC (P), mean average precision (mAP) for dense captioning, and CLIP-Score (CLIP-S) when applicable; best and second-best results are in \textbf{bold} and \underline{underlined}, respectively. %
}
\label{tab:results_consolidated}
\end{table*}

\begin{figure*}

\definecolor{Ours (Talk2DINO + Mem.)}{HTML}{D64550}        %
\definecolor{DeCap}{HTML}{2E86AB}       %
\definecolor{Ours (CLIP + Mem.)}{HTML}{F39C12}     %
\definecolor{ZeroCap}{HTML}{6E2C00}     %
\definecolor{DeCap (Crop)}{HTML}{287233}     %
\definecolor{ViECap}{HTML}{5D6D7E}       %
\definecolor{GT}{HTML}{000000}       %

\newcommand{\hcap}[2]{%
  {\color{#1}\textbf{#1}: #2}%
}

    \centering
    \tiny
    \renewcommand{\methodName}{Ours (Talk2DINO + Mem.)}
    \newcommand{\figw}{.26\textwidth}
    \setlength{\tabcolsep}{1pt}
    \resizebox{\textwidth}{!}{

    \begin{tabular}{*5{p{\figw}}}
    \m{Patch} &
    \m{Trace} &
    \m{Dense} &
    \m{Region-Set} &
    \m{Image} \\
    \includegraphics[width=\linewidth,valign=b,trim={0 5.5cm 0 3cm},clip]{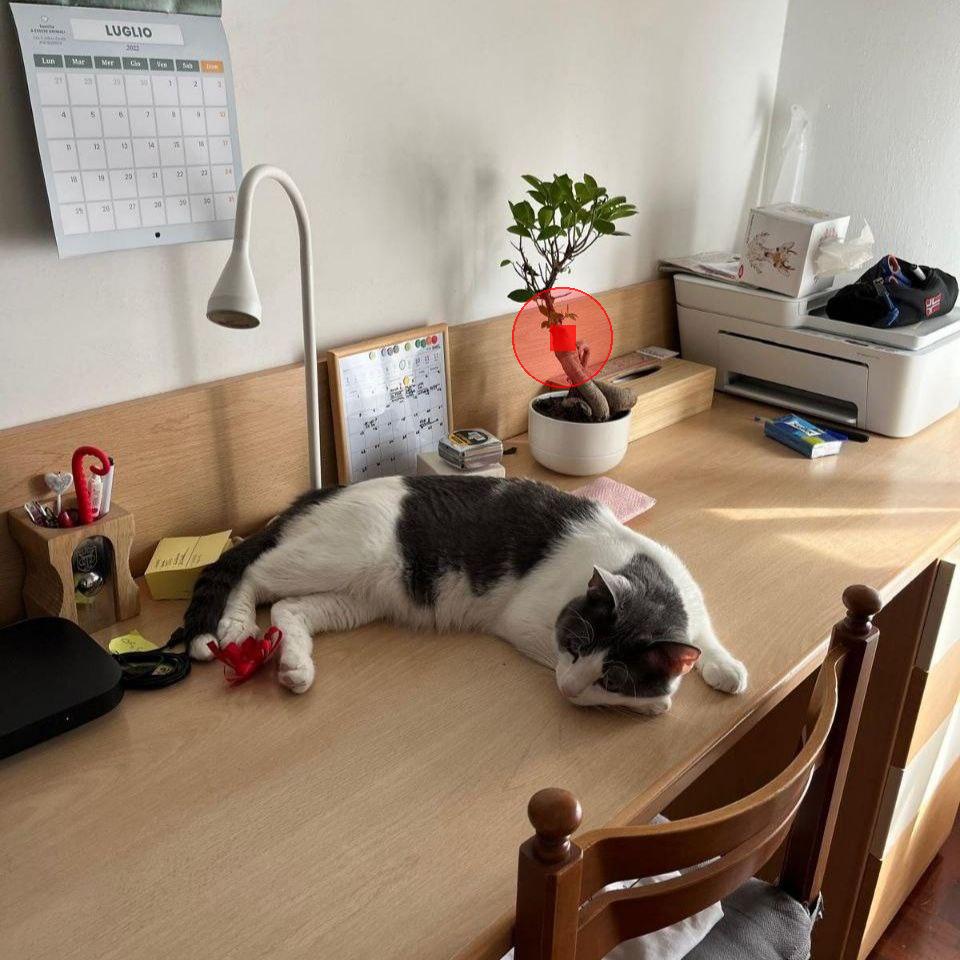} &
    \includegraphics[width=\linewidth,valign=b,trim={0 0 0 0},clip]{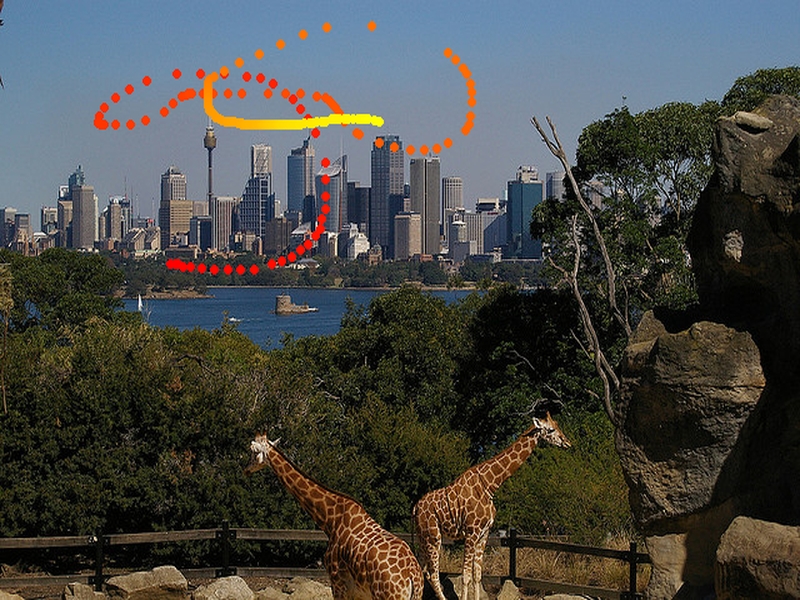} &
    \includegraphics[width=\linewidth,valign=b,trim={0 0 0 0},clip]{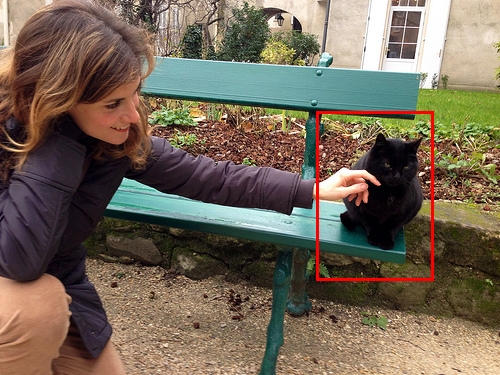} & 
    \includegraphics[width=\linewidth,valign=b,trim={2.35cm 0 0 0},clip]{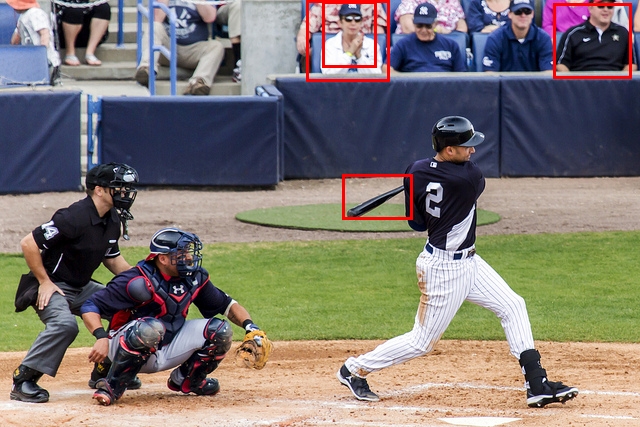}   &
    \includegraphics[width=\linewidth,valign=b,trim={0 0 0 0},clip]{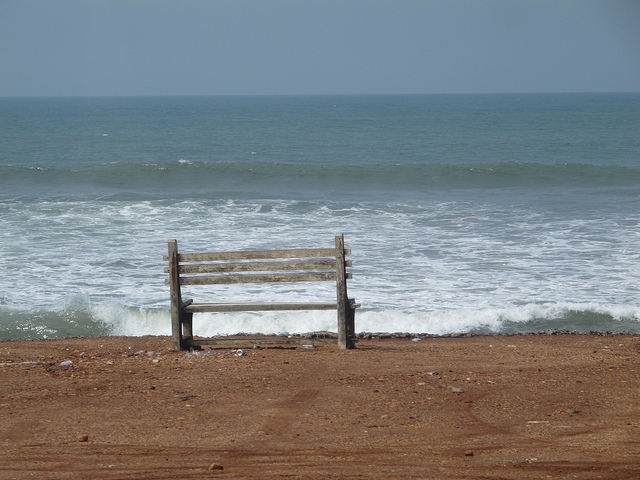} \\

    \hcap{DeCap}{a cat is sleeping on a cluttered desk.}
    \hcap{Ours (CLIP + Mem.)}{a cat is sitting on the bed and it’s contents.}
    \hcap{Ours (Talk2DINO + Mem.)}{a plant in a vase sitting on a table.} &
    
    \hcap{DeCap}{a giraffe in a zoo with a city in the background.}
    \hcap{Ours (CLIP + Mem.)}{there are some people that are out by a lot of trees.}
    \hcap{Ours (Talk2DINO + Mem.)}{a view of a city with a sky in the background.}
    \hcap{GT}{A sky.} &
    
    \hcap{DeCap}{a woman squatting on a bench with a cat.}
    \hcap{Ours (CLIP + Mem.)}{there is a person that is out on the kitchen.}
    \hcap{DeCap (Crop)}{a a close up of a person standing by a person holding a phone.}
    \hcap{Ours (Talk2DINO + Mem.)}{a black cat is sitting on a black bench.}
    \hcap{GT}{black cat sitting on a bench.} &
    
    \hcap{DeCap}{a baseball player at bat getting ready to hit the ball.}
    \hcap{Ours (CLIP + Mem.)}{some baseball players are on the field playing baseball.}
    \hcap{Ours (Talk2DINO + Mem.)}{a baseball player is swinging his bat as a crowd watches.}
    \hcap{GT}{a man swinging a baseball bat as another looks on.} &
    
    \hcap{DeCap}{a bench sits on the beach next to the ocean.}
    \hcap{ZeroCap}{a beachfront bench.}
    \hcap{ViECap}{A wooden bench sitting on top of a sandy beach.}
    \hcap{Ours (Talk2DINO + Mem.)}{a bench sitting on the beach next to the ocean.}
    \hcap{GT}{A wooden bench sitting on a beach.}\\[5ex]

\includegraphics[width=\linewidth,valign=b,trim={0 5.5cm 0 3cm},clip]{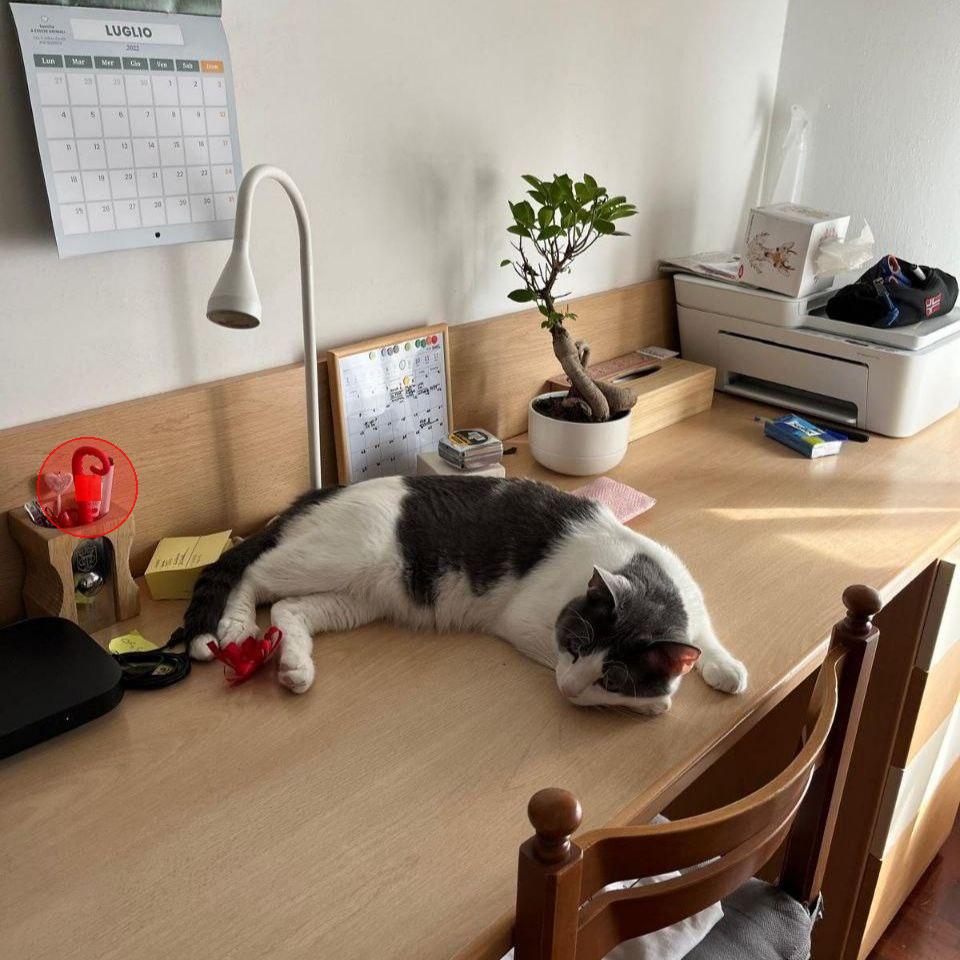} &
\includegraphics[width=\linewidth,valign=b,trim={0 0 0 0},clip]{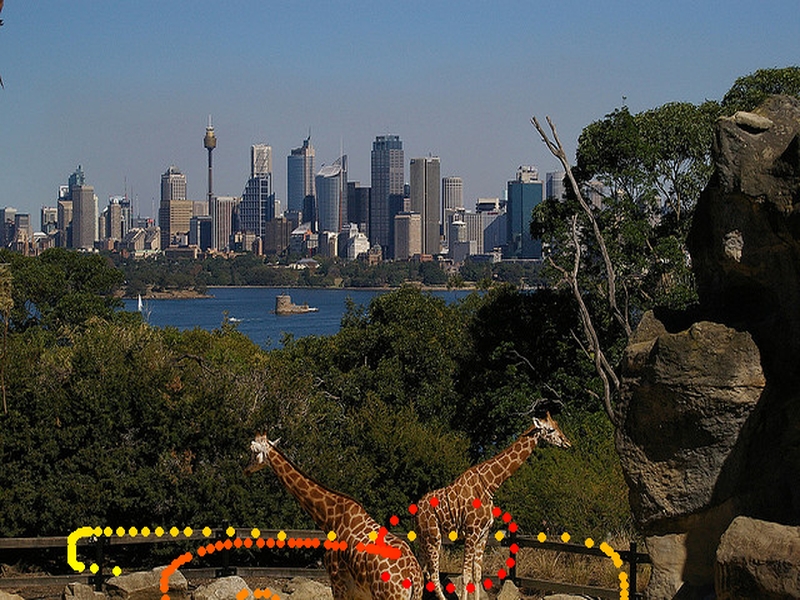} &
\includegraphics[width=\linewidth,valign=b,trim={0 0 0 0},clip]{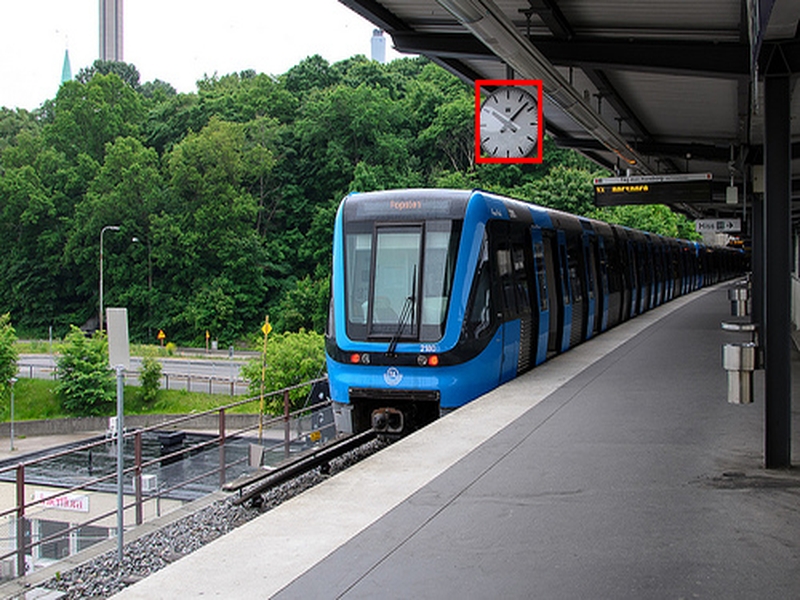} &
\includegraphics[width=\linewidth,valign=b,trim={0 0 2.5cm 0},clip]{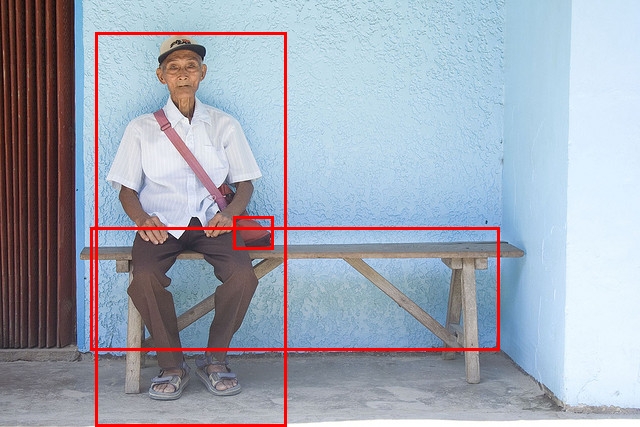} &
\includegraphics[width=\linewidth,valign=b,trim={0 0 0 0},clip]{images/qualitatives/image/example1.jpg} \\

\hcap{DeCap}{a cat is sleeping on a cluttered desk.}
\hcap{Ours (CLIP + Mem.)}{a cat is sitting at a table with a full laptop.}
\hcap{Ours (Talk2DINO + Mem.)}{office supplies, pens, toys, and other items on desk.} &

\hcap{GT}{Two giraffes, rocks, and a fence.}
\hcap{DeCap}{a giraffe in a zoo with a city in the background.}
\hcap{Ours (CLIP + Mem.)}{there are some people that are in a lot by a tree.}
\hcap{Ours (Talk2DINO + Mem.)}{two giraffes standing in a fenced area.} &

\hcap{GT}{a clock at a train station.}
\hcap{DeCap}{a train traveling along the platform of a public train.}
\hcap{Ours (CLIP + Mem.)}{a train is on the tracks and going by.}
\hcap{Ours (Talk2DINO + Mem.)}{a clock on a train station platform above a train.} &

\hcap{GT}{an old man sitting on a bench with a purse.}
\hcap{DeCap}{a man sitting on a bench while holding a door.}
\hcap{Ours (CLIP + Mem.)}{a bathroom has a blue toilet and the walls.}
\hcap{Ours (Talk2DINO + Mem.)}{a man sits on a wooden bench with a bag on his back.} &

\hcap{GT}{A black cat rubbing against a bottle of wine.}
\hcap{DeCap}{a black cat standing next to a bottle of wine glasses.}
\hcap{Ours (CLIP + Mem.)}{a cat sitting on the counter of a green bottle.}
\hcap{Ours (Talk2DINO + Mem.)}{a black cat sitting on a chair next to a bottle of wine.}
\\

    \end{tabular}
    }

    \caption{%
    \textbf{Qualitative results} from finer (left) to coarser (right) tasks.
    Note the discrepancy of predicted and ground-truth captions when an image-level (DeCap, DeCap (Crop)) or a CLIP-based regional (CLIP + Mem.) captioner is applied, with respect to Talk2DINO-based model.
    }
    \label{fig:examples}
\end{figure*}

\section{Conclusions}

We introduced a zero-shot framework that shifts from an image-centric to a patch-centric approach, enabling captioning for individual patches and arbitrary aggregations without any region or image supervision. We rely on the strong spatial awareness of DINOv2, whose local image patches have been bridged with the text modality, and we disentangle the training of the decoder network. %
This flexible and scalable method sets the new state of the art on various regional captioning tasks, including dense and region-based captioning, as well as our newly proposed trace captioning.
Results show that our patch-centric approach can effectively bridge the gap between local and global understanding in image captioning, providing a unified framework for multi-granularity captioning tasks in a zero-shot setting.
Moreover, our models require a single backbone forward pass to caption multiple regions, facilitating viability in interactive applications.

\paragraph{Limitations and Future Work.}
Despite strong zero-shot performance, our model still lags behind fully supervised, task-specific approaches. The contextual scope of each patch is fixed by the backbone and is not adjusted to meet the user's intents. Also, the modality gap introduces noise that can cause hallucinations.
Future work will focus on improving patch-level semantics, e.g., through image-level captioning loss in weakly-supervised settings, or refine the patch-to-text projection to reduce the modality gap in zero-shot settings.

{
    \small
    \section*{Acknowledgment} This work has been supported by the PNRRM4C2 project ``FAIR - Future Artificial Intelligence Research'', by the PRIN 2022-PNRR project ``MUCES'' (CUP E53D23016290001 and B53D23026090001), and by the PNRR project ``ITSERR - Italian Strengthening of Esfri RI Resilience'' (CUP B53C22001770006), all funded by the European Union - NextGenerationEU, and by the SUN XR project funded by the Horizon Europe Research \& Innovation Programme (GA n. 101092612).

    \bibliographystyle{ieeenat_fullname}
    \bibliography{main,supp}
}
\clearpage
\setcounter{page}{1}
\section*{Supplementary Material}

\newcommand{\patchesSymbol}{Patches}%
\newcommand{\gaussianWeighted}{gaussian}%
\newcommand{\attnMapWeighted}{attention}%
\newcommand{\cropCaptions}{Crop}%

\newcommand{\oneHot}{central patch}%
\newcommand{\avgPatches}{uniform}%
\newcommand{\resizeSymbol}{$\mathcal{R}$}
\newcommand{\avgTrace}{uniform}%

\section{Implementation Details}
\label{sec:implementation-details}

\begin{figure*}[h!]
    \centering
    \includegraphics[width=\linewidth]{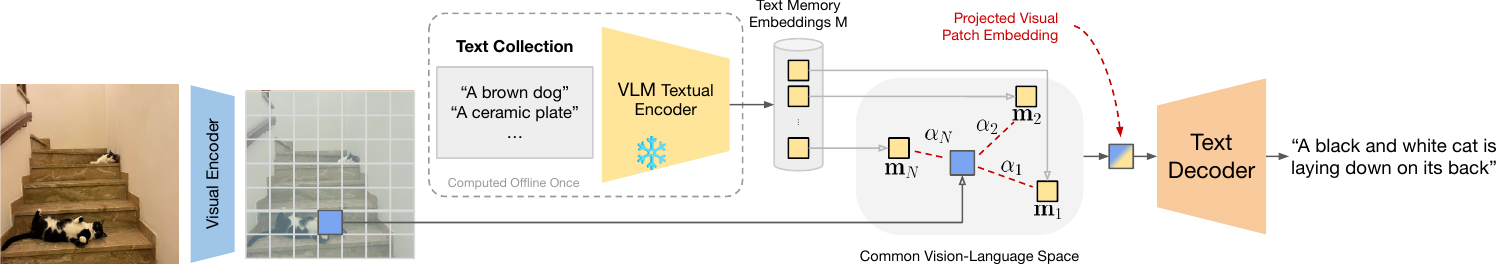}
    \caption{\textbf{Patch-level Captioning}. Given an input image, we first extract dense patch-level representations using a vision transformer backbone. For a selected patch, we apply the projection-based mechanism introduced by \cite{li2023decap} to mitigate the modality gap and align its representation with the text embedding space. Finally, the transformed embedding is fed into a text decoder trained on a text-only corpus, generating a zero-shot caption for the patch.}
    \label{fig:inference}
\end{figure*}

For the training of the textual decoder of the memory-based configuration $\phi$, we adopt a prefix GPT2-style decoder-only Transformer with 4 attention heads and 4 layers, following the architecture used by \cite{li2023decap}. We train the model on captions from the COCO training set, which also serves as the memory bank $M$ for the projection mechanism, comprising approximately 500k texts. We set the hyperparameters of the projection mechanism as in DeCap ($\tau = 0.01$), and use the AdamW optimizer with a weight decay of 0.01. Training proceeds for 10 epochs with a learning rate of $10^{-5}$
  and a batch size of 64. A comprehensive overview of the framework operating in the DeCap setting (with the projection mechanism as modality gap mitigation strategy) is provided in \autoref{fig:inference}.

For the external knowledge-based captioning models we performed a training of 15 epochs with a batch size of 80 captions on the same GPT2-style textual decoder using a learning rate of $2\times10^{-5}$ and a gaussian noise variance of $16\times10^{-3}$ to replicate the experimental settings of \cite{fei2023transferable} and \cite{zeng2024meacap}.

All experiments were conducted on a single NVIDIA H100 GPU with 80GB of HBM3 memory. Training took approximately 25 minutes per epoch.

\section{Backbone Details}
\label{sec:backbone-details}
We briefly summarize here the characteristics of the vision-language models we tested in our framework.
\begin{itemize}
    \item \textbf{CLIP}~\citep{radford2021learning}: A foundational model that learns a shared embedding space for images and text through contrastive learning. While being the most used model for global image-text alignment, its patch tokens are known to lack strong spatial and fine-grained semantic information. The input resolution of its training is 224 pixel.
    \item \textbf{DenseCLIP}~\citep{rao2022denseclip}:  A fine-tuned version of CLIP that incorporates a pixel-text matching loss to enhance the model's ability to understand local regions. The official implementation input resolution is 640 pixel for the ViT-B/16 version.
    \item \textbf{INViTE}~\citep{ICLR2024_99120406}: This method modifies CLIP's vision transformer to bring patch tokens in the text space by disabling the self-attention mechanism. It employs the same visual encoder of CLIP, trained at 224 pixel input resolution.
    \item \textbf{ProxyCLIP}~\citep{lan2024proxyclip}:  A model that leverages the local understanding of a DINO backbone to improve CLIP's patch-level representations. It achieves this by replacing the attention maps in CLIP's final layer with DINO's attention maps, effectively transferring DINO's fine-grained spatial awareness to the CLIP embedding space. The DINO ViT-B/8 version was tested with images at 296 pixel resolution, while the DINOv2 ViT-B/14 at 518 pixel. %
    \item \textbf{SigLIP2}~\cite{tschannen2025siglip2}: A multilingual vision–language model that improves upon CLIP by enhancing both global alignment and dense localization capabilities through refined training objectives and architectural adjustments. We employed the B/16 variant, which uses a ViT with 16-pixel patches and is trained at an input resolution of 512 pixels. To obtain text-aligned semantic patch representations, we follow the authors' dense feature extraction strategy and apply the MAP (Mean Attention Pooling) head---normally used to produce the global image representation embedding---individually to each patch token.
    \item \textbf{DINO.txt}~\citep{jose2025dinov2}: This model builds upon a frozen DINOv2 backbone, adding learnable transformer blocks on top. It is then trained with a contrastive objective against a text encoder to align both global and patch-level representations with language. The DINOv2 backbone was trained at the resolution of 518 pixel.
    \item \textbf{Talk2DINO}~\citep{barsellotti2024talking}: This model creates a bridge between the CLIP and DINOv2 embedding spaces. It trains a projection to map CLIP text embeddings into the DINOv2 patch space, using DINOv2's highly meaningful attention maps to identify and align with the most relevant patches during training. The DINOv2 backbone was trained at the resolution of 518 pixel.
\end{itemize}

\section{Comparison with LMMs}
\label{sec:suppl:comparison-lmms}
\begin{table*}[t]
\centering
\resizebox{\textwidth}{!}{
\begin{tabular}{lcccccccccc}
    \toprule
    \multicolumn{1}{l}{\makecell[l]{Captioning Task: \\ (Dataset)}} & Adaptation strategy & \multicolumn{2}{c}{\makecell{Trace \\ (COCO)}} & \multicolumn{2}{c}{\makecell{Dense \\ (VG v1.2)}} & \multicolumn{2}{c}{\makecell{Region-Set \\ (COCO Entities)}} & \multicolumn{3}{c}{\makecell{Image \\(COCO)}} \\
    \cmidrule(lr){1-1}\cmidrule(lr){2-2}\cmidrule(lr){3-4}\cmidrule(lr){5-6}\cmidrule(lr){7-8}\cmidrule(lr){9-11}
     & & C & P & C & P & C & P & C & P & CLIP-S \\
    \midrule
\rowcolor{lightgray}
\multicolumn{11}{l}{\textbf{LMMs}} \\
\toprule
Llava1.5 One-Vision (4B) & Visual Prompting & 21.2 & 75.9 & 25.3 & 76.2 & 72.2 & 87.3 & \textbf{91.6} & \textbf{91.1} & 80.1 \\
Qwen2.5 VL (3B) & Visual Prompting & 17.4  & 74.0  & 19.1  & 74.2  & 59.6  & 84.5 & 77.9  & 89.9  & 81.2  \\
Qwen3 VL (4B) & Visual Prompting & 9.1  & 70.7  & 10.8  & 74.2  & 10.8  & 74.2 & 19.4 & 84.2  & \textbf{85.3}  \\
\hdashline
Llava1.5 One-Vision (4B) & Crop & 19.7 & 75.4 & 15.1 & 72.7 &  75.9 & 86.7 & \textbf{91.6} & \textbf{91.1} & 80.1 \\
Qwen2.5 VL (3B) & Crop & 18.4  & 73.9  & 11.9  & 69.2  & 68.6  & 85.9  & 77.9  & 89.9  & 81.2  \\
Qwen3 VL (4B) & Crop & 7.6  & 69.3  & 3.5  & 68.8  & 20.2  & 80.4 & 19.4 & 84.2  & \textbf{85.3}  \\
\bottomrule
\rowcolor{lightgray}
\multicolumn{11}{l}{\textbf{Patch-ioner (Our Patch-based Framework)}} \\
\toprule
\rowcolor{LightCyan}
T2D + Mem. (0.21B) & & \textbf{27.9} & \textbf{78.7} & \textbf{31.9} & \textbf{78.8} & \textbf{109.1} & \textbf{87.5} & 69.2 & 87.4 & 72.8 \\
\bottomrule
\end{tabular}
}
\caption{\textbf{Patch-ioner framework vs. LMMs.}
}
\label{tab:patchioner_vs_vllms}
\end{table*}
Our framework is designed as a zero-shot regional captioner, relying solely on text-only corpora to train the decoder, and operating without any paired image-text or region-text supervision. 
While most of the Large Multi-modal Models (LMMs) are trained without region-text supervision, they are trained on massive datasets using image-text pairs, making them inherently supervised solutions.

While this difference in training paradigm places LMMs outside the core assumptions of our zero-shot setup, we include a comparison with representative, high-performing LMMs --- Llava1.5 OneVision (4B)~\cite{an2025llava}, Qwen2.5 VL ~\cite{bai2025qwen2} (3B) and Qwen3 VL (4B) --- to provide a strong, external supervised benchmark for our framework's capabilities.
\subsection{LMM Adaptation for Regional Tasks}
\begin{figure*}[h]
\centering
\resizebox{\textwidth}{!}{
\begin{tabular}{ccc}
    \toprule
    & \textbf{Original Image} & \\
    \textbf{Trace Captioning} & \textbf{Dense Captioning} & \textbf{Region-Set Captioning} \\
    \midrule
    \includegraphics[width=0.33\textwidth, height=0.25\textwidth]{images/supplementary/vllm_adaptation/trace_original.jpg} & 
    \includegraphics[width=0.33\textwidth, height=0.25\textwidth]{images/supplementary/vllm_adaptation/dense_original.jpg} & 
    \includegraphics[width=0.33\textwidth, height=0.25\textwidth]{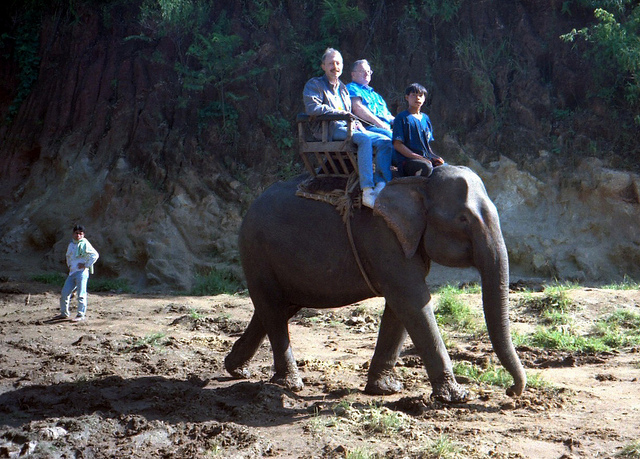} \\
    \midrule
    \multicolumn{3}{c}{\textbf{Visual Prompting Adaptation}} \\
    \midrule
    \includegraphics[width=0.33\textwidth, height=0.25\textwidth]{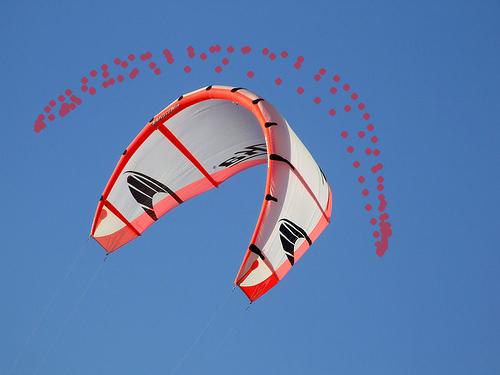} & 
    \includegraphics[width=0.33\textwidth, height=0.25\textwidth]{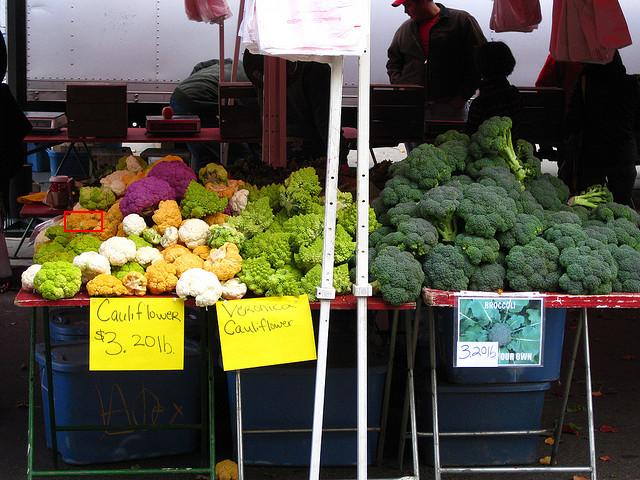} & 
    \includegraphics[width=0.33\textwidth, height=0.25\textwidth]{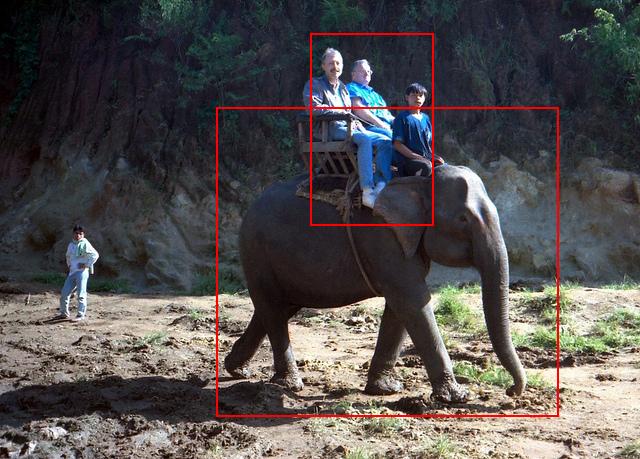} \\
     \makecell[l]{\scriptsize \textit{Provide a one-sentence caption for the provided region of} \\
     \scriptsize \textit{the image marked by the red dots. Do not mention the}\\
     \scriptsize \textit{red dots or the highlighting.}}
& 
     \makecell[l]{\scriptsize \textit{Provide a one-sentence caption for the provided region of }\\
     \scriptsize \textit{the image marked by the red bounding box. Do not mention}\\
     \scriptsize \textit{the red bounding box or the highlighting.}}
& 
     \makecell[l]{\scriptsize \textit{Provide a one-sentence caption for the provided regions of} \\
     \scriptsize \textit{the image marked by the red bounding boxes. Do not mention}\\
     \scriptsize \textit{the red bounding boxes or the highlighting.}}\\
    \midrule
    \multicolumn{3}{c}{\textbf{Cropping Adaptation}} \\
    \midrule
    \includegraphics[width=0.33\textwidth, height=0.25\textwidth]{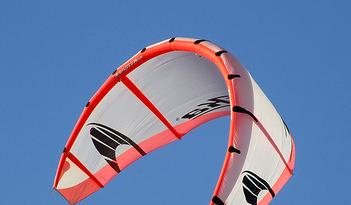} & 
    \includegraphics[width=0.33\textwidth, height=0.25\textwidth]{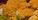} & 
    \includegraphics[width=0.33\textwidth, height=0.25\textwidth]{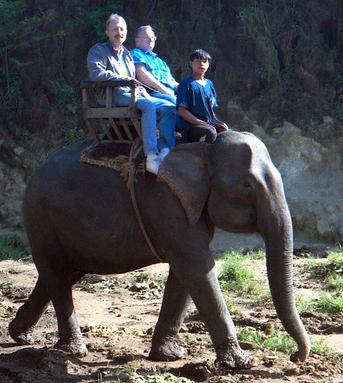} \\
    \scriptsize \textit{Provide a one-sentence caption for the provided image.}
    &
    \scriptsize \textit{Provide a one-sentence caption for the provided image.}
    &
    \scriptsize \textit{Provide a one-sentence caption for the provided image.} \\
    \bottomrule
\end{tabular}
}
\caption{\textbf{LMM Adaptation Strategies for Zero-Shot Regional Captioning.} This figure visualizes the two input-adaptation strategies used to benchmark Large Multimodal Models (LMMs) on Region-level Captioning tasks: Trace, Dense, and Region-Set Captioning. \textbf{Top Row:} The original image corresponding to each task. \textbf{Middle Row (Visual Prompting):} The task region is spatially indicated by superimposed red annotations (traces or bounding boxes). The LMM input includes this visually prompted image along with a detailed instruction (shown below the images) to describe the marked region while suppressing any mention of the annotations. \textbf{Bottom Row (Cropping):} The input is a minimal crop tightly encompassing the annotated region. The LMM is given a general instruction (shown below the images) to describe the presented image, implicitly forcing focus onto the region of interest.}
\label{fig:vllm_adaptation_grid}
\end{figure*}

LMMs typically process a whole image and cannot natively process explicit regional coordinates (such as bounding boxes or traces) as additional inputs, unless specifically trained with region-level box or mask annotations. To enable a fair comparison of our regional tasks, we propose two zero-shot adaptation strategies to inject regional information into the LMM.
\begin{itemize}
    \item \textbf{Visual Prompting:} The regional annotations (bounding boxes or traces) are drawn over the image, allowing the model to condition its generation on the spatial input. We then ask the model to describe the annotated image.
    \item \textbf{Cropping:} The image is cropped to the bounding box encompassing the regional annotation (be it a trace, a box, or a set of boxes). The cropped image is then fed to the LMM, forcing the model to focus its attention solely on the region of interest.
\end{itemize}
An example of adaptation for each task with the respective prompt is shown in \autoref{fig:vllm_adaptation_grid}.
\subsection{Results Analysis}
Table \ref{tab:patchioner_vs_vllms} presents the comparison results of our Patch-ioner framework against state-of-the-art LMMs across various captioning granularities.

In the finer-level tasks, specifically Dense Captioning (VG v1.2) and the Trace Captioning (COCO), the pronounced performance gap between our framework and LMMs highlights the fundamental inability of LMMs to effectively reason at the precise region level based solely on a visual annotation or a simple crop without sufficient global context. For these granular tasks, the Visual Prompting strategy is generally preferable for LMMs because it preserves the essential context of the entire image. Furthermore, simple cropping is problematic for concave traces as it fails to precisely delineate the intended region of the image.

In tasks involving broader regions, such as Region-Set Captioning (COCO Entities), the Patch-ioner framework achieves better performance despite LMMs showing closer results, highlighting our higher capability to generate captions specifically guided by the set of input regions. For these broader regions, the Cropping adaptation strategy generally leads to better LMM performance than Visual Prompting, suggesting that Region-Set Captioning demands a lower level of context from the uncropped parts of the image with respect to Dense and Trace Captioning.

For the Image Captioning task, which relies on global understanding, LMMs generally demonstrate superior performance compared to our zero-shot captioner baseline. This is largely expected, as LMMs are bigger and extensively trained on massive image-text pairs.

\subsection{Comparison Highlights}
The comparison with powerful LMM baselines adapted for regional tasks clearly underscores the unique advantages of the Patch-ioner framework in terms of design, data requirements, parameter efficiency, and inference speed.

\paragraph{Designed for Granularity} The fundamental difference lies in the design objective: our Patch-ioner framework is explicitly engineered for multi-granularity, region-level captioning , while LMMs are built primarily for global image understanding. The need for ad-hoc adaptation strategies for LMMs (Visual Prompting or Cropping) inherently limits their precision, resulting in significantly lower performance on fine-grained regional tasks --as shown in Table \ref{tab:patchioner_vs_vllms}-- compared to our method.

\paragraph{Data and Training Efficiency}
Unlike LMMs, which are trained on massive paired image-text datasets and are therefore inherently supervised solutions, our approach adheres to a strict zero-shot paradigm. The text decoder is trained solely on text corpora, eliminating the dependency on costly, large-scale image-text supervision for training the generative module.

\paragraph{Parameter and Computational Efficiency}
The Patch-ioner framework is dramatically more lightweight and parameter-efficient than LMMs. We provide a compact model (e.g., $0.21\text{B}$ parameters for the Talk2DINO configuration), contrasting sharply with the multi-billion parameter counts typical of LMMs (e.g., $3\text{B}$ to $4\text{B}$ parameters in the tested models).

\paragraph{Inference Speed and Scalability}
One of our framework's most critical advantages is its inference efficiency for multiple regions within a single image. For an image containing multiple annotations (such as in the Dense Captioning task, which can have over a hundred boxes), we require only a single forward pass of the frozen vision backbone to extract all patch features. These pre-computed patch features can then be reused to caption arbitrary regions. Conversely, an LMM adapted through Cropping or Visual Prompting requires a full inference of the vision-language model for every single annotation, leading to dramatically slower performance when scaling to dense or complex regional tasks

\paragraph{}

\section{Ablations on Text Decoder Architectures}

We tested different architectures and sizes for the text decoder network.
In particular, we trained 
GPT-2 small~\cite{radford2019language},
Gemma 3 270M~\cite{deepmind2025gemma3},
Qwen 3 0.6B~\cite{qwen3},
LLaMa 3.2 1B~\cite{grattafiori2024llama}
and Qwen 3 1.7B~\cite{qwen3}.
 Details on training hyperparameters are provided in \S\ref{sec:implementation-details}. The dataset adopted for these trainings is made of the ground truth captions of COCO train Karpathy split. The total number of different textual captions for that split is 566,747.
 Table \ref{tab:training-different-decoders} reports the scores obtained when varying the decoder in our framework.
 
 Across all tasks, we observe that increasing the capacity of the textual decoder does not translate into consistent improvements. GPT-2 small (124M) performs surprisingly strongly, outperforming larger models on Trace and Region-Set captioning in terms of CIDEr, while maintaining competitive RefPAC-S scores. Larger decoders such as Gemma 3 (270M), Qwen3 (0.6B), and Llama 3.2 (1B) yield similar or slightly worse results, and the largest tested model, Qwen3 1.7B, exhibits the lowest performance on all CIDEr metrics.
These findings suggest that, under our training regime and dataset size (COCO train Karpathy split, ~566k captions), model capacity is not the limiting factor. Larger decoders tend to overfit more easily and provide little benefit in a setting where the captioning supervision is narrow in distribution and relatively small compared to their pretraining scale. Moreover, since our overall architecture relies on a strong visual encoder capable of providing semantic local representations, the decoder’s primary role is to map visual representations to fluent text; in this context, compact autoregressive models appear sufficiently expressive.
Interestingly, RefPAC-S shows minimal variation across decoder sizes, indicating that semantic alignment with the image features (rather than surface-level text quality) is largely preserved even with larger models. However, the trend of decreasing CIDEr with increasing model size suggests that bigger decoders may introduce unnecessary linguistic diversity that hurts match-based metrics.
Overall, these results highlight that %
changing the decoder affects only marginally the performances.

\begin{table*}[t]
\centering
\begin{tabular}{lcccccccccc}
    \toprule
    \multicolumn{2}{l}{\makecell[l]{Captioning Task: \\ (Dataset)}} & \multicolumn{2}{c}{\makecell{Trace \\ (COCO)}} & \multicolumn{2}{c}{\makecell{Dense \\ (VG v1.2)}} & \multicolumn{2}{c}{\makecell{Region-Set \\ (COCO Entities)}} & \multicolumn{3}{c}{\makecell{Image \\(COCO)}} \\
    \cmidrule(lr){1-2}\cmidrule(lr){3-4}\cmidrule(lr){5-6}\cmidrule(lr){7-8}\cmidrule(lr){9-11}
    Textual Decoder & \# Parameters & C & P & C & P & C & P & C & P & CLIP-S \\
    \midrule
GPT2 & 124M & \textbf{27.9} & \textbf{78.7} & \textbf{31.9} & \textbf{78.8} & \textbf{109.1} & \textbf{87.5} & \textbf{69.2} & \textbf{87.4} & {72.8} \\ %
Gemma 3 & 270M & 23.5 & \underline{78.4} & 25.5 & 78.5 & 98.7 & \textbf{87.5} & \underline{61.0} & \underline{87.2} & 73.2 \\
Qwen3 & 0.6B & 23.5 & \underline{78.4} & 25.0 & 78.4 & 98.7 & \textbf{87.5} & 59.7 & \underline{87.2} & 73.4 \\
LLama 3.2 & 1B & \underline{24.2} & 78.3 & \underline{26.0} & 78.4 & \underline{101.0} & \underline{87.4} & 60.3 & 87.1 & \underline{73.5} \\
Qwen3 & 1.7B & 22.9 & 78.2 & 24.5 & 78.2 & 95.6 & \underline{87.4} & 57.9 & 87.1 & \textbf{73.7} \\
\bottomrule
\end{tabular}
\caption{\textbf{Training different decoders.} CIDEr (C) and RefPAC-S (P) across four captioning tasks. The model adopted is T2D + Memory ($\approx$ DeCap) trained on COCO train Karpathy split. %
}
\label{tab:training-different-decoders}
\end{table*}

\section{Ablations on Text-only Training Dataset}

In this section, we compare the results obtained when using the classical training strategy adopted by the models such as \cite{li2023decap,zeng2024meacap,nukrai2022text} in the zero-shot captioning task --- which consists of training for 10 epochs on the collection of texts from the COCO dataset --- with training on a different dataset, to assess generalization capabilities of our framework.

For this comparison, we took the best Patch-ioner configuration --- which consists of using Talk2DINO~\cite{barsellotti2024talking} as visual backbone and a memory bank of texts borrowed from COCO as in~\cite{li2023decap}---
and we trained it on a subset of ReLaion.

\subsection{Selected dataset: ReLaion 600M}

ReLaion is a large-scale image-text dataset containing approximately 600 million image-caption pairs sourced from LAION-2B~\cite{schuhmann2022laion5b}. Each entry includes multiple machine-generated captions of varying quality%
, along with a ``best caption'' field obtained by ranking the alternatives using a CLIP-based scoring function. In our ablation we exclusively use this ``best caption'' attribute, since it provides a higher-quality textual signal while avoiding noisy or inconsistent descriptions that typically arise in large crawled datasets.

To ensure a fair comparison with the standard COCO-based training strategy from prior work~\cite{li2023decap,zeng2024meacap,nukrai2022text}, we adopt a subset of ReLaion sized so that one training epoch over that matches %
the number of training steps of the standard COCO training lasting 10 epochs.
Since the COCO training split contains roughly 560k captions, one epoch on a 5.6M-sample subset of ReLaion results in approximately the same number of optimization steps, while a 28.3M-sample subset corresponds to five times more steps. In all experiments, we maintain the same Patch-ioner configuration (our best-performing model) and apply identical optimization settings.

This setup allows us to assess the generalization capability of our framework when trained on broader, noisier web-scale text compared to the highly curated COCO captions that are commonly used in zero-shot captioning pipelines.

\subsection{Discussion}

The results in Table~\ref{tab:training-on-different-datasets} highlight three main trends.

\paragraph{(1) ReLaion training improves cross-dataset generalization.}
Training on ReLaion --- even for a single epoch --- consistently improves performance across captioning tasks compared to the classical COCO-only training. The 5.6M-sample setting yields clear gains on Trace and Dense Captioning, and the larger 28.3M subset further enhances Region-Set and Image Captioning scores. These improvements indicate that Patch-ioner benefits substantially from the wider linguistic and visual variability captured in ReLaion, despite its noisier nature.

\paragraph{(2) Gains are not uniform across tasks.}
While the Trace and Dense tasks benefit the most from ReLaion (e.g., +2.4 CIDEr on Trace and +1.7 on Dense when moving from COCO to ReLaion 5.6M), the COCO-based Region-Set task exhibits a more nuanced behavior. The larger ReLaion subset (28.3M) significantly boosts Region-Set CIDEr, suggesting that recognizing localized entities and relations requires broader caption diversity, which becomes available only at larger scale.

\paragraph{(3) Memory Bank from ReLaion is less effective than the one from COCO.}
When employing 500k captions randomly sampled from ReLaion as memory bank, performance drops across many tasks. 
This suggests that using a collection of texts for the projection mechanism sampled from COCO leads the model to provide captions that are closer to the COCO ground-truth ones, particularly from a syntactic standpoint. In fact, we can notice how the larger performance drop is on the CIDEr metric.

Overall, these findings show that Patch-ioner adapts well to large-scale noisy text, outperforming COCO-trained baselines on most tasks, and that the advantages grow with the size of the ReLaion subset. This demonstrates that our framework can leverage broad web-scale text distributions to improve zero-shot captioning, even when training the textual decoder for a single epoch.
However, we pick the model trained on COCO as reference to have a fair comparison with existing models.

\begin{table*}[t]
\centering
\resizebox{\textwidth}{!}{
\begin{tabular}{lccccccccc}
    \toprule
    \multicolumn{1}{l}{\makecell[l]{Captioning Task: \\ (Dataset)}} & \multicolumn{2}{c}{\makecell{Trace \\ (COCO)}} & \multicolumn{2}{c}{\makecell{Dense \\ (VG v1.2)}} & \multicolumn{2}{c}{\makecell{Region-Set \\ (COCO Entities)}} & \multicolumn{3}{c}{\makecell{Image \\(COCO)}} \\
    \cmidrule(lr){1-1}\cmidrule(lr){2-3}\cmidrule(lr){4-5}\cmidrule(lr){6-7}\cmidrule(lr){8-10}
    Text-only Training Dataset & C & P & C & P & C & P & C & P & CLIP-S \\
    \midrule
COCO Train & 27.9 & \underline{78.7} & {31.9} & {78.8} & \underline{109.1} & {87.5} & 69.2 & 87.4 & \underline{72.8} \\ %
ReLaion 5.6M & \textbf{30.3} & \textbf{79.0} & \textbf{34.1} & \textbf{79.0} & 109.0 & \underline{87.7} & \underline{69.3} & \underline{87.5} & 72.5 \\
ReLaion 28.3M & \underline{29.7} & \textbf{79.0} & \underline{33.6} & \underline{78.9} & \textbf{113.5} & \textbf{87.9} & \textbf{70.6} & \textbf{87.7} & \textbf{73.0} \\
\hdashline
Using a Memory Bank of 500k ReLaion Captions \\
Relaion 5.6M & 26.7 & 78.4 & 33.8 & 79.2 & 86.3 & 85.3 & 54.1 & 85.0 & 70.1 \\
ReLaion 28.3M & 26.8 & 78.3 & 33.4 & 79.0 & 86.6 & 85.3 & 54.6 & 85.0 & 70.5 \\
\bottomrule
\end{tabular}
}
\caption{\textbf{Training on different datasets.} CIDEr (C) and RefPAC-S (P) across four captioning tasks. The model adopted is T2D + Memory ($\approx$ DeCap) using the GPT2 textual decoder.
}
\label{tab:training-on-different-datasets}
\end{table*}

\section{Additional Results: Aggregation Strategies, Input Resolution, Text Collection}
\label{sec:aggr-size}

\sethlcolor{LightCyan}

We tested several patch aggregation strategies and input resolutions for our model and the other baselines.

\paragraph{Patch Aggregation.}
\label{sec:patch-aggregation}
In cases where we are not captioning a single patch, %
we test different aggregation functions for merging the $\mathbf{v}_i$ in a selected set $S$ of visual patches:
\begin{enumerate}[label=\alph*)]
    \item \textbf{uniform}, the average box patch representations; %
    \item \textbf{gaussian}, for rectangular configurations of contiguous patches --- i.e., either the full image or a bounding box; we consider a weighted average of patches representations where central patches weigh more; specifically, we assign to each patch $(a,b)$ coordinates in a uniform square grid $[-1,1]^2$ (i.e., the top-left and bottom-right patches have $(-1, -1)$ and $(1,1)$ coordinates, respectively), and a weight of $e^{-\left(a^2+b^2\right)}$ in the average, and
    \item \textbf{attention}, a weighted average of box patches representations, with patch weights defined as the average attention map of the last layer of $\psi_\text{v}$.
\end{enumerate}

\paragraph{Input Resolution.}
For patch-based captioning, we followed Talk2DINO~\cite{barsellotti2024talking} and used an input image resolution of 518x518, obtaining 37 14x14 patches per side when using the Talk2DINO backbone.
The original DeCap~\cite{li2023decap}, ViECap~\cite{fei2023transferable}, MeaCap~\cite{zeng2024meacap} implementations uses the CLIP B/32 backbone with 224x224 input images with 7 patches per side.
We also tested with the CLIP B/16 backbone, resulting in 14 patches per side at 224x224 resolution, and with 592x592 input image size, to obtain the same number of patches as in our framework (37 per side).

We report results of these additional configurations for all baselines: DeCap, ViECap, and MeaCap.
While the main paper reports only the best configuration per task and model, in this section, we report and discuss the results of all the tested configurations.
We perform these tests on COCO-derived datasets and on VG v1.2 for dense captioning.
\hl{We highlight the rows in the tables corresponding to the configurations reported in the main paper.}

\paragraph{Trace Captioning.}

Table~\ref{fullTableTraceCaptCOCO} reports trace captioning results.
We did not apply the \textit{gaussian} weighting scheme for this task, as the sparse discontinuous traces often do not identify a rectangular region needed to apply this scheme.
We notice that a) the simple average of the trace patches provides the best performance in our framework b) as expected, using the CLIP B16 backbone, that extracts finer patches, improves over the standard CLIP B32  backbone used in the baseline methods, and c) resolutions higher than 224 only marginally improve performance for baselines in this task.

\begin{table*}
\centering
\resizebox{\linewidth}{!}{
\begin{tabular}{lclcc*6r}
\toprule
Model & \# Patches & Backbone & Input & Weighting & B & M & R & C & S & P \\
\rowcolor{lightgray}
\midrule
\multicolumn{11}{l}{\textbf{Image-based}} \\
DeCap@224 & 7 & CLIP B32 & CLS & - & 2.1 & 9.7 & 21.7 & 21.1 & 8.8 & 75.2 \\
DeCap@224 & 14 & CLIP B16 & CLS & - & 2.2 & 9.8 & 21.8 & 21.3 & 8.7 & 75.4 \\
\rowcolor{LightCyan}
DeCap@592 & 37 & CLIP B16 & CLS & - & 2.0 & 9.6 & 21.5 & 20.5 & 8.7 & 75.3 \\
\midrule
ViECap@224 & 7 & CLIP B32 & CLS & - & 2.5 & 9.8 & 22.4 & 24.8 & 9.3 & 74.1 \\
ViECap@224 & 14 & CLIP B16 & CLS & - & 2.5 & 9.9 & 22.3 & 24.7 & 9.5 & 74.5 \\
\rowcolor{LightCyan}
ViECap@592 & 37 & CLIP B16 & CLS & - & 2.3 & 9.6 & 22.1 & 24.3 & 9.5 & 74.4 \\
\midrule
MeaCap@224 & 7 & CLIP B32 & CLS & - & 2.3 & 9.4 & 21.5 & 23.1 & 8.9 & 74.2 \\
MeaCap@224 & 14 & CLIP B16 & CLS & - & 2.3 & 9.3 & 20.8 & 23.4 & 9.0 & 74.6 \\
\rowcolor{LightCyan}
MeaCap@592 & 37 & CLIP B16 & CLS & - & 2.2 & 9.1 & 20.3 & 22.5 & 9.0 & 74.4 \\
\midrule
\rowcolor{lightgray}
\multicolumn{11}{l}{\textbf{Patch-based (Our Framework)}} \\
\rowcolor{LightCyan}
T2D + Mem. ($\simeq$ DeCap) @518 & 37 & DINOv2 B14 & \patchesSymbol{} & \avgTrace{} & \textbf{2.5} & \textbf{10.7} & \textbf{23.2} & \textbf{27.9} & \textbf{12.6} & \textbf{78.7} \\
T2D + Mem. ($\simeq$ DeCap) @518 & 37 & DINOv2 B14 & \patchesSymbol{} & \attnMapWeighted{} & 2.4 & 10.4 & 22.7 & 27.6 & 12.0 & 78.1 \\
\bottomrule
\end{tabular}
}
\caption{Trace Captioning results on COCO test set.}
\label{fullTableTraceCaptCOCO}
\end{table*}

\paragraph{Dense Captioning.}

Table~\ref{tab:fullTableDenseCaptvg12} reports the results of the dense captioning task.
For our framework, changing the weighting strategy does not cause significant performance changes.
The best baselines are the ones Region-based, which consist of applying the captioners to the CLS tokens of image crops specified by the bounding boxes (e.g., DeCap@224 Crop).
The difference between the CLIP B/16 and B/32 versions is usually small or negligible.

\begin{table*}
\centering
\resizebox{\textwidth}{!}{
\begin{tabular}{lclcc*7r}
\toprule
Model & \# Patches & Backbone & Input & Weighting & mAP & M & B & R & C & S & P \\
\midrule
\rowcolor{lightgray}
\multicolumn{12}{l}{\textbf{Image-based}} \\
DeCap@224 & 7 & CLIP B32 & CLS & - & 0.15 & 8.40 & 0.94 & 15.61 & 19.38 & 9.38 & 73.71 \\
\rowcolor{LightCyan}
DeCap@224 & 14 & CLIP B16 & CLS & - & 0.14 & 8.48 & 0.95 & 15.70 & 19.11 & 9.40 & 73.94 \\
DeCap@592 & 37 & CLIP B16 & CLS & - & 0.15 & 8.37 & 0.92 & 15.67 & 18.53 & 9.26 & 73.91 \\
\hdashline
ViECap@224 & 7 & CLIP B32 & CLS & - & 0.13 & 8.25 & 1.02 & 16.06 & 24.18 & 9.97 & 73.03 \\
ViECap@224 & 14 & CLIP B16 & CLS & - & 0.14 & 8.30 & 1.01 & 15.86 & 23.81 & 9.91 & 73.49 \\
\rowcolor{LightCyan}
ViECap@592 & 37 & CLIP B16 & CLS & - & 0.14 & 8.17 & 1.00 & 15.86 & 23.26 & 9.82 & 73.35 \\
\hdashline
MeaCap@224 & 7 & CLIP B32 & CLS & - & 0.13 & 8.04 & 0.98 & 15.40 & 23.22 & 9.66 & 72.97 \\
MeaCap@224 & 14 & CLIP B16 & CLS & - & 0.13 & 8.01 & 1.03 & 15.15 & 23.37 & 9.49 & 73.55 \\
\rowcolor{LightCyan}
MeaCap@592 & 37 & CLIP B16 & CLS & - & 0.13 & 7.86 & 0.99 & 15.13 & 22.77 & 9.43 & 73.50 \\
\midrule
\rowcolor{lightgray}
\multicolumn{12}{l}{\textbf{Region-based}} \\
DeCap@224 \cropCaptions{} & 7 & CLIP B32 & CLS & - & 0.17 & 10.03 & 1.35 & 18.20 & 23.61 & 10.90 & 77.09 \\
\rowcolor{LightCyan}
DeCap@224 \cropCaptions{} & 14 & CLIP B16 & CLS & - & 0.18 & 10.33 & 1.40 & 18.44 & 24.56 & 11.28 & 77.76 \\
DeCap@592 Crop & 37 & CLIP B16 & CLS & - & 0.14 & 8.30 & 1.05 & 16.39 & 17.20 & 7.78 & 75.47 \\
\hdashline
\rowcolor{LightCyan}
ViECap@224 Crop & 7 & CLIP B32 & CLS & - & 0.15 & 9.32 & 1.42 & 17.79 & 26.40 & 10.07 & 74.34 \\
ViECap@224 Crop & 14 & CLIP B16 & CLS & - & 0.16 & 9.59 & 1.46 & 18.03 & 27.13 & 10.43 & 75.62 \\
ViECap@592 Crop & 37 & CLIP B16 & CLS & - & 0.12 & 7.83 & 1.14 & 16.02 & 20.20 & 7.44 & 73.26 \\
\hdashline
\rowcolor{LightCyan}
MeaCap@224 Crop & 7 & CLIP B32 & CLS & - & 0.15 & 9.64 & 1.46 & 18.00 & 28.62 & 10.98 & 75.08 \\
MeaCap@224 Crop & 14 & CLIP B16 & CLS & - & 0.16 & 10.03 & 1.57 & 18.45 & 30.53 & 11.51 & 76.35 \\
MeaCap@592 Crop & 37 & CLIP B16 & CLS & - & 0.12 & 7.93 & 1.19 & 16.17 & 21.32 & 7.86 & 73.69 \\
\midrule
\rowcolor{lightgray}
\multicolumn{12}{l}{\textbf{Patch-based (Our Framework)}} \\
\rowcolor{LightCyan}
T2D + Mem. ($\simeq$ DeCap) @518 & 37 & DINOv2 B14 & \patchesSymbol{} & \avgPatches{} & 0.21 & 10.63 & 1.36 & 18.59 & 31.94 & 15.03 & 78.82 \\
T2D + Mem. ($\simeq$ DeCap) @518 & 37 & DINOv2 B14 & \patchesSymbol{} & \gaussianWeighted{} & \textbf{0.22} & \textbf{10.82} & \textbf{1.43} & \textbf{18.82} & \textbf{32.80} & \textbf{15.48} & \textbf{79.14} \\
T2D + Mem. ($\simeq$ DeCap) @518 & 37 & DINOv2 B14 & \patchesSymbol{} & \attnMapWeighted{} & 0.21 & 10.31 & 1.27 & 18.17 & 30.58 & 14.72 & 78.69 \\
\bottomrule
\end{tabular}
}
\caption{Dense Captioning results on VG v1.2 test set.}
\label{tab:fullTableDenseCaptvg12}
\end{table*}

\paragraph{Region-Set Captioning.}

Table \ref{fullTableRegionSetCaptCOCO}
shows the results in the region-set captioning task on COCO Entities. %
The gap between zero-shot image captioners and our framework's captioners narrows in this task due to the more global nature of it, which requires the model to produce a caption for the whole image while focusing on certain regions.
Also in this task, the choice of weighting strategy only marginally affects the performance of the models in our framework.

\begin{table*}
\centering
\resizebox{\textwidth}{!}{
\begin{tabular}{lclcc*6r}
\toprule
Model & \# Patches & Backbone & Input & Weighting & B & M & R & C & S & P \\
\midrule
\rowcolor{lightgray}
\multicolumn{11}{l}{\textbf{Image-based}} \\
DeCap@224 & 7 & CLIP B32 & CLS & - & 10.1 & 19.0 & 38.0 & 94.4 & 26.4 & 86.9 \\
\rowcolor{LightCyan}
DeCap@224 & 14 & CLIP B16 & CLS & - & 10.0 & 19.4 & 38.3 & 95.1 & 26.8 & 87.4 \\
DeCap@592 & 37 & CLIP B16 & CLS & - & 9.6 & 18.6 & 37.5 & 91.4 & 25.9 & 86.7 \\
\hdashline
\rowcolor{LightCyan}
ViECap@224 & 7 & CLIP B32 & CLS & - & 11.2 & 18.2 & 38.9 & 102.7 & 27.0 & 85.0 \\
ViECap@224 & 14 & CLIP B16 & CLS & - & 11.3 & 18.3 & 38.6 & 102.2 & 26.9 & 85.4 \\
ViECap@592 & 37 & CLIP B16 & CLS & - & 10.8 & 17.8 & 37.9 & 99.2 & 26.5 & 85.0 \\
\hdashline
\rowcolor{LightCyan}
MeaCap@224 & 7 & CLIP B32 & CLS & - & 10.4 & 17.7 & 37.0 & 97.9 & 25.9 & 85.2 \\
MeaCap@224 & 14 & CLIP B16 & CLS & - & 10.1 & 17.5 & 35.5 & 96.5 & 25.7 & 85.4 \\
MeaCap@592 & 37 & CLIP B16 & CLS & - & 9.3 & 16.9 & 34.6 & 91.1 & 25.5 & 85.0 \\
\midrule
\rowcolor{lightgray}
\multicolumn{11}{l}{\textbf{Patch-based (Our Framework)}} \\
T2D + Mem. ($\simeq$ DeCap) @518 & 37 & DINOv2 B14 & CLS & - & 9.1 & 16.9 & 35.0 & 89.4 & 25.4 & 85.5 \\
\rowcolor{LightCyan}
T2D + Mem. ($\simeq$ DeCap) @518 & 37 & DINOv2 B14 & \patchesSymbol{} & \avgPatches{} & 11.5 & 19.3 & 38.8 & 109.1 & 29.4 & 87.5 \\
T2D + Mem. ($\simeq$ DeCap) @518 & 37 & DINOv2 B14 & \patchesSymbol{} & \gaussianWeighted{} & \textbf{11.6} & \textbf{19.6} & \textbf{39.3} & \textbf{111.6} & \textbf{30.1} & \textbf{87.7} \\
T2D + Mem. ($\simeq$ DeCap) @518 & 37 & DINOv2 B14 & \patchesSymbol{} & \attnMapWeighted{} & 11.0 & 19.0 & 38.3 & 107.0 & 29.3 & 87.4 \\
\bottomrule
\end{tabular}
}
\caption{Region-Set Captioning results for COCO Entities test set.}
\label{fullTableRegionSetCaptCOCO}
\end{table*}

\paragraph{Image Captioning.}
In Table \ref{fullTableImgCaptCOCO}, we report the results of standard zero-shot image captioning.
In addition to the already described weighting schemes, we test one additional configuration for our framework that is \textit{\oneHot{}}, where the decoding is applied to the central patch of the image. %
We can observe that the most effective strategy for the image captioning task %
is \textit{\attnMapWeighted{}}. This is coherent with results from~\cite{barsellotti2024talking}, where they suggest the attention-weighted patch means to use Talk2DINO for global tasks such as image-text retrieval.

\newcommand{\memTrucc}{GT Memory}%

\paragraph{Memory Bank.}
Considering that in the memory-based model of our framework (that is similar to \cite{li2023decap}) we tackle the modality gap through a projection based on a collection of texts, we tested how much the selection of the texts in the memory bank influences performance.
In Table \ref{fullTableImgCaptCOCO}, we also report the results obtained by that model when in its memory bank there are also ground-truth captions of the test set (rows marked with \textit{\memTrucc{}}).
This provides a sort of upper bound to performance when varying the text collection used as memory.
We observe that in this configuration, the performance only slightly improves (+0.5\%), indicating that the model is robust to the choice of the memory bank.

\begin{table*}
\centering
\resizebox{\textwidth}{!}{
\begin{tabular}{lclcc*6r}
\toprule
Model & \# Patches & Backbone & Input & Weighting & B & M & R & C & S & P \\
\midrule
\rowcolor{lightgray}
\multicolumn{11}{l}{\textbf{Image-based}} \\
\rowcolor{LightCyan}
DeCap@224 & 7 & CLIP B32 & CLS & - & 23.46 & 25.12 & 50.06 & 87.40 & 19.14 & 90.58 \\
DeCap@224 & 14 & CLIP B16 & CLS & - & \textbf{23.89} & \textbf{25.51} & \textbf{50.34} & \textbf{89.64} & \textbf{19.52} & \textbf{91.05} \\
DeCap@592 & 37 & CLIP B16 & CLS & - & 22.43 & 24.64 & 49.25 & 84.57 & 18.66 & 90.36 \\
\hdashline
\rowcolor{LightCyan}
ViECap @224 & 7 & CLIP B32 & CLS & - & 26.70 & 23.99 & 50.85 & 89.67 & 17.54 & 88.45 \\
ViECap@224 & 14 & CLIP B16 & CLS & - & 26.3 & 24.0 & 50.3 & 89.5 & 17.6 & 88.8 \\
ViECap @592 & 37 & CLIP B16 & CLS & - & 25.60 & 23.38 & 49.52 & 86.84 & 17.08 & 88.33 \\
\hdashline
\rowcolor{LightCyan}
MeaCap@224 & 7 & CLIP B32 & CLS & - & 24.57 & 23.12 & 47.68 & 86.66 & 17.27 & 88.76 \\
MeaCap@224 & 14 & CLIP B16 & CLS & - & 23.6 & 22.7 & 45.5 & 85.1 & 17.3 & 89.0 \\
MeaCap@592 & 37 & CLIP B16 & CLS & - & 22.01 & 21.87 & 44.72 & 80.84 & 16.69 & 88.41 \\
\midrule
\rowcolor{lightgray}
\multicolumn{11}{l}{\textbf{Patch-based (Our Framework)}} \\
T2D + Mem. ($\simeq$ DeCap) @518 & 37 & DINOv2 B14 & \patchesSymbol{} & \oneHot{} & 15.68 & 18.46 & 40.84 & 55.53 & 12.66 & 84.26 \\
T2D + Mem. ($\simeq$ DeCap) @518 & 37 & DINOv2 B14 & \patchesSymbol{} & \avgPatches{} & 19.52 & 21.49 & 44.88 & 69.19 & 15.59 & 87.36 \\
T2D + Mem. ($\simeq$ DeCap) @518 & 37 & DINOv2 B14 & \patchesSymbol{} & \gaussianWeighted{} & 21.17 & 22.62 & 46.62 & 76.79 & 16.73 & 88.36 \\
\rowcolor{LightCyan}
T2D + Mem. ($\simeq$ DeCap) @518 & 37 & DINOv2 B14 & \patchesSymbol{} & \attnMapWeighted{} & 23.64 & 23.93 & 48.54 & 88.46 & 18.21 & 90.21 \\
\midrule
\rowcolor{Gray}
T2D + Mem. ($\simeq$ DeCap) @518 \memTrucc{} & 37 & DINOv2 B14 & CLS & - & 23.58 & 23.54 & 47.71 & 85.67 & 17.86 & 89.53 \\
\rowcolor{Gray}
T2D + Mem. ($\simeq$ DeCap) @518 \memTrucc{} & 37 & DINOv2 B14 & \patchesSymbol{} & \attnMapWeighted{} & 25.66 & 24.77 & 49.83 & 93.87 & 19.09 & 90.70 \\
\bottomrule
\end{tabular}
}
\caption{Image Captioning results on COCO test set.}
\label{fullTableImgCaptCOCO}
\end{table*}

\section{Modality Gap: Projection to Textual Space vs Training with Noise}
\label{sec:suppl:modality-gap}

In this section, we quantitatively assess the performance of two state-of-the-art solutions to overcome the modality gap.
In particular, we compared the configuration based on a memory bank of texts --- the one introduced in \S\ref{sec:method} --- with an alternative solution based on noise injection during the decoder training.
Additionally, we include in our comparison a baseline with no modality gap mitigation (\modalityGapNoStrategy{}), to highlight the benefits brought by each strategy.

\paragraph{Training with Noise.} 
Various works~\cite{gu2023can,nukrai2022text} proposed zero-shot image captioning solutions based on noise injection during the training of the text decoder. Through this strategy, the trained decoders are more effective in understanding semantic representations, even when those are not coming from the text modality.
To implement this strategy in our framework, we trained the textual decoder on the same collection of captions as for the memory bank-based configuration. We adopted Talk2DINO~\cite{barsellotti2024talking} textual space for the decoder input space, which is aligned to DINOv2~\cite{oquab2023dinov2} with registers~\cite{jose2024dinov2}.
Following the setting of \cite{gu2023can}, we added Gaussian noise with $\sigma^2= 0.08$ to the textual embeddings while leaving the other parameters unchanged (as defined in \S\ref{sec:implementation-details}).
In the next paragraphs, we report and compare the results for each task of Talk2DINO within our framework with the memory bank (\imProjSymbol{}) and with the training with noise (\noiseSymbol{}).

\begin{table*}
\setlength{\tabcolsep}{3pt}
\centering
\caption{\textbf{Mitigation of Modality Gap.} Comparison of Memory-based Projection (\imProjSymbol{}) vs Noise-trained Decoder (\noiseSymbol{}) across tasks.
}
\label{tab:ablation-modgap}
\resizebox{\textwidth}{!}{
\begin{tabular}{l*6c*7c*6c*7c}
\toprule
 & \multicolumn{6}{c}{Trace Captioning (COCO)} & \multicolumn{7}{c}{Dense Captioning (VG v1.2)} & \multicolumn{6}{c}{Region-Set Captioning (COCO Entities)} & \multicolumn{7}{c}{Image Captioning (COCO)} \\
\cmidrule(lr){2-7} \cmidrule(lr){8-14} \cmidrule(lr){15-20} \cmidrule(lr){21-27}
Mitigation & B & M & R & C & S & P 
           & mAP & M & B & R & C & S & P 
           & B & M & R & C & S & P 
           & B & M & R & C & S & P & CLIP-S \\
\midrule
\modalityGapNoStrategy & 1.2 & 9.1 & 18.3 & 14.7 & 8.5 & 75.1 
                       & 0.18 & 9.7 & 0.7 & 15.9 & 17.8 & 10.2 & 75.2 
                       & 5.0 & 15.0 & 29.4 & 59.4 & 21.1 & 82.2 
                       & 9.9 & 17.7 & 36.8 & 43.7 & 12.3 & 82.2 & 69.6 \\
\noiseSymbol           & \textbf{3.0} & \textbf{11.5} & \textbf{24.7} & \textbf{29.3} & 12.3 & 78.1 
                       & 0.20 & 10.4 & 1.2 & 17.8 & 26.3 & 12.6 & 77.0 
                       & 10.5 & 18.4 & 37.2 & 97.5 & 26.7 & 85.6 
                       & 19.6 & 21.5 & 45.4 & 65.5 & 15.5 & 86.2 & 70.9 \\
\imProjSymbol          & 2.5 & 10.7 & 23.2 & 27.9 & \textbf{12.6} & \textbf{78.7} 
                       & 0.21 & 10.6 & 1.4 & 18.6 & 31.9 & 15.0 & 78.8 
                       & 11.5 & 19.3 & 38.8 & 109.1 & 29.4 & 87.5 
                       & 19.5 & 21.5 & 44.9 & 69.2 & 15.6 & 87.4 & 72.8 \\
\bottomrule
\end{tabular}
}
\end{table*}

In Table~\ref{tab:ablation-modgap}, we compare the two modality gap mitigation strategies across multiple captioning tasks, and also report the performance of a baseline without any mitigation (\modalityGapNoStrategy{}). 
The baseline consistently underperforms compared to both the \imProjSymbol{} and \noiseSymbol{} configurations, indicating that, like other contrastively learned image-text encoders~\cite{liang2022mind}, Talk2DINO is also affected by the modality gap. These results highlight the importance of explicitly addressing this gap to achieve strong captioning performance.
For Trace Captioning, the \imProjSymbol{} method is slightly more effective in the semantic metric RefPAC-S, while the \noiseSymbol{} variant achieves marginally better scores in CIDEr, ROUGE-L, METEOR, and BLEU@4, with a minimal gap between the two approaches. In Dense Captioning, the \imProjSymbol{} model consistently outperforms the \noiseSymbol{} model across all metrics. Similarly, for Region-Set Captioning, both methods achieve strong results, but the \imProjSymbol{} method shows a clearer advantage, particularly in tasks closer to the patch level. Finally, in Image Captioning, the performance gap between the two architectures narrows, especially on the Flickr30k test split. In this scenario, the \imProjSymbol{} method performs significantly better when applied to the CLS token, whereas patch aggregation produces comparable results. However, the metrics reveal conflicting trends across different datasets.

\paragraph{Chosen Strategy.} Based on the observed results, we selected the projection-based approach (\imProjSymbol{}) as the primary strategy for overcoming the modality gap in our framework. 
While the noise injection method (\noiseSymbol{}) yielded competitive performance across multiple tasks, %
the \imProjSymbol{} method demonstrated superior performance in dense captioning and region-set captioning, as well as a clear advantage when applied to the CLS token in image captioning. Given these trends, and considering the stability of the projection-based approach across different evaluation settings, we adopted \imProjSymbol{} as the default configuration for our framework.

\section{Trace Captioning Benchmark Generation}
\label{sec:trace-details}
We construct our Trace Captioning dataset from the Localized Narratives dataset~\cite{pont2020connecting}. This dataset consists of mouse traces and their corresponding speech transcriptions, where annotators describe objects in images while moving the mouse pointer over them. 

The initial dataset samples include timestamped mouse traces and are composed of multiple sentences that thoroughly describe the trace, with the generated descriptions following the order of the mouse movement. However, our task does not require strict temporal coherence. Instead, we aim to generate a single, concise caption that describes the specific area covered by the localized trace, rather than a multi-sentence description.

To achieve this, we split the descriptions into individual sentences and align the traces accordingly. We then refine the traces by removing intermediate periods caused by transitions between sentences, which often occur when the annotator moves to a different region of the image. Specifically, we trim each trace by removing the first and last 15\% of points, eliminating these transitional segments.

Furthermore, we refine the captions by prompting the Llama3 8B model to rephrase the sentences, removing vague or subjective phrases such as "there is," "we can see," or "on the left of the image," and replacing them with concise, objective descriptions that refer specifically to the region covered by the trace. This rephrasing is crucial to ensure that each caption adheres to the standard format of image-captioning datasets and focuses only on the precise part of the image that the trace corresponds to. The LLM also helps identify and remove irrelevant sentences (e.g., "the image is blurred," "the image is edited"), which are then discarded along with their associated traces from the final benchmark.

\begin{figure*}[t]
        \newcommand{\colwid}{0.225\linewidth}
        \centering
        \sffamily

        \begin{subfigure}[t]{\colwid}
            \caption{\textsf{Localized Narrative}} 
            \includegraphics[width=\linewidth]{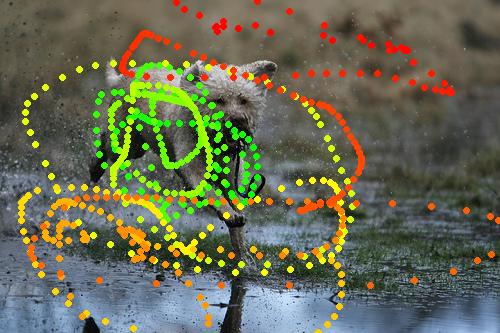}
            \tiny\raggedright
In this picture I can observe a dog running on the land. I can observe water and grass on the ground. The background is blurred.
        \end{subfigure}
        \hfill
        \begin{subfigure}[t]{\colwid}
            \caption{\textsf{Track 1}} %
            \includegraphics[width=\linewidth]{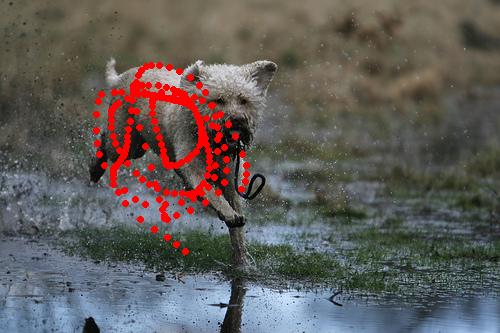}
            \tiny\raggedright
\textbf{Original}: In this picture I can observe a dog running on the land.
\\
\textbf{Processed}: A dog runs on the land.
        \end{subfigure}
        \hfill
        \begin{subfigure}[t]{\colwid}
            \caption{\textsf{Track 2}} %
            \includegraphics[width=\linewidth]{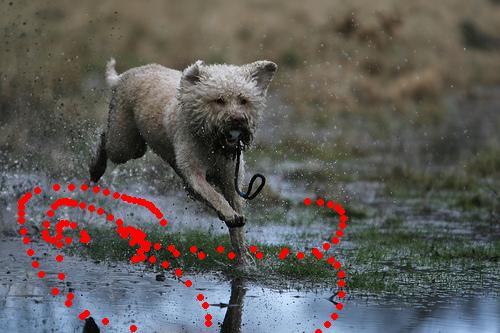}
            \tiny\raggedright
\textbf{Original}:  I can observe water and grass on the ground.
\\
\textbf{Processed}: Water and grass on the ground.
        \end{subfigure}
        \hfill
        \hfill
        \begin{subfigure}[t]{\colwid}
            \caption{\textsf{Track 3}} %
            \includegraphics[width=\linewidth]{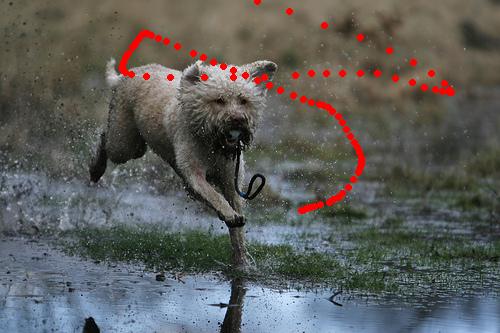}
            \tiny\raggedright
\textbf{Original}:  The background is blurred.
\\
\textbf{Processed}: \texttt{<INVALID>}
        \end{subfigure}
        \hfill

        \begin{subfigure}[t]{\colwid}
            \includegraphics[width=\linewidth]{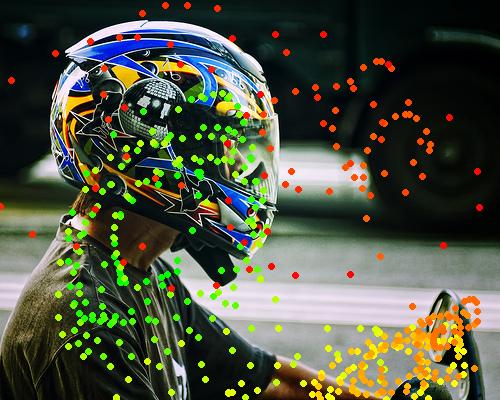}
            \tiny\raggedright
In this image there is a person wearing a helmet is on a vehicle. At the bottom of the image there are side mirrors. The background of the image is blurred.

        \end{subfigure}
        \hfill
        \begin{subfigure}[t]{\colwid}
            \includegraphics[width=\linewidth]{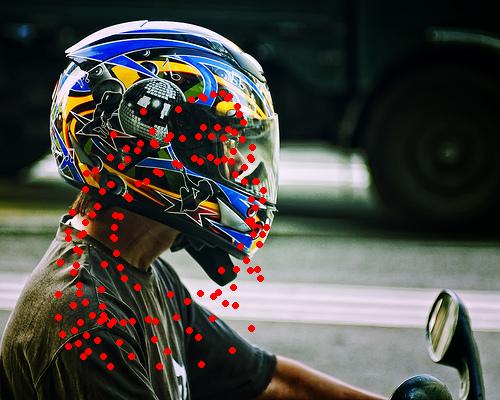}
            \tiny\raggedright
\textbf{Original}: In this image there is a person wearing a helmet is on a vehicle.
\\
\textbf{Processed}: A person wearing a helmet rides a vehicle.
        \end{subfigure}
        \hfill
        \begin{subfigure}[t]{\colwid}
            \includegraphics[width=\linewidth]{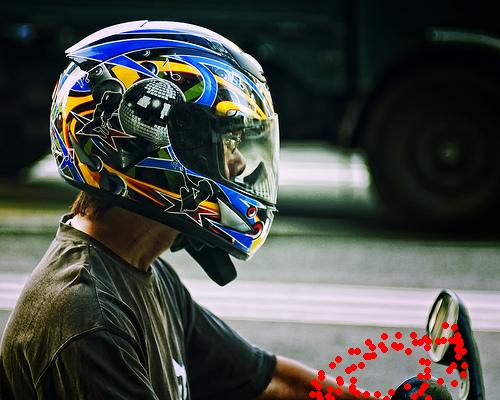}
            \tiny\raggedright
\textbf{Original}:  At the bottom of the image there are side mirrors.
\\
\textbf{Processed}: Side mirrors. 
        \end{subfigure}
        \hfill
        \hfill
        \begin{subfigure}[t]{\colwid}
            \includegraphics[width=\linewidth]{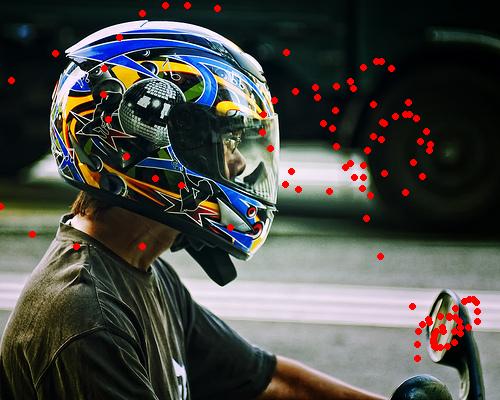}
            \tiny\raggedright
\textbf{Original}:  The background of the image is blurred.
\\
\textbf{Processed}: \texttt{<INVALID>}
        \end{subfigure}
        \hfill

        \begin{subfigure}[t]{\colwid}
            \includegraphics[width=\linewidth]{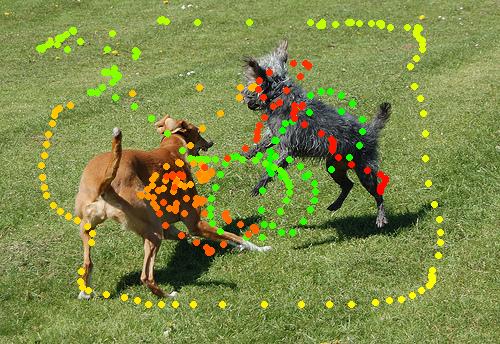}
            \tiny\raggedright
This image is taken outdoors. In this image we can see the green grass on the ground. In the middle of the image we can see there are two dogs.
        \end{subfigure}
        \hfill
        \begin{subfigure}[t]{\colwid}
            \includegraphics[width=\linewidth]{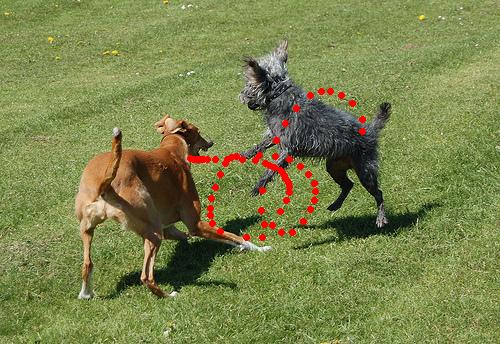}
            \tiny\raggedright
\textbf{Original}: This image is taken outdoors.
\\
\textbf{Processed}: \texttt{<INVALID>}
        \end{subfigure}
        \hfill
        \begin{subfigure}[t]{\colwid}
            \includegraphics[width=\linewidth]{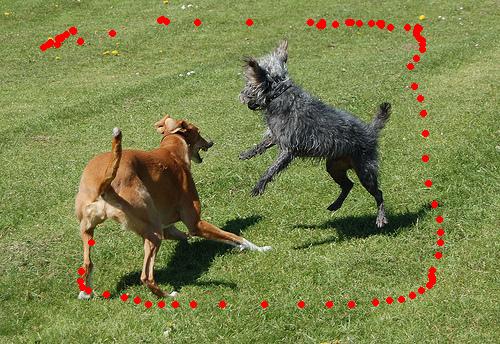}
            \tiny\raggedright
\textbf{Original}:  In this image we can see the green grass on the ground.
\\
\textbf{Processed}: Green grass on the ground. 
        \end{subfigure}
        \hfill
        \hfill
        \begin{subfigure}[t]{\colwid}
            \includegraphics[width=\linewidth]{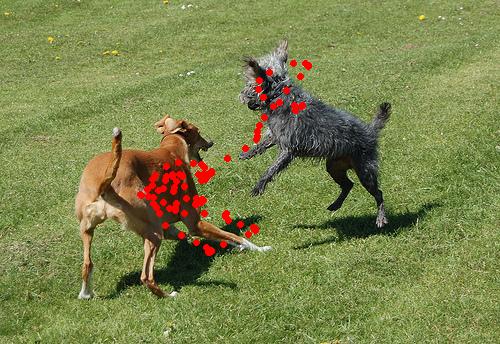}
            \tiny\raggedright
\textbf{Original}:  In the middle of the image we can see there are two dogs.
\\
\textbf{Processed}: Two dogs.
        \end{subfigure}
        \hfill
        
        \caption{\textbf{Narrative vs. Trace Samples}. The first column displays sample images from the Localized Narrative dataset~\cite{pont2020connectingLocalizedNarratives}. The remaining three columns show the corresponding mouse traces, along with the captions generated by the LLM. Captions marked with \texttt{<INVALID>} are removed from the dataset.}
        \label{fig:narratives_vs_trace}
    \end{figure*}

Figure~\ref{fig:prompt} shows the full prompt used to guide the Llama model in refining and cleaning the descriptions.
Figure~\ref{fig:narratives_vs_trace} illustrates how the initial narrative samples are transformed into final trace captioning samples through the process of trace splitting and caption rephrasing.

\begin{figure*}[ht]
\centering
\fontsize{8pt}{9.5pt}\selectfont
\fvset{breaklines, frame=single}
\begin{Verbatim}
I have image descriptions derived from spoken narratives. These need to be rewritten as concise, stand-alone captions in the style of the image-caption datasets. Follow these rules:

- Remove unnecessary narrative phrases like "we can see," "there is," "in this image," etc.  
- Ensure the caption is standalone and descriptive.  
- Use simple, objective language that highlights key elements.  
- Keep it concise—just a single phrase.  
- Follow the classical style of caption datasets.  
- If the description is vague, subjective, or does not describe a concrete visual element (e.g., "The image is taken indoor," "This image is blurred"), return `<INVALID>`.  
- Wrap the output in `{}` and add nothing else.

### **Examples:**  
- **Input:** "We can see a young elephant stands which is near the water in a wooded area."  
  **Output:** {A young elephant stands near the water in a wooded area.}  

- **Input:** "In this image I can see some young children kicking a soccer ball in a field."  
  **Output:** {A group of young children kicking a soccer ball around a field.}  

- **Input:** "In the left of the image, we see a pole that has two green street signs on it."  
  **Output:** {A pole has two green street signs on it.}  

- **Input:** "We can see two surfboards which are stuck in the sand along the seashore."  
  **Output:** {Two surfboards stuck in the sand along the seashore.}  

- **Input:** "This image consists of a man which rides a wakeboard behind a boat."  
  **Output:** {A man rides a wakeboard behind a boat.}  

- **Input:** "In the background, there are a bunch of sticky notes and a pair of scissors."  
  **Output:** {A bunch of sticky notes and a pair of scissors.}  

- **Input:** "It looks like a sepia-toned photograph of a motorcycle underneath the shadow of a
tree."  
  **Output:** {A sepia-toned photograph of a motorcycle underneath the shadow of a tree.}  
  
- **Input:** "There is a sky"  
  **Output:** {A sky.}  

- **Input:** "She is smiling."  
  **Output:** {A smiling girl.}  

- **Input:** "The image is taken indoor."  
  **Output:** {<INVALID>}  

- **Input:** "This image is edited."  
  **Output:** {<INVALID>}  

- **Input:** "The image is blurred."  
  **Output:** {<INVALID>}  

- **Input:** "I think he is about to jump."  
  **Output:** {<INVALID>}  

Now, rewrite the following captions accordingly. Wrap each in `{}` and add nothing else:  
<INPUT CAPTION>

\end{Verbatim}
\caption{\textbf{LLM Prompt for rephrasing trace captions}.}
\label{fig:prompt}
\end{figure*}

\section{Learned Patch Aggregation via Attention}

In the main paper we employ a parameter-free aggregation strategy to combine patch embeddings belonging to a region. While this choice preserves the zero-shot nature of our framework, we additionally explored whether a lightweight learned aggregation module could further improve the quality of region representations when local supervision is available.

\paragraph{Attention-based aggregation.}
Let $S = \{v_i\}_{i=1}^{N}$, with $v_i \in \mathbb{R}^D$, denote the set of patch embeddings corresponding to a selected region. In the default formulation of our framework, the region representation is obtained through mean aggregation:
\begin{equation}
v_{\text{mean}} = \frac{1}{N}\sum_{i=1}^{N} v_i .
\end{equation}

To investigate a learned alternative, we introduce a lightweight attention-based aggregation module that summarizes the patch set through a single transformer-style attention layer. The patch embeddings are processed jointly and produce a summary token that attends over the region patches. Formally, given the matrix of patch embeddings $V \in \mathbb{R}^{N \times D}$, we compute

\begin{equation}
v_{\text{att}} = \mathrm{Attn}(V),
\end{equation}

where $\mathrm{Attn}(\cdot)$ denotes a self-attention block that outputs a single \texttt{[CLS]} token representing the aggregated region information.

Rather than replacing the mean representation, we combine the two signals to preserve the stability of the parameter-free aggregation while allowing the model to learn refinements. The final region embedding is defined as

\begin{equation}
v_R = v_{\text{mean}} + \alpha \, v_{\text{att}},
\end{equation}

where $\alpha$ is a learnable scalar controlling the contribution of the attention-based summary. The parameter $\alpha$ is initialized to $0.1$, ensuring that the aggregation initially behaves close to the mean operator and gradually learns to incorporate the attention-based refinement during training.

\paragraph{Training setup.}
The attention-based aggregator is trained with region-level supervision for a single epoch using the COCO Trace Captioning training set derived from the LocNar COCO train split. We optimize the parameters of the attention layer and the scalar weight $\alpha$ using AdamW with learning rate $10^{-4}$, while keeping the visual backbone and the text decoder frozen.

\begin{table*}[h]
\definecolor{lightergray}{gray}{0.95}
\newcolumntype{g}{>{\columncolor{lightergray}}c}
\setlength{\tabcolsep}{3pt}
\centering
\resizebox{\textwidth}{!}{
\begin{tabular}{lggccgggcccggccgggccc}
\toprule
& \multicolumn{4}{c}{Trace Captioning} & \multicolumn{6}{c}{Dense Captioning} & \multicolumn{4}{c}{Region-Set Captioning} & \multicolumn{6}{c}{Image Captioning} \\
\cmidrule(lr){2-5} \cmidrule(lr){6-11} \cmidrule(lr){12-15} \cmidrule(lr){16-21}
& \multicolumn{2}{c}{COCO} & \multicolumn{2}{c}{Flickr30k} & \multicolumn{3}{c}{VG v1.2} & \multicolumn{3}{c}{VG-COCO} & \multicolumn{2}{c}{COCO Entities} & \multicolumn{2}{c}{Flickr30k Entities} & \multicolumn{3}{c}{COCO} & \multicolumn{3}{c}{Flickr30k} \\
\cmidrule(lr){2-3} \cmidrule(lr){4-5} \cmidrule(lr){6-8} \cmidrule(lr){9-11} \cmidrule(lr){12-13} \cmidrule(lr){14-15} \cmidrule(lr){16-18} \cmidrule(lr){19-21}
Model & \m{C} & \m{P} & \m{C} & \m{P} & \m{mAP} & \m{C} & \m{P} & \m{mAP} & \m{C} & \m{P} & \m{C} & \m{P} & \m{C} & \m{P} & \m{C} & \m{P} & \m{CLIP-S} & \m{C} & \m{P} & \m{CLIP-S} \\
\toprule
Learned Aggregation (T2D + Noise) & \textbf{73.9} & \textbf{81.9} & \textbf{34.8} & \textbf{77.2} & 17.8 & \textbf{49.4} & \textbf{81.7} &  17.7 & \textbf{48.9} & \textbf{81.7} & 67.8 & 83.6 & 25.7 & 73.3 & \textbf{74.2} & 86.5 & 70.9 & 27.3 & 79.7 & 66.0 \\
\rowcolor{LightCyan} 
Fixed Aggregation (T2D + Noise) & 29.3 & 78.1 & 19.3 & 75.6 & \textbf{20.3} & 26.3 & 77.0 & \textbf{20.3} & 26.4 & 76.9 & \textbf{97.5} & \textbf{85.6} & \textbf{37.1} & \textbf{76.5} & 65.5 & \textbf{86.2} & \textbf{70.9} & \textbf{27.8} & \textbf{80.8} & \textbf{67.0} \\
\bottomrule
\end{tabular}
}
\caption{
Learned vs. Fixed Aggregation
}
\label{tab:learned_aggr}
\end{table*}

\paragraph{Discussion.}
Table~\ref{tab:learned_aggr} reports the results of this experiment. The learned aggregator yields substantial improvements when evaluated on tasks with the same spatial granularity used during training (trace and dense captioning). However, these gains do not consistently transfer to tasks involving different region granularities (e.g., region-set or image captioning), where the fixed aggregation remains more balanced.

These findings suggest that when task-specific local supervision is available, learned aggregation can effectively refine region representations. However, in a unified zero-shot framework where no region-level annotations are assumed, the parameter-free aggregation strategy remains preferable due to its robustness across captioning granularities.

\section{More Qualitative Results}
\begin{figure*}[t]
    \centering
    \tiny
    \newcommand{\hd}[1]{\textbf{#1}}
    \renewcommand{\methodName}{Ours (Talk2DINO + Mem.)}
    \newcommand{\figw}{.26\textwidth}
    \setlength{\tabcolsep}{1pt}
    \resizebox{.9\textwidth}{!}{
    \begin{tabular}{r>{\hspace{0.1cm}}*2{p{\figw}>{\hspace{2pt}}}p{\figw}>{\hspace{1pt}}p{\figw}}
        \large
        \raisebox{1.4cm}{\rotatebox{90}{\sc \sffamily Patch}} &
        \includegraphics[width=\linewidth]{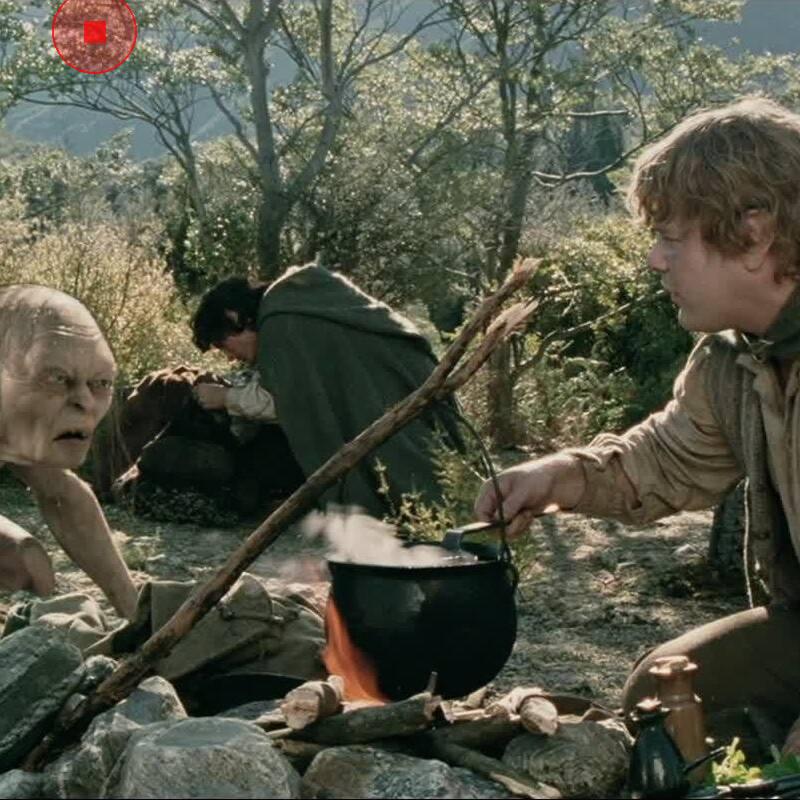} &
        \includegraphics[width=\linewidth]{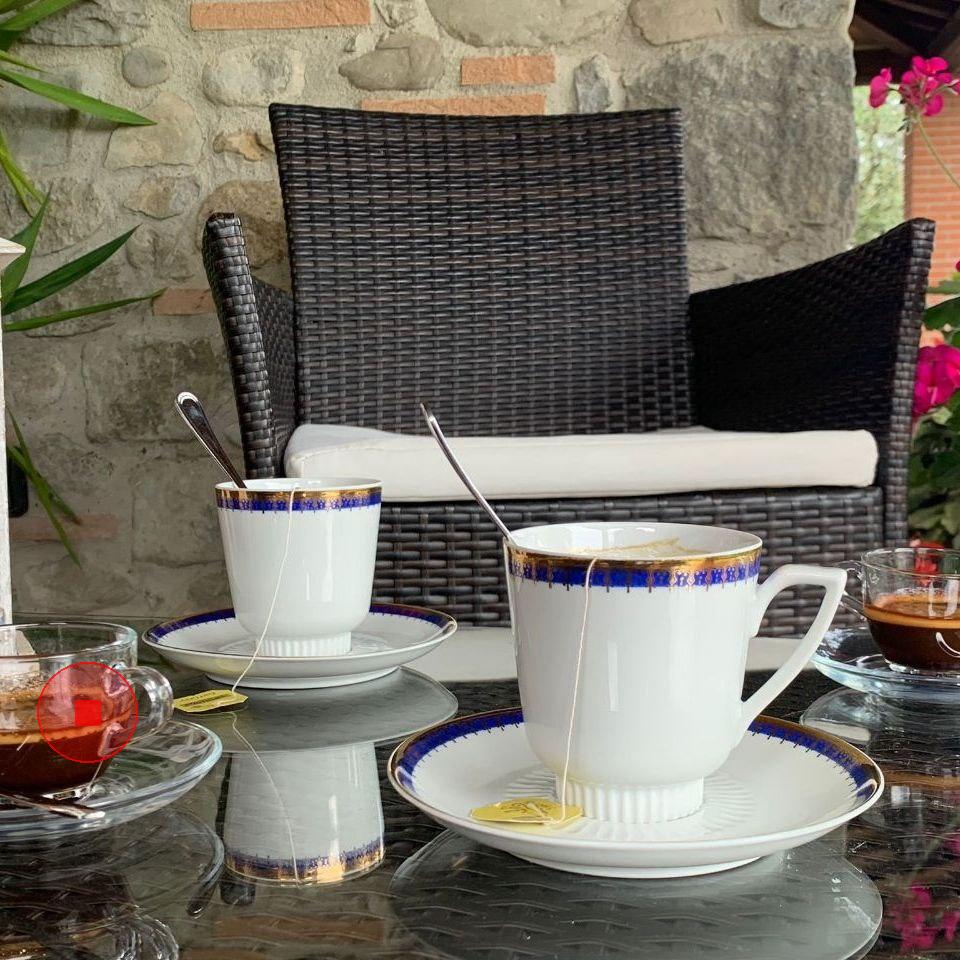} &
        \includegraphics[width=\linewidth]{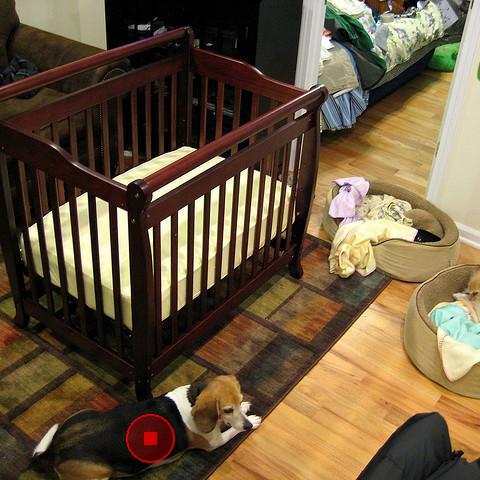} &
        \includegraphics[width=\linewidth]{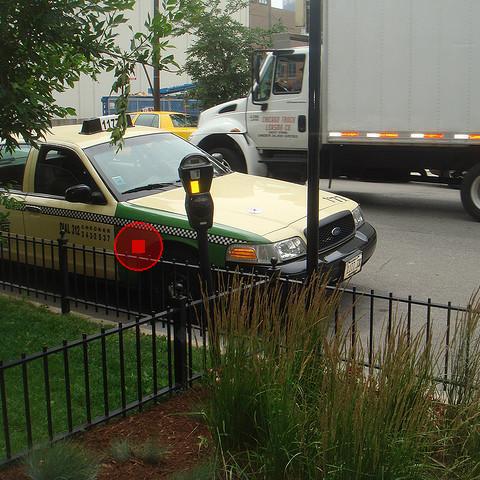} \\

        \hd{DeCap} &
        a group of people in a kitchen are cooking food. &
        a table with a cup of coffee and plates of silverware. &
        a small bed is curled up in a cluttered room. &
        a police car is parked on the side of a street. \\

        \rowcolor[gray]{.9}
        \hd{Ours (CLIP + Mem.)} &
        a couple of people that are standing around each other. &
        a bunch of people are sitting at the table together. &
        aa baby is in a bedroom with a white sink and toilet. &
        there are a few street signs in the middle of the neighborhood. \\
 
        \hd{\methodName{}} &
        a forest with trees in the background. &
        a cup of coffee with a spoon sitting on a plate. &
        a dog laying on a rug in a living room. &
        a fence that is next to a road. \\[2ex]

        \large
        \raisebox{.7cm}{\rotatebox{90}{\sc \sffamily Trace}} &
        \includegraphics[width=\linewidth]{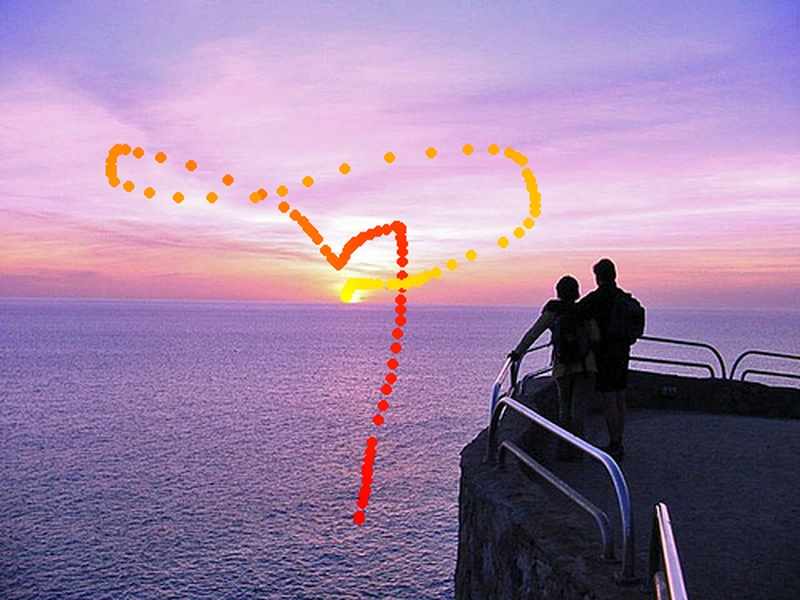} &
        \includegraphics[width=\linewidth]{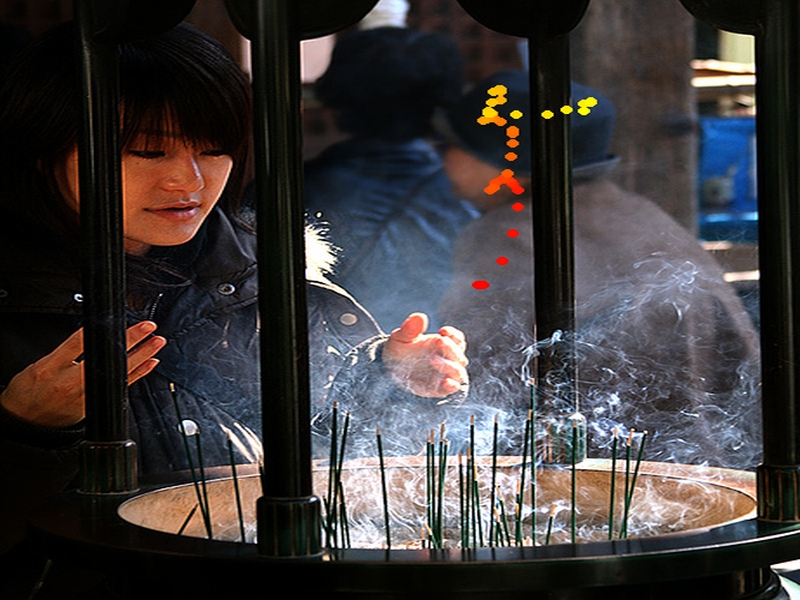} &
        \includegraphics[width=\linewidth]{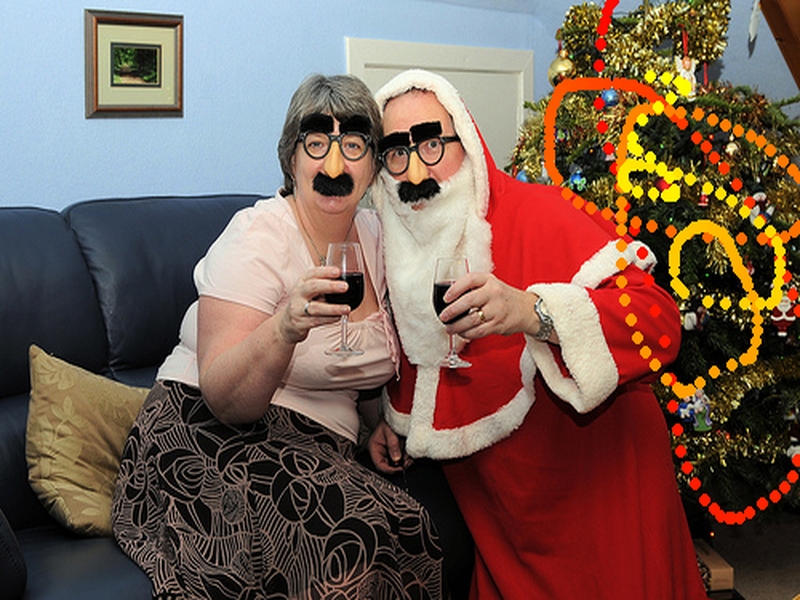} &
        \includegraphics[width=\linewidth]{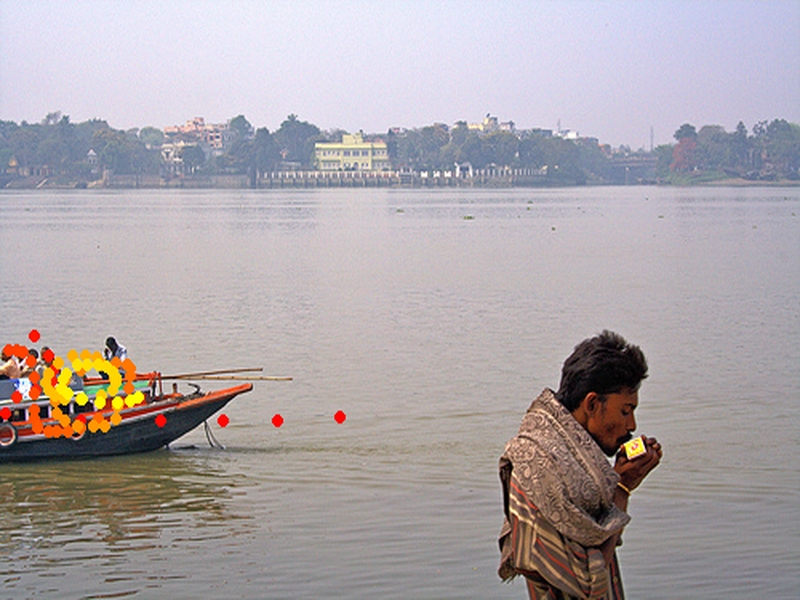} \\
 
        \rowcolor[gray]{.9}
        \hd{GT} &
Clouds and the sun in the sky. & 
A person wearing a cap. & 
A Christmas tree decorated with balls and toys. & 
Few people on a boat. \\ 
        
        \hd{DeCap} &
a couple of people are sitting on a bench looking at the ocean. & 
a woman at a table putting food in a pot. & 
two people posing with a man and woman having a glass of wine. & 
a man on a boat in a body of water with other boats. \\ 

        \rowcolor[gray]{.9}
        \hd{Ours (CLIP + Mem.)} &
a couple of people are on a boat by the ocean. & 
a couple of people are in a kitchen making food. & 
there are two people in a kitchen with a red sweater. & 
there are some people in the water by a boat. \\ 
        
        \hd{\methodName{}} &
a sunset in the distance in the sun & 
a person wearing a hat looking at something in the background & 
the christmas tree is decorated for christmas & 
a boat with many people on it \\

        \large
        \raisebox{.9cm}{\rotatebox{90}{\sc \sffamily Dense}} &
        \includegraphics[width=\linewidth]{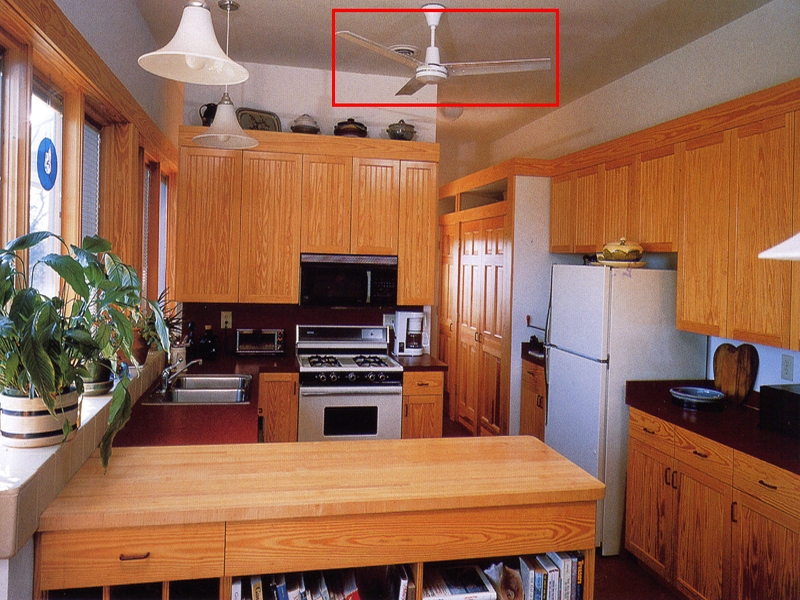} &
        \includegraphics[width=\linewidth]{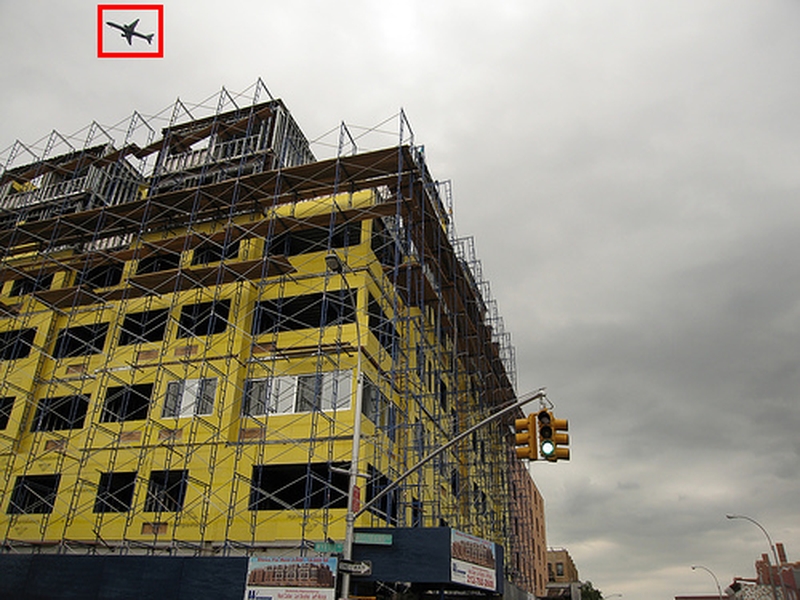} &
        \includegraphics[width=\linewidth]{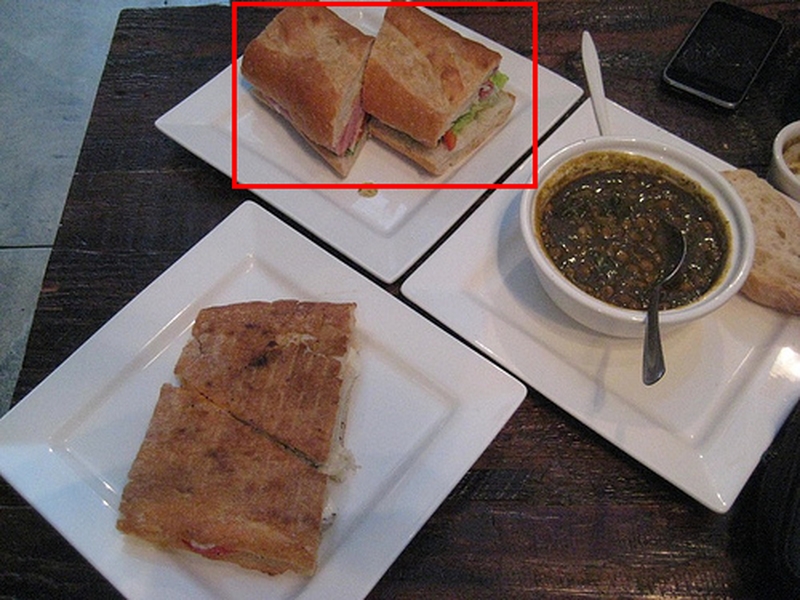} &
        \includegraphics[width=\linewidth]{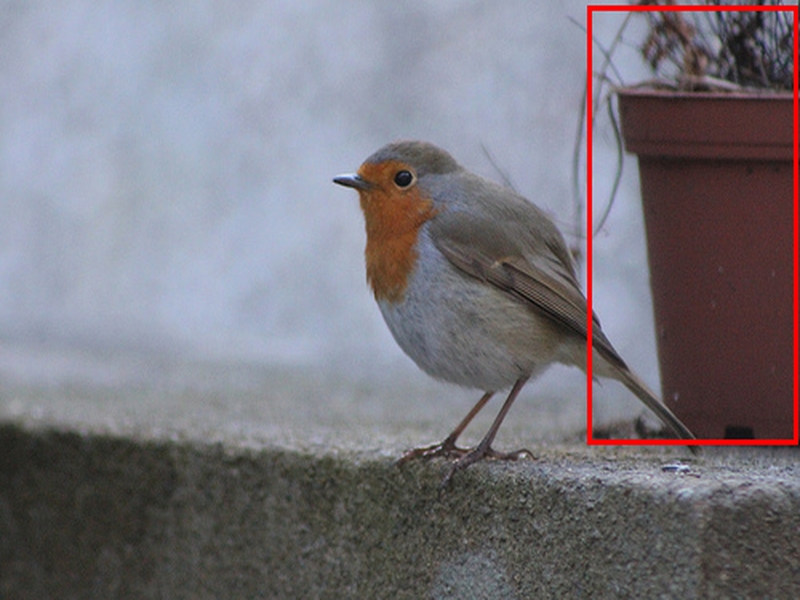} \\
 
        \rowcolor[gray]{.9}
                \hd{GT} &
a white ceiling fan hanging in the kitchen. & 
a plane flying in the sky. & 
two sandwiches on a plate. & 
potted plant on the ledge. \\ 
        
        \hd{DeCap} &
a kitchen with a large refrigerator , cabinets and stove. & 
a building is flying under a traffic light in the air near a building. & 
a sandwich and a plate of soup on a table. & 
a bird that is perched on a small bird. \\ 

        \rowcolor[gray]{.9}
        \hd{DeCap (Crop)} &
a bathroom sink with a variety of toilet above the wall. & 
a large airplane is in flight on the airport. & 
a sandwich on a plate containing a sandwich. & 
a a close up of a person standing by a person holding a phone. \\ 

        \hd{Ours (CLIP + Mem.)} &
a kitchen has a lot of fridge and a stove in it. & 
a lot of a building is outside of a yellow car. & 
the couple of food are in the kitchen with a meal. & 
there is a man that is about to take a trick. \\

        \rowcolor[gray]{.9}
        \hd{\methodName{}} &
a ceiling fan is hanging in the kitchen. & 
there is a plane flying high in the sky. & 
a plate topped with two sandwiches on a table. & 
a potted plant sitting on a ledge. \\ [2ex]

        \large
        \raisebox{.3cm}{\rotatebox{90}{\sc \sffamily Region Set}} &
        \includegraphics[width=\linewidth]{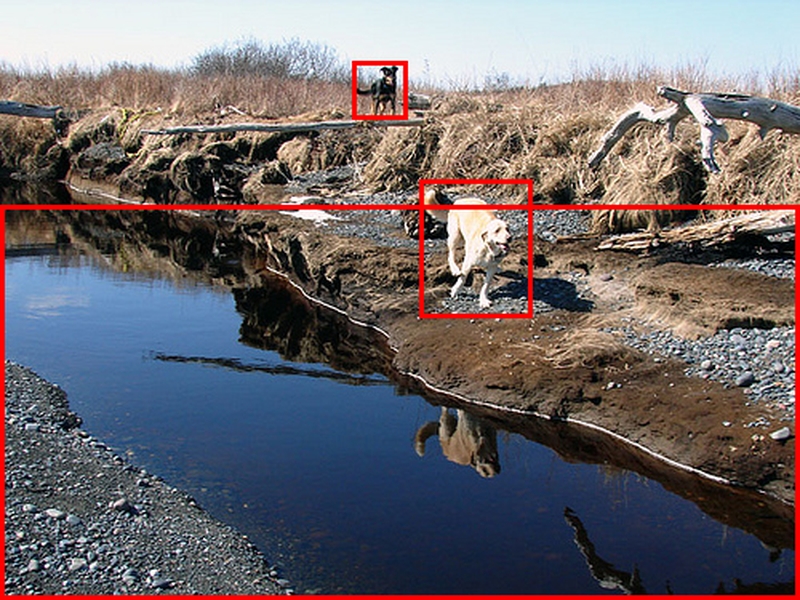} &
        \includegraphics[width=\linewidth]{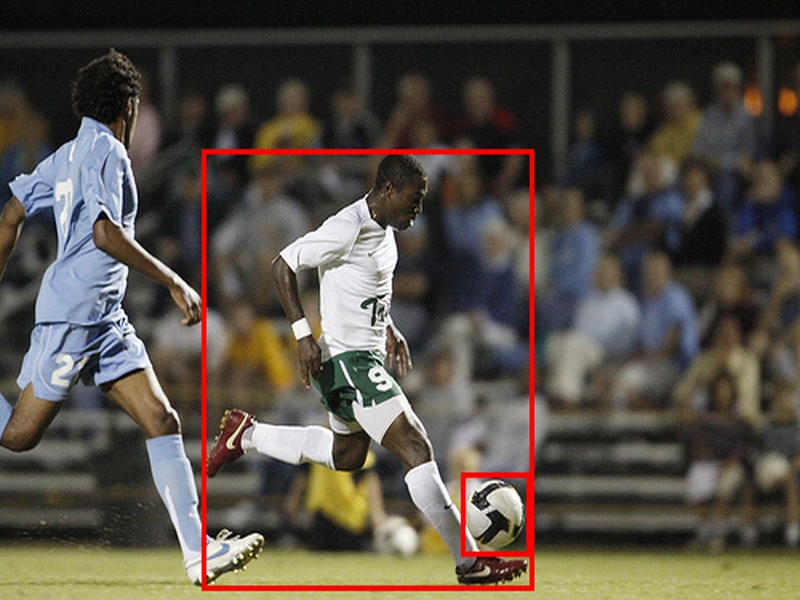} &
        \includegraphics[width=\linewidth]{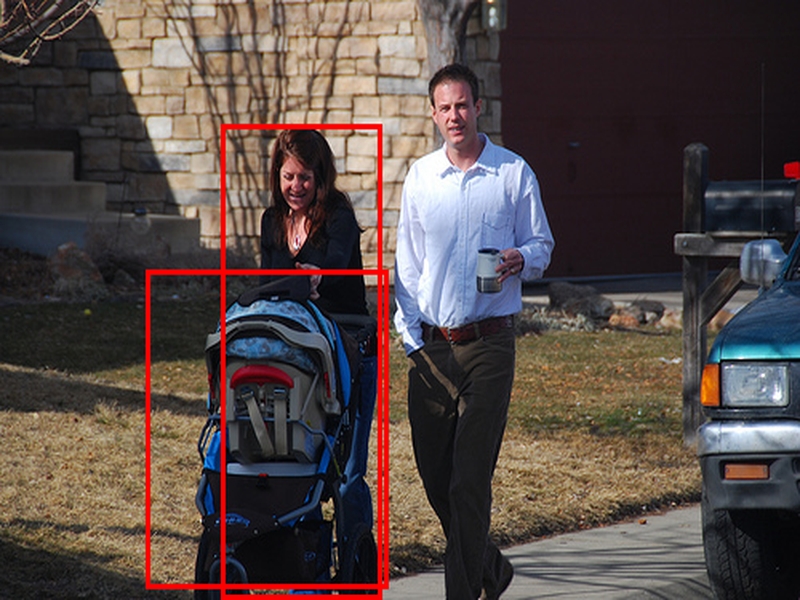} &
        \includegraphics[width=\linewidth]{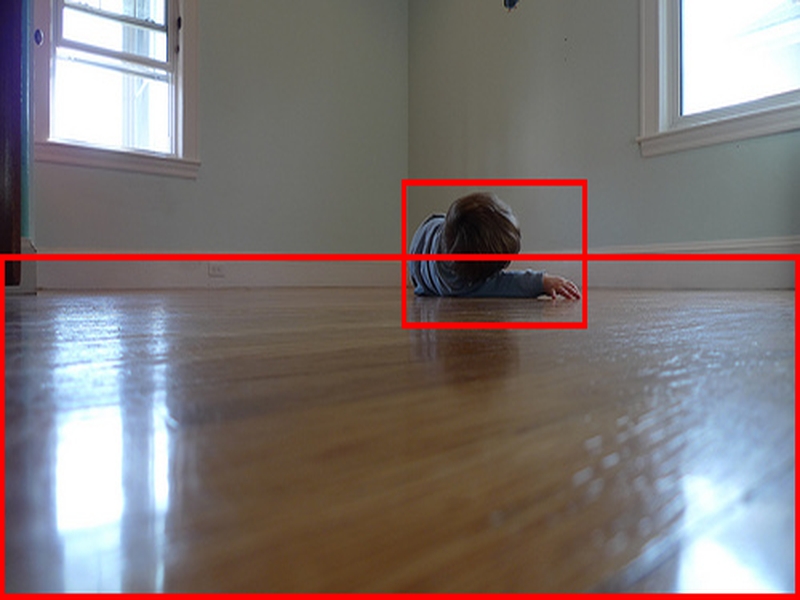} \\
 
        \rowcolor[gray]{.9}
        \hd{GT} &
Dogs near the edge of water . & 
A soccer player is running while kicking a ball . & 
A brown-haired woman is pushing a baby stroller . & 
The child lays on the hardwood floor .\\ 
        
        \hd{DeCap} &
a dog and his dogs are wading in the muddy water. & 
a soccer player in the soccer uniform tries to kick the ball. & 
a man and a child walking in the street while holding a stroller. & 
a young boy sitting on the floor in a room. \\ 

        \rowcolor[gray]{.9}
        \hd{Ours (CLIP + Mem.)} &
there are many things that are out in the water. & 
there are some people on a baseball field playing a game. & 
there are some cars and a man about to go down the street. & 
the kitchen is very clean and has an open door. \\ 
 
        \hd{\methodName{}} &
two dogs near one another near water. & 
a soccer player getting ready to kick the ball. & 
a woman pushing a stroller with a child inside. & 
a young child is laying on the floor. \\ 

        \large
        \raisebox{.8cm}{\rotatebox{90}{\sc \sffamily Image}} &
        \includegraphics[width=\linewidth]{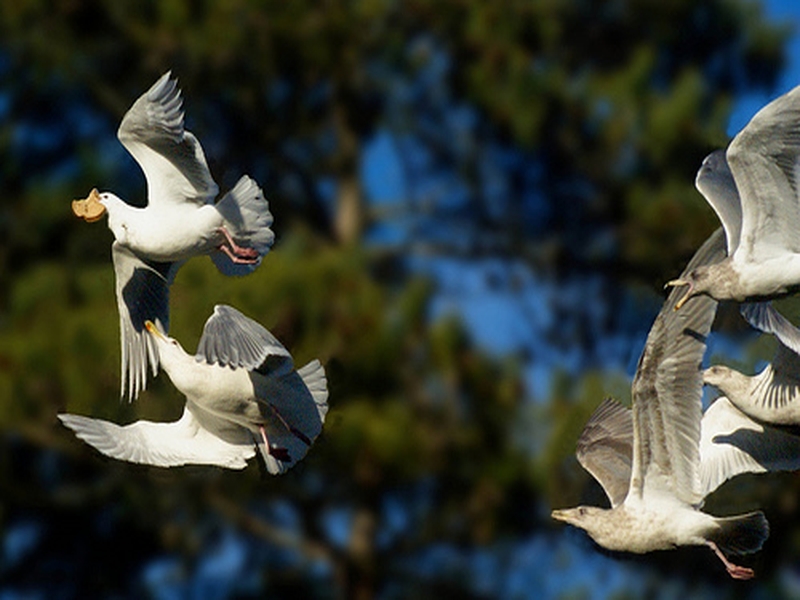} &
        \includegraphics[width=\linewidth]{images/supplementary/qualitatives1/image/example2.jpg} &
        \includegraphics[width=\linewidth]{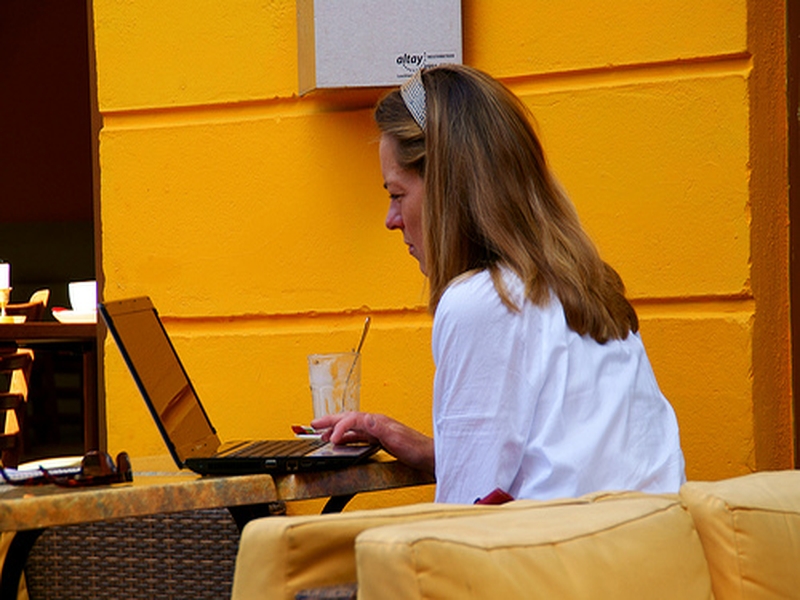} &
        \includegraphics[height=0.195\textwidth]{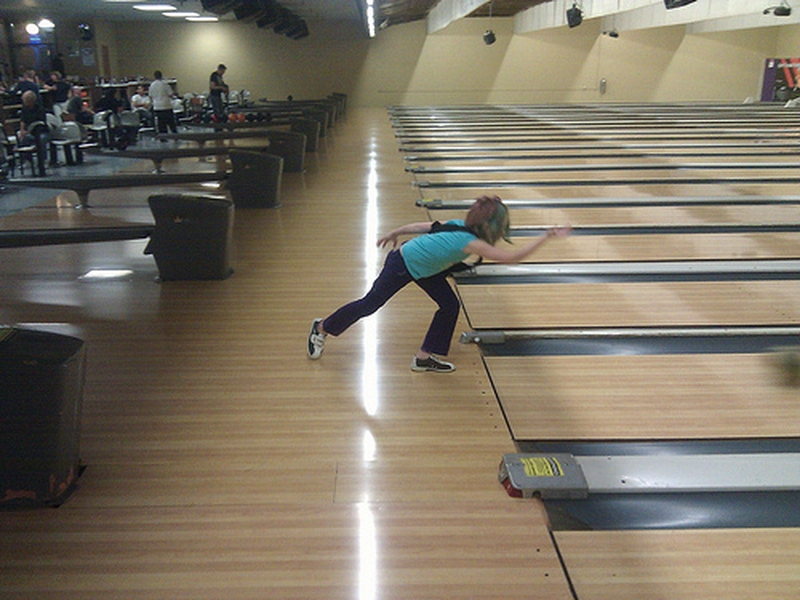} \\

        \rowcolor[gray]{.9}
        \hd{GT} &
Four birds are chasing another bird which has a piece of food in its mouth. & 
Brown-haired girl wearing a green tank top, talking on a cellphone. & 
A woman with blond-hair is sitting in a booth with a drink working on her laptop. & 
A young girl in a blue shirt is in a bowling alley, and is casting her ball down a lane. \\
        
        \hd{DeCap} &
a flock of birds flying over the water. & 
a woman talking on a cell phone while on a street. & 
a woman sitting at a table using a laptop. & 
a young girl playing a bowling game on wii. \\ 

        \rowcolor[gray]{.9}

        \hd{ZeroCap} &
        a gull mating. &
        a man in the back of a pickup truck with blood on the back. &
        a reader's writing on a laptop on desk-mounted computer. &
        a view hitting the deck pin at the end of the row stretch. \\

                \hd{CLOSE} &
        a group of birds flying over a body of water. &
        a woman looking at her cell phone while standing in a street. &
        a woman sitting at a table with a laptop and a drink. &
        on a person on a skateboard doing a trick. \\

        \rowcolor[gray]{.9}
        \hd{\methodName{}} &
a flock of birds flying in the sky. & 
a woman talking on a cell phone in a market. & 
a woman sitting at a cafe using her laptop. & 
a woman playing a bowling game on the bowling. \\ 
        
    \end{tabular}
    }
        \caption{\textbf{Qualitative results.} We report four predictions of our model and compare baselines from the finer (top) to the coarser (bottom) task. For trace captioning examples, the trace time is color-coded from start (red) to end (yellow). \textbf{DeCap} = DeCap applied on the whole image. 
    \textbf{DeCap (Crop)} = DeCap applied on cropped box. 
    \textbf{ZeroCap} = ZeroCap~\cite{tewel2022zerocap} applied to the whole image.
    \textbf{CLOSE} = CLOSE~\cite{gu2023can} applied to the whole image.
    \textbf{Ours (CLIP + Mem.)} = Our patch-based framework using CLIP as backbone and the projection as modality gap mitigation strategy.
     \textbf{Ours (Talk2DINO + Mem.)} = Our patch-based framework using Talk2DINO as backbone and the projection as modality gap mitigation strategy.
    \textbf{GT} = ground-truth caption.}
    \label{fig:more-examples}
\end{figure*}

Additional qualitative results are shown in Figures~\ref{fig:examples} and \ref{fig:more-examples}.
Note that the first rows of Figures \ref{fig:examples} and \ref{fig:more-examples} contain also qualitative results for single patch captioning, for which we do not have annotated data to report quantitative results.

As can be noticed in Figures \ref{fig:examples} and \ref{fig:more-examples}, the Region-Set Captioning task tends to align more closely with image-level captioning rather than strictly focusing on localized regions. 
This is expected since the ground-truth captions in the COCO Entities dataset originate from the image-level annotations of COCO, as stated in \cite{cornia2019show}.
\begin{figure*}[t]
    \centering
    \tiny
    \newcommand{\hd}[1]{\textbf{#1}}
    \renewcommand{\methodName}{Ours (Talk2DINO + Mem.)}
    \newcommand{\figw}{.26\textwidth}
    \setlength{\tabcolsep}{1pt}
    \resizebox{.9\textwidth}{!}{
    \begin{tabular}{r>{\hspace{0.1cm}}*2{p{\figw}>{\hspace{2pt}}}p{\figw}>{\hspace{1pt}}p{\figw}}
        \large
        \raisebox{1.4cm}{\rotatebox{90}{\sc \sffamily Patch}} &
        \includegraphics[width=\linewidth]{images/qualitatives/patch/example1.jpg} &
        \includegraphics[width=\linewidth]{images/qualitatives/patch/example2.jpg} &
        \includegraphics[width=\linewidth]{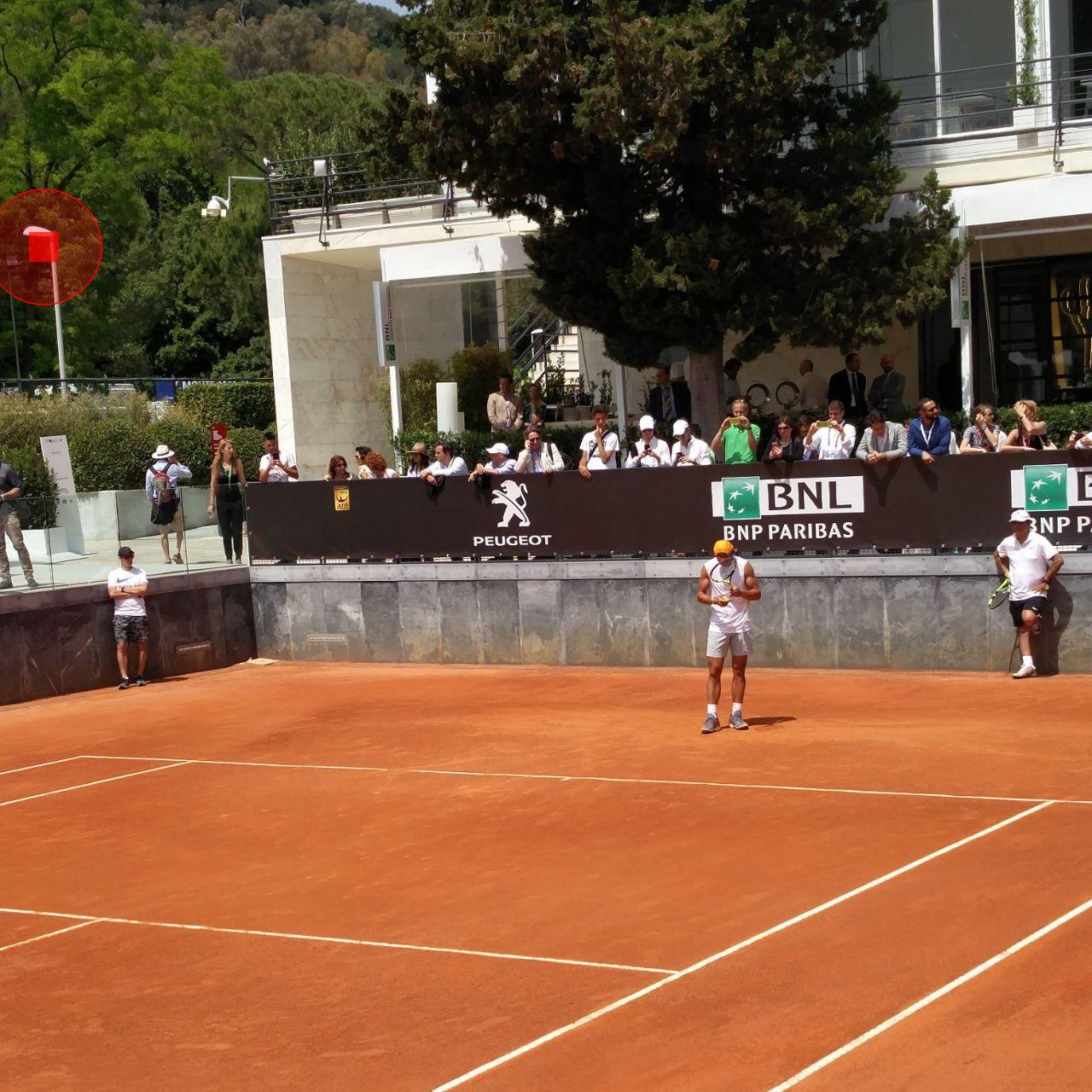} &
        \includegraphics[width=\linewidth]{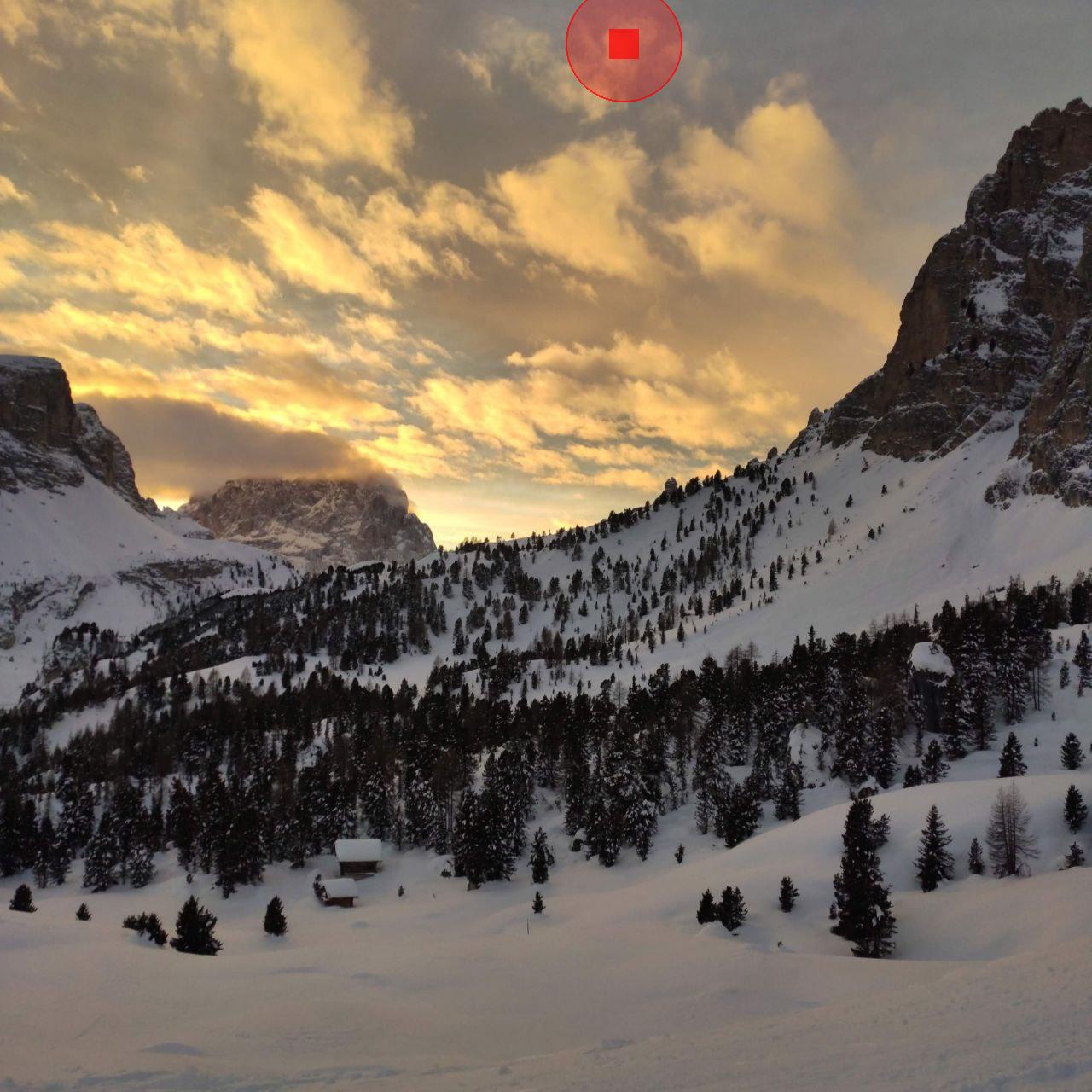} \\

        \hd{DeCap} &
        a cat is sleeping on a cluttered desk. &
        a cat is sleeping on a cluttered desk. &
        a tennis player is playing tennis on the court for a serve. &
        a few people are skiing on a snowy mountain. \\

        \rowcolor[gray]{.9}
        \hd{Ours (CLIP + Mem.)} &
        a cat is sitting on the bed and it’s contents. &
        a cat is sitting at a table with a full laptop . &
        a couple of people are in the middle of a tennis court. &
        a few people are skiing in a snowy mountain. \\
 
        \hd{\methodName{}} &
        a plant in a vase sitting on a table. & 
        office supplies , pens , toys , and other items on desk. &
        a street light in front of a large building. &
        a cloudy sky is seen in this cloudy day. \\

        \large
        \raisebox{.7cm}{\rotatebox{90}{\sc \sffamily Trace}} &
        \includegraphics[width=\linewidth]{images/qualitatives/track/example1.jpg} &
        \includegraphics[width=\linewidth]{images/qualitatives/track/example2.jpg} &
        \includegraphics[width=\linewidth]{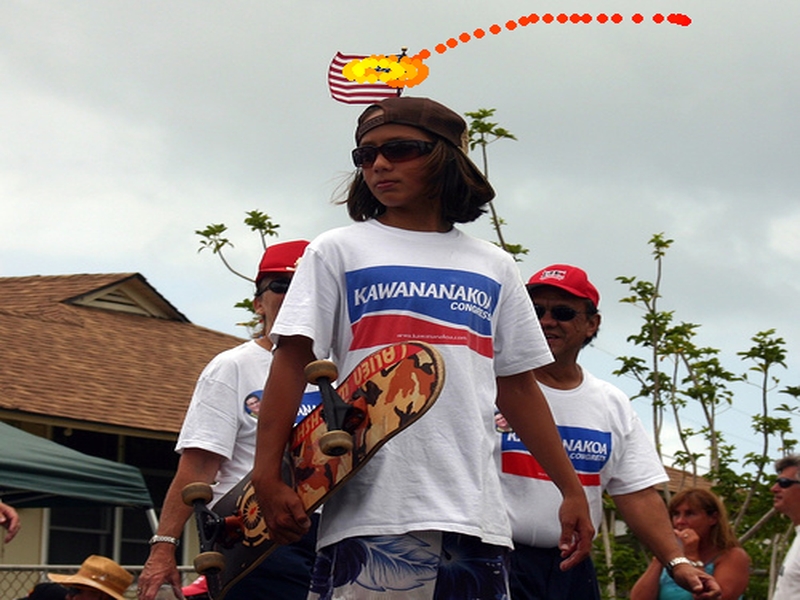} &
        \includegraphics[width=\linewidth]{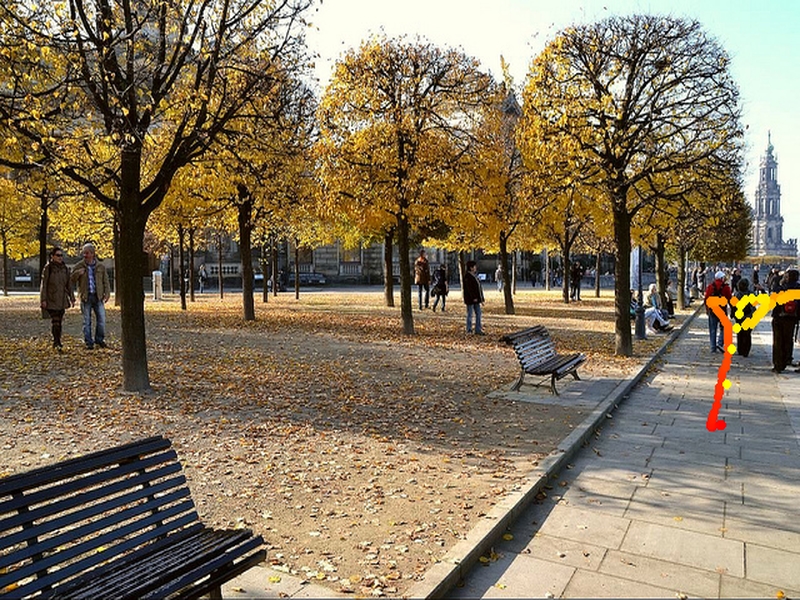} \\
 
        \rowcolor[gray]{.9}
        \hd{GT} &
        Two giraffes, rocks, and a fence. &
        A sky. &
        A flag. &
        People walking on a walkway.\\
        
        \hd{DeCap} &
        a giraffe in a zoo with a city in the background. &
        a giraffe in a zoo with a city in the background. &
        a man on a skateboard who is holding onto a skateboard. &
        a park filled with people sitting on benches near trees. \\

        \rowcolor[gray]{.9}
        \hd{Ours (CLIP + Mem.)} &
        there are some people that are in a lot by a tree. &
        there are some people that are out by a lot of trees. &
        there are some people that are in the water with a couple of them. &
        there are several traffic lights out in the wild. \\
        
        \hd{\methodName{}} &
        two giraffes standing in a fenced area. &
        a view of a city with a sky in the background. &
        a flag is flying high in the air. &
        a large group of people walking on a sidewalk. \\[2ex]

        \large
        \raisebox{.9cm}{\rotatebox{90}{\sc \sffamily Dense}} &
        \includegraphics[width=\linewidth]{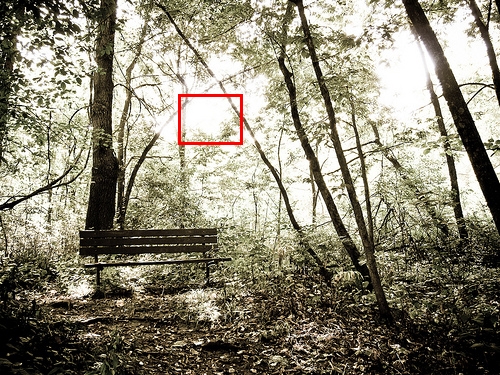} &
        \includegraphics[width=\linewidth]{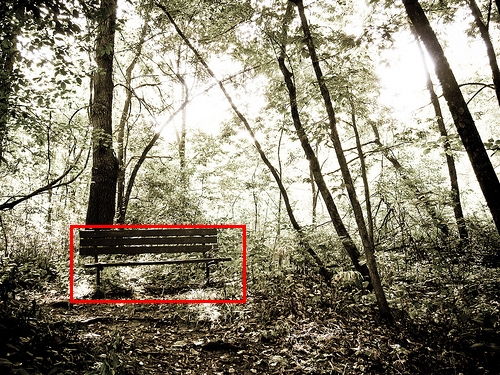} &
        \includegraphics[width=\linewidth]{images/qualitatives/dense/example3.jpg} &
        \includegraphics[width=\linewidth]{images/qualitatives/dense/example4.jpg} \\
 
        \rowcolor[gray]{.9}
        \hd{GT} &
        light shining through the trees. &
        bench sitting in the woods. &
        a clock at a train station. &
        black cat sitting on a bench. \\
        
        \hd{DeCap} &
        a bench sits in the middle of a wooded area. &
        a bench sits in the middle of a wooded area. &
        a train traveling along the platform of a public train. &
        a woman squatting on a bench with a cat. \\

        \rowcolor[gray]{.9}
        \hd{DeCap (Crop)} &
        a person in a tree is standing in the wild near trees. & 
        a bench sitting in the middle of a wooded area. &
        a black cat is leaning on a black cat. & 
        a a close up of a person standing by a person holding a phone. \\ 
        
        \hd{Ours (CLIP + Mem.)} &
        a bear is in the woods among the trees. &
        there are many trees that are standing in the woods. &
        a train is on the tracks and going by. &
        there is a person that is out on the kitchen. \\

        \rowcolor[gray]{.9}
        \hd{\methodName{}} &
        sun shining through the trees at sunset. &
        a park bench sitting in the middle of a wooded area. &
        a clock on a train station platform above a train. &
        a black cat is sitting on a black bench. \\[2ex]
        
        \large
        \raisebox{.3cm}{\rotatebox{90}{\sc \sffamily Region-Set}} &
        \includegraphics[width=\linewidth]{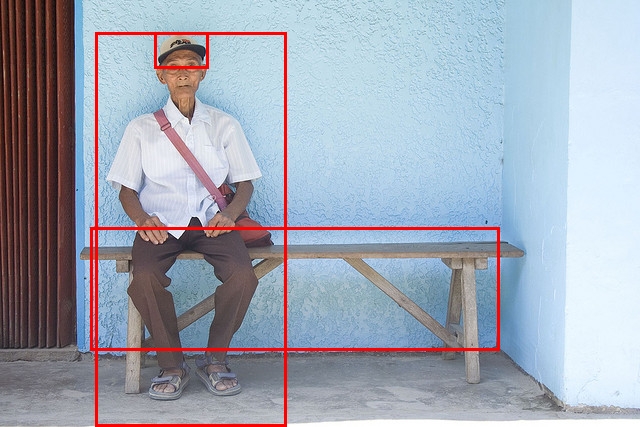} &
        \includegraphics[width=\linewidth]{images/qualitatives/set/example2.jpg} &
        \includegraphics[width=\linewidth,trim={0 .5cm 0 1cm},clip]{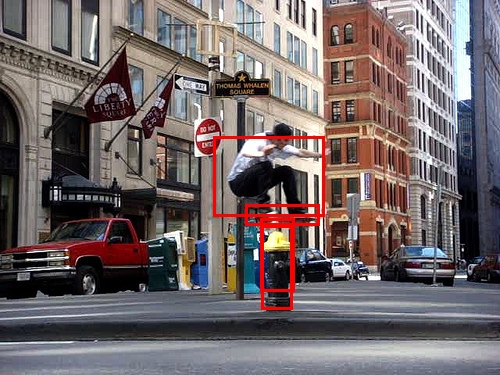} &
        \includegraphics[width=\linewidth]{images/qualitatives/set/example4.jpg} \\
 
        \rowcolor[gray]{.9}
        \hd{GT} &
        an elderly man in a cap sitting on a bench. &
        an old man sitting on a bench with a purse. &
        a man performing a trick near a fire hydrant. &
        a man swinging a baseball bat as another looks on. \\
        
        \hd{DeCap} &
        a man sitting on a bench while holding a door. &
        a man sitting on a bench while holding a door. &
        a man on a skateboard doing a trick. &
        a baseball player at bat getting ready to hit the ball. \\

        \rowcolor[gray]{.9}
        \hd{Ours (CLIP + Mem.)} &
        a bathroom has a blue floor and it is very clean. &
        a bathroom has a blue toilet and the walls. &
        there are many cars driving down the street corner. &
        some baseball players are on the field playing baseball. \\
 
        \hd{\methodName{}} &
        a man in a hat sitting on a bench. &
        a man sits on a wooden bench with a bag on his back. &
        a fire hydrant on a sidewalk next to a street pole. &
        a baseball player is swinging his bat as a crowd watches. \\[2ex]

        \large
        \raisebox{.8cm}{\rotatebox{90}{\sc \sffamily Image}} &
        \includegraphics[width=\linewidth]{images/qualitatives/image/example1.jpg} &
        \includegraphics[width=\linewidth]{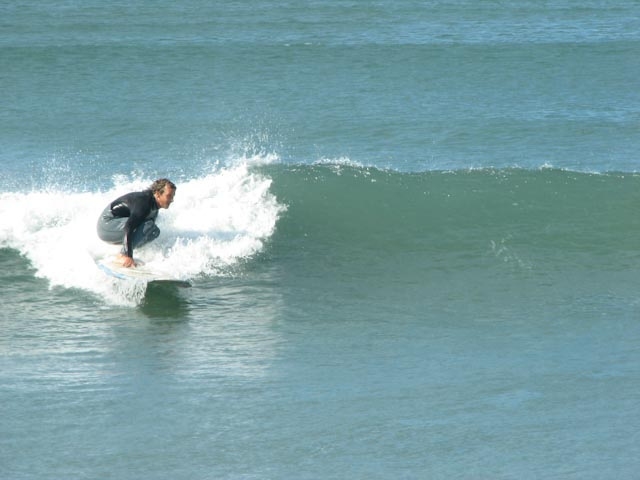} &
        \includegraphics[width=\linewidth]{images/qualitatives/image/example3.jpg} &
        \includegraphics[height=0.195\textwidth]{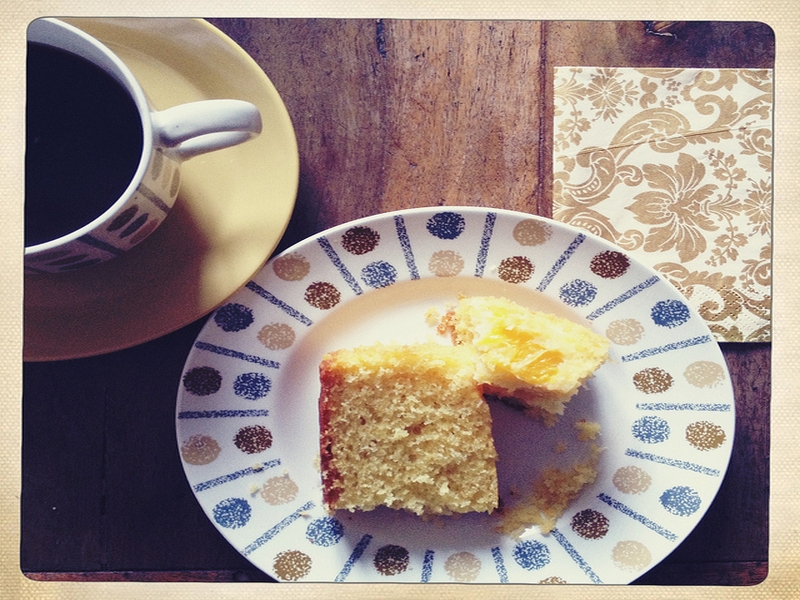} \\

        \rowcolor[gray]{.9}
        \hd{GT} &
        A black cat rubbing against a bottle of wine. &
        A man in a wetsuit rides a wave. &
        A wooden bench sitting on a beach. &
        A wooden table with a plate of cake and coffee. \\ 
        
        \hd{DeCap} &
        a black cat standing next to a bottle of wine glasses &
        a man on a surf board riding a wave in the water &
        a bench sits on the beach next to the ocean &
        a slice of cake on a plate with a cup of cake \\
        
        \rowcolor[gray]{.9}
        \hd{ZeroCap} &
        a Wine dro Pet Cat. &
        a man surfing in the area 0. &
        a beachfront bench. &
        a sunny cake with tea. \\    
        
        \hd{CLOSE} &
        a cat sitting on the counter of a green bottle. &
        a man on a surf board riding a wave in the ocean. &
        a wooden bench sitting in the sand near the ocean. &
        and a cake is sitting on a white plate. \\

        \rowcolor[gray]{.9}
        \hd{\methodName{}} &
        a black cat sitting on a chair next to a bottle of wine. &
        a man on a surfboard riding a wave. &
        a bench sitting on the beach next to the ocean. &
        a piece of cake on a plate with a cup of coffee. \\
        
    \end{tabular}
    }
    \caption{\textbf{Qualitative results.} We report four predictions of our model and compare baselines from the finer (top) to the coarser (bottom) task. For trace captioning examples, the trace time is color-coded from start (red) to end (yellow). \textbf{DeCap} = DeCap applied on the whole image. 
    \textbf{DeCap (Crop)} = DeCap applied on cropped box. 
    \textbf{ZeroCap} = ZeroCap~\cite{tewel2022zerocap} applied to the whole image.
    \textbf{CLOSE} = CLOSE~\cite{gu2023can} applied to the whole image.
    \textbf{Ours (CLIP + Mem.)} = Our patch-based framework using CLIP as backbone and the projection as modality gap mitigation strategy.
     \textbf{Ours (Talk2DINO + Mem.)} = Our patch-based framework using Talk2DINO as backbone and the projection as modality gap mitigation strategy.
    \textbf{GT} = ground-truth caption.}
    \label{fig:examples}
\end{figure*}

\end{document}